\documentclass{article}
\usepackage{iclr27,times}
\iclrfinalcopy
\usepackage{amsmath,amsfonts,bm}

\def\eqref#1{equation~\ref{#1}}

\def\1{\bm{1}}

\DeclareMathAlphabet{\mathsfit}{\encodingdefault}{\sfdefault}{m}{sl}
\SetMathAlphabet{\mathsfit}{bold}{\encodingdefault}{\sfdefault}{bx}{n}

\newcommand{\E}{\mathbb{E}}

\usepackage{hyperref}
\hypersetup{hidelinks,
    pdftitle={Harness Compilation: Which Decisions Should a Small Vision-Language Model Keep?},
    pdfauthor={Minhao Fan, Yinyi Liu, Jiayu Zhao, Zihan Teng, Song Chen, Weichen Liu}}
\usepackage{url}
\usepackage{graphicx}
\usepackage{booktabs}
\usepackage{multirow}
\usepackage{array,tabularx}
\usepackage{longtable}
\usepackage{needspace}
\usepackage{flafter} 
\usepackage{amsmath,amssymb}
\usepackage{xcolor}
\usepackage{listings}
\usepackage{placeins} 
\lstdefinestyle{prompt}{basicstyle=\fontencoding{T1}\selectfont\scriptsize\ttfamily, breaklines=true, breakatwhitespace=false, columns=fullflexible, keepspaces=true, frame=single, framerule=0.3pt, rulecolor=\color{black!40}, backgroundcolor=\color{black!3}, xleftmargin=2pt, xrightmargin=2pt, aboveskip=4pt, belowskip=8pt, captionpos=t, showstringspaces=false, upquote=true}

\usepackage{caption}
\usepackage{subcaption}

\usepackage{titletoc}

\newcommand{\hc}{\textsc{HC}}
\newcommand{\dof}{DoF}
\newcolumntype{Y}{>{\raggedright\arraybackslash}X}
\newcommand{\ns}{^{\mathrm{ns}}}
\newcommand{\bfs   }[1]{\textbf{#1}}

\newcommand{\tablestyle}{
    \scriptsize
    \setlength{\tabcolsep}{0.1pt}
    \renewcommand{\arraystretch}{1.0}
}

\newcommand{\appendixorganization}{%
    \begin{center}
        {\Large\bfseries Appendix Contents}
    \end{center}
    \vspace{6pt}
    \noindent\textbf{Organization.}
    The appendices group implementation, controlled comparisons, task-specific evidence, and reliability checks.
    \par\smallskip
    \begingroup
    \setlength{\parskip}{0pt}
    \setcounter{tocdepth}{2}
    \titlecontents{section}
        [2.8em]
        {\fontsize{9}{10.5}\selectfont\bfseries}
        {\hyperlink{appendix.\thecontentslabel}{\contentslabel{2.8em}}}
        {}
        {\titlerule*[0.55pc]{.}\contentspage}
        [\vspace{3pt}]
    \titlecontents{subsection}
        [4.3em]
        {\fontsize{9}{10.5}\selectfont}
        {\hyperlink{subsection.\thecontentslabel}{\contentslabel{2.8em}}}
        {}
        {\titlerule*[0.55pc]{.}\contentspage}
        [\vspace{0pt}]
    \hrule height 0.4pt
    \vspace{4pt}
    \printcontents[appendix]{}{1}{}
    \vspace{1pt}
    \hrule height 0.4pt
    \endgroup
    \vspace{6pt}
}

\title{Harness Compilation: Which Decisions Should a Small Vision-Language Model Keep?\thanks{An earlier version of this work is currently under review.}}

\author{%
    \begingroup
        \newcommand{\namefont}{\bfseries\fontsize{10.5}{12}\selectfont}
        \newcommand{\affilfont}{\normalfont\mdseries\fontsize{10}{12}\selectfont}
        \hspace*{-7pt}\begin{tabular}{@{}l@{}}
            {\namefont Minhao Fan$^{1}$%
            \thanks{\raggedright Correspondence to: \texttt{minhao001@e.ntu.edu.sg} and \texttt{liu@ntu.edu.sg}.}%
            \quad Yinyi Liu$^{1}$ \quad
            Jiayu Zhao$^{1,2}$ \quad Zihan Teng$^{1,2}$} \\
            {\namefont Song Chen$^{2}$ \quad Weichen Liu$^{1}$} \\[3pt]
            {\affilfont $^{1}$College of Computing and Data Science, Nanyang Technological University} \\
            {\affilfont $^{2}$School of Microelectronics, University of Science and Technology of China}
        \end{tabular}
    \endgroup
}

\begin{document}
\maketitle
\lhead{Preprint.}

\begin{abstract}

Small vision-language models may be able to read external evidence yet struggle to obtain it. We introduce Harness Compilation (HC), an offline procedure that adapts the division of work between a frozen small VLM and its external harness. A large teacher uses student execution traces to revise reusable content and control, while a separate validation set selects the deployed harness. Deployment requires neither weight updates nor teacher calls. Across seven visual question-answering settings with students of at most 9B parameters, HC improves scores over bare students by 9.9--23.9 points, averaged over three independent builds per setting. Interventions on five runtime decision types (invocation, selection, argument generation, evidence integration and abstention) show why this allocation matters: requesting evidence and generating open queries can be costly, whereas bounded choices and reading supplied text can remain useful student work. Fact cards benefit all ten evaluated students, but decision policies transfer unevenly. Recompilation for a new student model helps when the transferred interface no longer fits the student. With 100 practice items, HC exceeds answer-only LoRA on three tasks. Larger training budgets can match or surpass a fixed harness, while combining the two improves SlideVQA beyond either alone. These findings support allocating work from measured student behavior rather than uniformly removing decisions.

\end{abstract}

\section{Introduction}\label{sec:intro}

Consider a visual assistant that answers image queries locally under limited memory and compute. Reusing a small vision-language model (VLM) avoids maintaining separate weights for each workflow. We focus on models with at most 9B parameters, adapted without weight updates or large-model calls at deployment. Privacy, unreliable connectivity and scarce labels motivate this setting. A large teacher is available only during offline adaptation.

An external harness supplies the instructions, memory and execution procedures around the model. These resources help only if the student can use them. Accessing evidence and interpreting it are different parts of the task: a student may answer correctly when a relevant fact card is supplied, yet fail when it must first request a lookup, choose an entry or generate a useful query. In that case, the obstacle arises before the evidence is read. A harness can therefore impose demands that exceed the capabilities of the model it is meant to assist.

Our central hypothesis is that a harness's benefit depends not only on what it provides, but also on which decisions it leaves to the model. Which decisions should a small VLM retain? Recent work automates harness construction through code search and execution feedback~\citep{metaharness,autosaddler,bhsm}, including systems for frozen VLMs~\citep{dynamo}. We ask how the executor's observed behavior should shape this division of work.

\begin{figure}[!t]
    \centering
    \includegraphics[width=\linewidth]{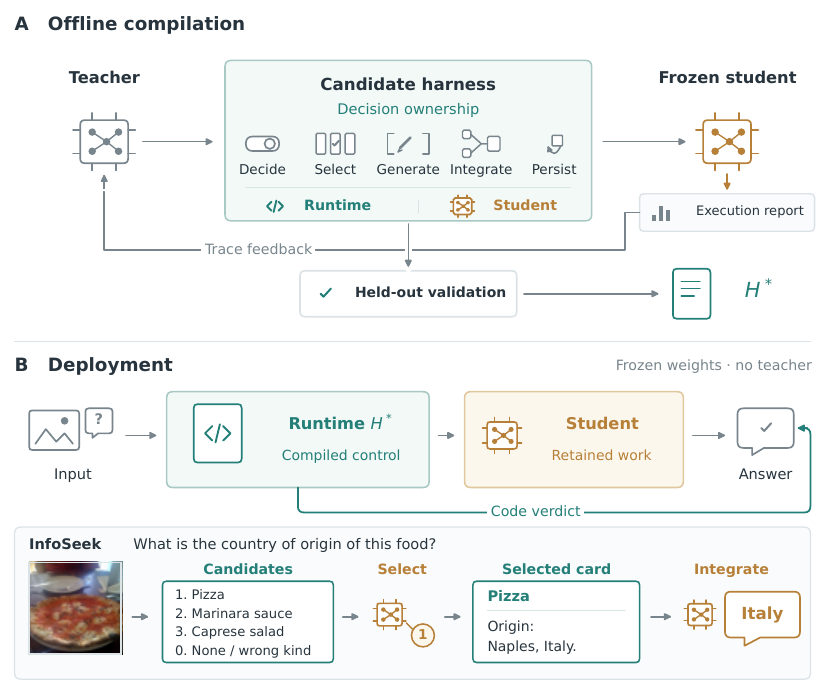}
    \caption{\textbf{Compile the division of work for reuse at deployment.} (A) Within a fixed runtime, the teacher revises the latest proposal from probe traces, while disjoint validation selects $H^*$. (B) Local control invokes the frozen student or emits a permitted code verdict. In the archived InfoSeek test example, the student selects Pizza, then reads its card to answer Italy. Both calls receive the image, with fresh context for answering. Card text is excerpted (Appendix~\ref{app:prompts}).}
    \label{fig:overview}
\end{figure}

We introduce \emph{Harness Compilation} (\hc{}), an offline procedure that fits this division of work to a target student. The teacher proposes reusable content and control, executes the candidate through the student's runtime, and revises it from reports. Validation selects the deployed harness. The same frozen model serves different tasks by loading different harnesses. In Figure~\ref{fig:overview}, code supplies candidate names, the student chooses among them, and code provides a fact card for the student to read.
The objective is to retain work the student executes effectively. On InfoSeek, open entity naming is costly, whereas choosing among supplied names can remain useful. Other tasks benefit from a fixed rule or an evidence-page selector. We organize these choices into five decision types and test them through interface interventions. The allocation can change with the task and student even when the external content stays the same.

Across seven settings, HC raises the macro-average task score from 48.3 to 62.9, improving over bare students in all 21 main builds. Six settings use the same frozen Qwen3-VL-4B with task-specific harnesses. HC also remains useful with 100 practice items and can complement weight adaptation.

\Needspace{18\baselineskip}
Our contributions are:
\begin{itemize}
    \item \textbf{Harness compilation with execution feedback.} We develop an offline procedure in which a teacher revises reusable content and executable control within a developer-specified runtime. Student traces guide revisions, and disjoint validation selects the deployed harness. One frozen model serves different tasks by loading different harnesses, without teacher calls or weight updates at deployment.
    \item \textbf{An empirical study of runtime decision allocation.} Interventions on five decision types distinguish costly access and argument generation from useful bounded choices and evidence reading. Cross-student experiments show how this allocation changes with the executor and when recompilation can repair a transferred interface.
    \item \textbf{Low-data adaptation compatible with weight updates.} HC outperforms answer-only LoRA with 100 practice items on three tasks. Combining the same harness with tuned students improves both document and multi-page reading. Label-scaling experiments also identify when a stronger student needs less external control.
\end{itemize}

\section{Related work}\label{sec:related}

\paragraph{Test-time scaling.} These methods increase the work performed for each query. Self-consistency samples and votes over reasoning paths~\citep{selfconsistency}, adaptive visual inference allocates rollouts by query~\citep{avis}, and test-time augmentation combines predictions from transformed inputs~\citep{ttaug}. Visual programs and tool-using agents also place decisions inside the inference loop~\citep{vipergpt,visprog,toolllm}. HC fits reusable content and control offline. Our compute-only and student-driven tool controls compare these alternatives using the target students and available resources.

\paragraph{Prompt and context optimization.} These methods improve what a model is asked to read and follow. DSPy and MIPROv2 optimize instructions and demonstrations within a program~\citep{dspy,miprov2}. GEPA uses reflective feedback~\citep{gepa}, TextGrad provides textual critiques~\citep{textgrad}, and ACE evolves a standing context~\citep{ace}. We distinguish the search procedure from the object it can edit. HC exposes memory, routing and executable control as well as prompts. Section~\ref{sec:compiler} examines this distinction by allowing another optimizer to edit the same harness.

\paragraph{Harness optimization.} This line of work revises the system around a model using execution evidence. Meta-Harness and AutoSaddler search over harness code~\citep{metaharness,autosaddler}, while related work adapts harnesses and contexts to smaller or video models~\citep{ai4ai,bhsm,videoharness}. HARNESSEVO studies which components account for gains~\citep{harnessevo}, while \citet{harnessupdating} examine whether models can activate and follow useful updates. Reusable tools and controllers also move work outside the model~\citep{latm,agentoptimizer,agentdistill}, including Dynamo for frozen VLMs~\citep{dynamo}. HC belongs to this line of work. Its focus is runtime decision allocation, studied through interface interventions, cross-student transfer and recompilation after weight adaptation.

\section{Problem: which decisions should the student keep?}\label{sec:problem}

We study a task with labelled practice data and a held-out test set. A frozen student $S$ answers under a harness $H$, which supplies prompts, tools, routing, retrievable content and, where needed, bounded candidate choices. The practice data are divided into a probe subset that produces execution feedback and a disjoint validation subset that selects the deployed harness. A large teacher may revise the harness offline but cannot be called at deployment. An output contract specifies the expected answer format and whether abstention is allowed.

\Needspace{8\baselineskip}
Harness components describe what the system contains, while decision types describe what the student is asked to do. We use five operational categories, or degrees of freedom (\dof{}):
\begin{quote}
\textbf{Decide:} whether to invoke a tool or memory.\\
\textbf{Select:} which tool or candidate entry to use.\\
\textbf{Generate:} an open argument, such as a query, entity name or program.\\
\textbf{Integrate:} how to use supplied evidence to answer.\\
\textbf{Persist:} when to stop or abstain.
\end{quote}
These categories describe exposed interfaces, not independent capability axes. A task need not expose all five, and changing one interface can affect another. Generating a final answer does not require an open lookup argument. A retrieved fact card and a rule-based verdict offer different ways to help the student, but each must be useful on the inputs routed to it. A strong offline teacher alone does not guarantee this.
The practical question is therefore twofold: does the resource improve answers, and can the student reliably make the decisions needed to use it? Compilation tests this division of work through execution rather than assuming that fewer student decisions are always better.

\section{Harness Compilation}\label{sec:method}

\paragraph{Compile a runtime specification.} The compiler prompt is a manual for a task-specific runtime (Figure~\ref{fig:overview}). It provides the task, bare-student probe score and resource examples, then specifies allowed observations, editable JSON fields and Python hook contracts. The teacher fills this developer-specified interface with reusable content and control, returning one complete harness JSON with hook source code. Table~\ref{tab:ownership_main} illustrates the resulting division of work.

\begin{table}[htbp]
    \centering\small\renewcommand{\arraystretch}{1.12}\setlength{\tabcolsep}{5pt}
    \caption{\textbf{Examples of the division of work.} Each row describes a fixed harness used in the diagnostic studies. Policies can differ across builds, so these allocations are not universal policies for each task.}
    \label{tab:ownership_main}
    \begin{tabularx}{\linewidth}{@{}lYY@{}}
        \toprule
        Setting & Compiled work & Student retains \\
        \midrule
        InfoSeek & Route retrieval and prepare a card & Choose among names on ambiguous inputs, then answer \\
        DocVQA & Supply OCR and an answer format & Read the page and supplied text \\
        SlideVQA & Rank and select evidence slides & Read selected slides and answer \\
        LiveVQA & Supply three article cards & Select and combine evidence while answering \\
        BLINK & Compute a verdict from measurements & Answer when the rule defers \\
        \bottomrule
    \end{tabularx}
\end{table}

\paragraph{Execute content and control.} Content fields set prompt framing, answer format and memory use. The router's \texttt{select(item)} chooses what evidence to show and how to invoke the student. On InfoSeek, it may supply cards directly or ask the student to choose or name an entity before answering. These modes change the decisions left to the student (Section~\ref{sec:problem}). The router reads observable inputs, excluding benchmark question-type and condition annotations. The \texttt{parse(text)} hook only extracts a span of the student's output. BLINK separately permits code verdicts with student fallback.

\paragraph{Revise from execution.} The teacher first writes $H_0$, which the runtime executes on the probe subset. Main builds use three reflection rounds and, except on DocVQA, no separate student profile. Each round edits the latest proposal, including a rejected one, while validation tracks the best harness for deployment.
Execution reports connect failures to editable decisions: a wrong entity suggests checking routing or selection, while a wrong answer with the correct entity suggests checking evidence sufficiency and reading instructions. This separation helps target revisions, although a teacher rationale does not establish causality when multiple components change together. InfoSeek reports include 30 stratified failure traces, pick precision and abstention, with one additional trace query per round.

\paragraph{Select and deploy.} A disjoint validation subset selects the deployed harness $H^*$. A proposal replaces the incumbent only when its unrounded mean validation score strictly improves, and no candidate below bare validation performance is retained. InfoSeek and DocVQA allow up to two restarts when $H_0$ scores below bare on the probe. Test items never reach the compiler. Deployment executes $H^*$ with local tools and the frozen student, without teacher calls. Appendix~\ref{app:implementation} provides the runtime contracts, prompts and evaluation protocols.

\Needspace{8\baselineskip}
\paragraph{When should a decision stay with the student?}\label{sec:decisioncriteria}
First, can the available resource improve answers? A channel $c$ is a deployed path, such as answering from a retrieved card or using a tool verdict. Let $D_R$ be the items routed through it. Its \emph{channel edge} compares that path's answer accuracy with the bare student on the same items:
\begin{equation}
    \Delta_c(D_R)=\mathrm{acc}_c(D_R)-\mathrm{acc}_S(D_R).
    \label{eq:channel_edge}
\end{equation}
The edge must be positive on a subset the router can identify from observable inputs. A strong offline teacher does not ensure this: the runtime must retrieve useful evidence and the student must consume it reliably.

Second, should the student make a decision within that path? Choosing among supplied names may beat taking the linker's first match, while unrestricted naming may fail. Let $\pi_S$ and $\pi_0$ denote student and mechanical decision precision on $D_R$. The student decision is worth retaining when
\begin{equation}
    \pi_S(D_R)-\pi_0(D_R)-a_S(D_R)>0,
    \label{eq:retain_criterion}
\end{equation}
where $a_S$ is the accuracy lost to contract-induced abstentions, not their raw frequency. All terms use the same routed items and accuracy scale. These criteria interpret interface interventions. HC selects the complete harness by validation without optimizing either expression directly. Failure may reflect missing tools, poor retrieval or an unsuitable contract, rather than reasoning that cannot be compiled.

\section{Experiments}\label{sec:results}

We test overall effectiveness and execution feedback, then examine the decision interfaces behind the gains, their response to executor changes, and generation costs. The seven settings cover knowledge retrieval (InfoSeek), document reading (DocVQA), multi-page reading (SlideVQA), news-image question answering (LiveVQA), and three perceptual settings from BLINK. All use Qwen3-VL-4B except DocVQA, which uses Gemma-4-E4B. GPT-5.5 is the default teacher. Student weights are frozen and decoding is greedy unless stated otherwise. DocVQA uses ANLS, SlideVQA exact match, and the other settings accuracy, with task-specific scoring details in Appendix~\ref{app:protocol}.

Each main build uses 150 probe and 300 validation items, except the smaller BLINK folds. Selecting the gate additionally consumed 300 disjoint items on each of InfoSeek and DocVQA, bringing development use to 750 unique items on those tasks. This outer selection budget is separate from the 450 items used within a build. We repeat compilation three times per setting. BLINK uses three cross-build folds within each replicate, so every item is tested only with a harness built on the other folds.

Overall and initial-to-selected results average three independent builds per setting. Later mechanism studies fix a diagnostic harness within each comparison. The token study pairs scores and generation counts from single executions of the formal replicate-1 harnesses. These HC scores are not repeats of the main means or stages in a common progression. Each difference compares the named arms within its own experiment.

\subsection{Overall effectiveness}\label{sec:main}

Table~\ref{tab:main} compares the bare student, a student-routed control that exposes available resources with a plain instruction, a manually specified human recipe, and HC. Human recipes are alternative task-specific policies rather than an ordered level above student routing. HC improves over bare students in all seven settings and all 21 builds, raising the macro-average task score from 48.3 to 62.9. Resources alone recover part of this gain. HC adds most over student routing on InfoSeek, SlideVQA, reflectance and depth. On LiveVQA and correspondence, the three-build means differ by less than two points.

\begin{table}[!htb]
    \centering\small\setlength{\tabcolsep}{7pt}\renewcommand{\arraystretch}{1.12}
    \caption{\textbf{Overall test scores.} HC averages three independent builds per setting. Other columns are the specified comparators. Higher is better. The macro-average weights all seven task-score scales equally. Human recipes are alternative policies, not an upper bound.}
    \label{tab:main}
    \begin{tabular}{lrrrr}
        \toprule
        Setting & Bare & Student-routed & Human recipe & HC (3 builds) \\
        \midrule
        InfoSeek & 18.0 & 29.1 & 33.8 & \textbf{37.7} \\
        DocVQA & 72.7 & 80.4 & 82.4 & \textbf{85.0} \\
        SlideVQA & 44.7 & 44.4 & 48.7 & \textbf{54.6} \\
        LiveVQA & 14.7 & 36.9 & 25.5 & \textbf{38.6} \\
        BLINK reflectance & 50.8 & 50.0 & 59.0 & \textbf{62.9} \\
        BLINK correspondence & 54.9 & 64.9 & 64.2 & \textbf{65.7} \\
        BLINK depth & 82.3 & 73.4 & 94.4 & \textbf{95.4} \\
        \midrule
        Average & 48.3 & 54.2 & 58.3 & \textbf{62.9} \\
        \bottomrule
    \end{tabular}
\end{table}

Preselecting evidence can be restrictive. On LiveVQA, the human recipe first asks the student to choose one headline and then supplies only that article, scoring 25.5. The student-routed control reads all three candidate cards and scores 36.9. Choosing before reading can discard evidence useful for answering. These arms also differ in prompts and output formats, so the gap does not isolate selection accuracy. HC largely retains the student's reading and evidence choice, consistent with its smaller additional gain here.

Well-designed fixed interfaces remain competitive. Fixed BLINK rules are strong comparators, and a separate unoptimized typed-input/output program reaches 86.2 on DocVQA, above HC's 85.0 mean. This program is distinct from the 82.4 human recipe in Table~\ref{tab:main}. Compilation finds a useful interface for the target student, but does not necessarily improve on every well-designed fixed interface.

\subsection{What does execution feedback contribute?}\label{sec:compiler}

Feedback-guided revision and validation improve the initial harnesses from a macro-average test score of 60.4 to 62.9 (Table~\ref{tab:compilation_main}). InfoSeek, DocVQA and LiveVQA improve in every build. SlideVQA improves in two of three, while BLINK changes little, with depth retaining its initial rule in every fold. One validation-accepted SlideVQA edit loses 1.8 points on test. Earlier independent one-shot builds also match the loop on some tasks, although their different manual prevents isolating the effect of iteration.

\begin{table}[!htb]
    \centering\small\setlength{\tabcolsep}{16pt}\renewcommand{\arraystretch}{1.1}
    \caption{\textbf{Revision and validation selection after the initial teacher-built harness.} Means over the same three builds per setting, evaluated on identical test items. Validation alone selects the retained round.}
    \label{tab:compilation_main}
    \begin{tabular}{lrr}
        \toprule
        Setting & Initial $H_0$ & Selected $H_{\mathrm{val}}$ \\
        \midrule
        InfoSeek & 33.5 & 37.7 \\
        DocVQA & 78.4 & 85.0 \\
        SlideVQA & 54.4 & 54.6 \\
        LiveVQA & 34.1 & 38.6 \\
        BLINK reflectance & 61.7 & 62.9 \\
        BLINK correspondence & 65.2 & 65.7 \\
        BLINK depth & 95.4 & 95.4 \\
        \midrule
        Average & 60.4 & 62.9 \\
        \bottomrule
    \end{tabular}
\end{table}

Stage diagnostics distinguish retrieval failures from failures to use supplied evidence, directing revisions to different interfaces. Validation then chooses among the proposals, so the initial-to-selected gain measures search and selection together. The setup budget includes the extra data used to select the gate.
What the optimizer can edit also matters. When GEPA revises the same executable harness with a fixed store, its mean score is similar to our loop on two matched runs, with build variation exceeding the difference. Prompt-only search exposes fewer structural choices. These controls support editable content and control without establishing a universally best search algorithm or feedback format (compiler ablations in Appendix~\ref{app:compiler}, matched optimizer controls in Appendix~\ref{app:gepa}).

\subsection{Which decisions should the student keep?}\label{sec:carry}

\begin{figure}[!htb]
    \centering
    \includegraphics[width=\linewidth]{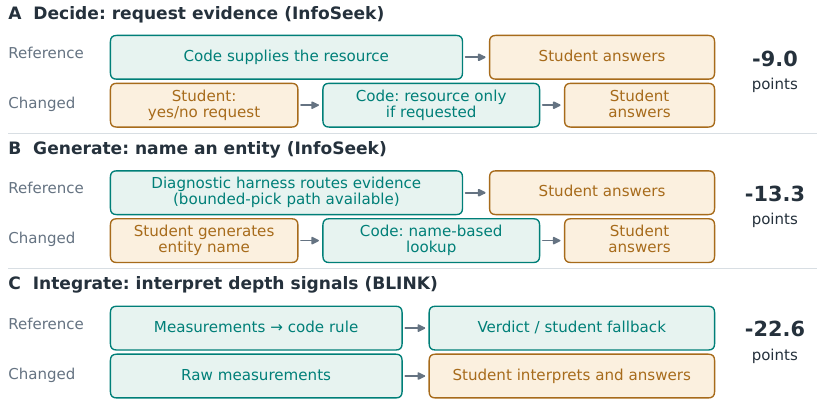}
    \caption{\textbf{Changing the interface changes the work left to the student.} Three interventions from Table~\ref{tab:onedec_main}. Differences are changes from the fixed diagnostic reference, not additive capability effects. Naming changes the lookup interface, while raw BLINK signals also remove compiled verdict control.}
    \label{fig:interfaces}
\end{figure}

We test the division of work by changing interfaces within fixed diagnostic harnesses. Figure~\ref{fig:interfaces} illustrates three changes, and Table~\ref{tab:onedec_main} reports all comparisons. Negative changes mean that an intervention hurts its reference harness. These are operational comparisons, not independent capability measurements.

\begin{table}[!htb]
    \centering\small\setlength{\tabcolsep}{5pt}\renewcommand{\arraystretch}{1.13}
    \caption{\textbf{Effect of changing one decision interface.} Paired score changes from one fixed reference harness per setting, with negative values indicating harm. These references differ from the three-build means in Table~\ref{tab:main}. $^{\mathrm{ns}}$: $p>0.05$. n/a: untested. Each column changes how a decision is executed, rather than measuring an isolated capability. Appendix~\ref{app:free} defines the interventions and comparison conditions.}
    \label{tab:onedec_main}
    \begin{tabular}{lrrrrr}
        \toprule
        Setting & Decide & Select & Generate & Integrate & Persist \\
        \midrule
        InfoSeek & $-9.0$ & $-0.5\ns$ & $-13.3$ & $-3.1$ & $-4.3$ \\
        DocVQA & $-2.7$ & $+0.5\ns$ & n/a & $+0.1\ns$ & $+0.6\ns$ \\
        SlideVQA & $-0.5\ns$ & $-12.1$ & n/a & $+0.6\ns$ & $-2.8$ \\
        LiveVQA & n/a & n/a & n/a & $-0.1\ns$ & $+0.2\ns$ \\
        BLINK reflectance & $+0.0\ns$ & n/a & n/a & $-14.2$ & $-4.5\ns$ \\
        BLINK correspondence & $-2.5$ & n/a & n/a & $-2.3\ns$ & $-1.4\ns$ \\
        BLINK depth & $+0.0\ns$ & n/a & n/a & $-22.6$ & $-4.0\ns$ \\
        \bottomrule
    \end{tabular}
\end{table}

A student cannot use evidence it never obtains. Requiring an explicit resource request costs 9.0 points on InfoSeek and 2.7 on DocVQA. Removing SlideVQA's evidence-slide selector costs 12.1 points. These interfaces put an obstacle before reading, although the request intervention has no statistically supported cost on SlideVQA, reflectance or depth.
The form of the access decision also matters. Requiring an entity name for lookup costs 13.3 points on InfoSeek, whereas removing the diagnostic harness's bounded pick changes the score by only $-0.5$ points, nonsignificantly. That pick is used on some inputs, and its effect varies across builds. Removing it also changes context, so this comparison does not isolate decision ownership.

Processing can make numerical evidence usable. On BLINK, code converts measurements into a verdict. Replacing this interface with raw signals costs 14.2 points on reflectance and 22.6 on depth. The raw-signal arm also removes compiled verdict control, so these results concern the operational interface rather than isolated integration ability.

Readable text presents a different trade-off: further processing can remove useful evidence. In the six formal InfoSeek harnesses tested in Table~\ref{tab:mechanism_main}, restoring the stored cards' original lines under the same character budget improves scores by 0.27--2.87 points. These are different reference harnesses from the integration contrast in Table~\ref{tab:onedec_main}. Short guidance derived from the question text can still help: removing such hints hurts two of the three reported-protocol harnesses, while the third change is not significant. A selected harness can therefore contain both helpful guidance and harmful evidence compression.

\begin{table}[!htb]
    \centering\small\setlength{\tabcolsep}{5pt}\renewcommand{\arraystretch}{1.14}
    \caption{\textbf{Controls within formal InfoSeek harnesses.} Each change is paired with its own fixed harness on 6,000 test items. Ranges span harnesses, not confidence intervals. All use interface v2, with A editing the last proposal and B the incumbent. Extractor and contract rows report A and B separately and do not decompose the diagnostic $-4.3$ effect. Full controls: Appendix~\ref{app:rcontrols}.}
    \label{tab:mechanism_main}
    \begin{tabularx}{\linewidth}{@{}Xlr@{}}
        \toprule
        Change to the selected interface & Evaluated harnesses & Score-change range \\
        \midrule
        Restore stored cards' original lines & A and B (6) & $+0.27$ to $+2.87$ \\
        Remove question-text hints & A (3) & $-2.23$ to $-0.13$ \\
        Default extractor with contract & A (3) / B (3) & $-0.02$ to $0.00$ / $-0.07$ to $0.00$ \\
        Artifact extractor without contract & A (3) / B (3) & $-6.07$ to $-1.78$ / $-2.57$ to $-1.72$ \\
        \bottomrule
    \end{tabularx}
\end{table}

Output contracts matter where response behavior is limiting. Removing the contract and parser together costs 4.3 points on the InfoSeek diagnostic harness and 2.8 on the SlideVQA formal artifact, but little on DocVQA or LiveVQA. Separate controls on the three formal InfoSeek harnesses of each protocol show that the contract carries this effect and the extractor does not (Table~\ref{tab:mechanism_main}). They do not decompose the earlier 4.3-point effect or isolate stopping ability. The objective also matters: LiveVQA student routing fares better under an abstention-sensitive F-score, although HC has higher accuracy (Appendix~\ref{app:livevqa}).
Together, the interventions identify interfaces that block useful evidence and work the student can retain. Their effects are not additive and favor task- and student-specific allocations.

\Needspace{6\baselineskip}
\subsection{Student-dependent decisions and recompilation}\label{sec:predict}

\begin{figure}[!htb]
    \centering
    \includegraphics[width=\linewidth]{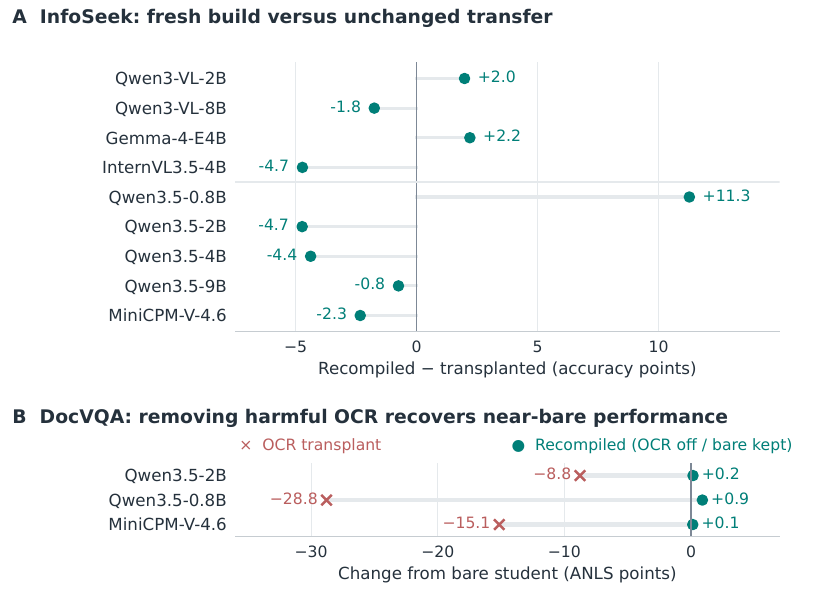}
    \caption{\textbf{Recompilation can repair a mismatched interface.} (A) Fresh InfoSeek builds versus unchanged transfer of the Qwen3-VL-4B diagnostic harness (37.8 on its source student): nine recipients, one fresh build each, interface v2. Positive values favor recompilation (reported mean $-0.4$). (B) DocVQA changes are relative to each bare student. The transplant is the formal replicate-1 Gemma harness, and fresh builds use the current gate with its bare floor. The panels use different baselines. Source harnesses and paired tests are in Appendix~\ref{app:matrix}.}
    \label{fig:students}
\end{figure}

Changing students can change the useful allocation. InfoSeek cards benefit all ten students in the content diagnostics, but bounded selection transfers less reliably: MiniCPM picks no better than the linker, while Qwen3.5-2B often chooses ``none'' and discards useful evidence. A separate comparison in Figure~\ref{fig:students} tests unchanged whole-harness transfer against one fresh build for each of nine recipients, using a different diagnostic reference.

Recompilation helps when it repairs a mismatch, but is not automatically better. Across the nine InfoSeek recipients, fresh builds average 0.4 points below unchanged transfer and help only three students. Qwen3.5-0.8B gains the most, 11.3 points, after struggling with the transferred pick-and-read program. Each recipient has one fresh build, not repeated teacher draws.
A mismatch can also come from the evidence channel. The transferred DocVQA OCR harness harms the receiving document readers. Fresh builds under the current gate turn OCR off or retain the bare model, staying within one point of bare (Figure~\ref{fig:students}). Here, removing an unhelpful input improves the interface.
Weight updates can likewise make an old decision policy redundant, motivating the component comparisons below.

\Needspace{6\baselineskip}
\subsection{Data efficiency and weight adaptation}\label{sec:weights}\label{sec:bounds}

HC is useful when task labels are scarce. With 100 practice items, split into 50 probe and 50 validation items, HC exceeds answer-only LoRA on all three tasks in Table~\ref{tab:lowdata_main}. On InfoSeek, ten-epoch trajectory training (35.3) exceeds the HC mean (33.6), whose three builds span 25.9 to 38.0. HC is therefore a useful low-data alternative, without uniformly outperforming weight adaptation. Matched item counts also do not imply equal supervision: HC uses an offline teacher and external resources, whereas answer-only LoRA learns from gold answers.

\begin{table}[!htb]
    \centering\small\setlength{\tabcolsep}{10pt}\renewcommand{\arraystretch}{1.12}
    \caption{\textbf{Adaptation with 100 practice items.} Mean test scores over three runs. HC uses 50 probe and 50 validation items. Training columns use the stronger reported epoch variant. Schedules and build-level results are in Appendix~\ref{app:ladder}.}
    \label{tab:lowdata_main}
    \begin{tabular}{lrrr}
        \toprule
        Setting & HC & Answer LoRA & Trajectory SFT \\
        \midrule
        InfoSeek & 33.6 & 23.3 & \textbf{35.3} \\
        DocVQA & \textbf{84.7} & 81.1 & n/a \\
        SlideVQA & \textbf{54.7} & 47.3 & n/a \\
        \bottomrule
    \end{tabular}
\end{table}

The two adaptation routes can complement one another. On SlideVQA, the fixed harness scores 58.1 and the student tuned with 1,300 labels scores 53.2. Applying the same harness to the tuned student reaches 63.7, while on DocVQA the harness adds 1.8 points to a 999-label adapter (Table~\ref{tab:combined_main}). This comparison establishes compatibility, without identifying which skills entered the weights. Weight adaptation can also make an old interface harmful. With enough external reflectance labels, LoRA surpasses the frozen harness, whose control then reduces the tuned student's performance. Recompilation removes most of that control and recovers the tuned student's score. Weight updates can thus complement a harness or change which parts remain useful.

\begin{table}[!htb]
    \centering\small\setlength{\tabcolsep}{7pt}\renewcommand{\arraystretch}{1.12}
    \caption{\textbf{Fixed-harness component comparisons.} The same serialized harness, verified by identical hashes, is used before and after tuning. LoRA labels count training items. Harness practice budgets are 450 (SlideVQA) and 999 (DocVQA). These are separate component studies, not the formal three-build means. Training and label-scaling protocols: Appendix~\ref{app:weights}.}
    \label{tab:combined_main}
    \begin{tabular}{lrrrr}
        \toprule
        Setting & LoRA labels & Fixed HC & LoRA & LoRA + HC \\
        \midrule
        SlideVQA & 1,300 & 58.1 & 53.2 & \textbf{63.7} \\
        DocVQA & 999 & 85.4 & 84.7 & \textbf{86.4} \\
        \bottomrule
    \end{tabular}
\end{table}

\subsection{Generation efficiency and offline cost}\label{sec:token_efficiency}\label{sec:cost}

HC reduces online generation through reusable content that shortens the student's response process and code verdicts that avoid a student call altogether. On the formal replicate-1 harnesses, HC scores higher than chain-of-thought sampling with voting on InfoSeek, DocVQA and SlideVQA while generating about 69--189 times fewer student tokens (Table~\ref{tab:tokens_main}). On BLINK, voting is not significantly different from HC on reflectance or correspondence and is below it on depth. Fractional token counts average over all items, including those answered without a student call.

\begin{table}[!htb]
    \centering\small\setlength{\tabcolsep}{8pt}\renewcommand{\arraystretch}{1.1}
    \caption{\textbf{Task performance and generated student tokens.} Each HC row pairs the score and token count from one execution of the formal replicate-1 harness (fold artifacts on BLINK). CoT+vote records the comparator runs on the same items. Counts include all sampled rollouts and student passes, excluding input prefill, tools and offline compilation. Appendix~\ref{app:cost} gives the accounting.}
    \label{tab:token_efficiency}\label{tab:tokens_main}
    \begin{tabular}{lrrrr}
        \toprule
        \multirow{2}{*}{Setting} & \multicolumn{2}{c}{Test score $\uparrow$} & \multicolumn{2}{c}{Tokens per item $\downarrow$} \\
        \cmidrule(lr){2-3}\cmidrule(lr){4-5}
         & CoT + vote & HC & CoT + vote & HC \\
        \midrule
        InfoSeek & 24.6 & 37.3 & 1,120 & 5.9 \\
        DocVQA & 80.4 & 85.7 & 580 & 8.5 \\
        SlideVQA & 51.1 & 58.1 & 970 & 5.8 \\
        BLINK reflectance & 64.9 & 64.2 & 2,000 & 0.1 \\
        BLINK correspondence & 68.5 & 67.4 & 2,510 & 0.9 \\
        BLINK depth & 85.5 & 96.0 & 2,080 & 0.1 \\
        \bottomrule
    \end{tabular}
\end{table}

These generation savings do not establish end-to-end speedups, as the counts exclude input and image prefill, tool computation and offline teacher use. Reported VQA builds cost about \$0.3--7 in teacher calls and 1--35 student GPU-minutes. On InfoSeek, compilation with an existing store costs about \$1, and constructing the reusable store adds about \$6. Both regimes eliminate teacher calls during deployment but retain local input and tool costs.

\section{Discussion and limitations}\label{sec:discussion}

HC allocates work to match the executor, but decision counts alone do not measure harness quality. Fixed interfaces and other optimizers can also produce effective harnesses.
Our builds depend on initial teacher draws and validation selection. Three builds only partly characterize variation, and BLINK test pools are small. The accepted SlideVQA regression and harmful InfoSeek card compression show that selection can retain poor decisions. The validation floor and branch-support audits do not guarantee improvement on new items. Appendix~\ref{app:reliability} collects the branch-support, overlap and evaluation audits.

The formal evaluation uses observation-derived routing and an extraction-only parser. InfoSeek retrieves over a fixed entity inventory, and the reused store has practice-informed lineage. Formal runs repeat compilation on benchmark splits already used in this research, measuring build variation rather than confirmation on a newly collected test population. Transfer diagnostics likewise characterize the studied behavior rather than establish an independent prospective prediction test.

Our measurements concern task scores and generation costs, not end-to-end industrial deployment or hardware-specific latency. Offline teacher access to practice traces is still required. Residual teacher-student gaps can reflect missing tools, retrieval failures or interface limits and do not establish that reasoning cannot be compiled. Exploratory LIBERO-plus results are reported separately in Appendix~\ref{app:vla}, with completion, refusal and false-abort outcomes kept distinct and excluded from the VQA average.

\section{Conclusion}\label{sec:conclusion}

Harness Compilation uses offline execution feedback and validation to allocate work around a frozen small VLM, raising the macro-average task score from 48.3 to 62.9 across seven VQA settings. Its interventions show that obtaining evidence and generating open lookup arguments can be costly, while bounded choices and evidence reading can remain useful student work. This suggests a practical approach for repeatedly used local document assistants and image-query services: adapt resources and control offline to reduce online generation. On InfoSeek, DocVQA and SlideVQA, HC scores higher than chain-of-thought sampling with voting while generating about 69--189 times fewer student tokens, although these savings do not establish end-to-end speedups. HC also offers an option when labels are scarce, outperforming answer-only LoRA with 100 practice items on three tasks. When weight adaptation is available, the same harness improves tuned students on both SlideVQA and DocVQA beyond either component alone. Together, these results support allocating work from measured student behavior and reassessing that allocation as the model improves.

\bibliography{iclr27}
\bibliographystyle{iclr27}

\clearpage
\appendix
\startcontents[appendix]
\appendixorganization
\clearpage

\section{Implementation, prompts and evaluation protocols}
\label{app:implementation}
\subsection{Tasks, splits and evaluation}
\label{app:protocol}
\label{app:triage}
\paragraph{How to read the main comparisons.}
Table~\ref{tab:main_settings} records the evaluation settings used in the main results. Table~\ref{tab:recipe_registry} identifies the human comparator in each row. Tables~\ref{tab:main} and \ref{tab:compilation_main} summarize the formal three-build evaluation. Later studies use fixed diagnostic harnesses, task-specific adaptation budgets, or historical runs with recorded token counts. Table~\ref{tab:build_registry} identifies those artifacts. Their scores are not interchangeable with the main means. A delta compares the two named arms within its experiment, not arbitrary HC scores across tables. Teacher scores are listed separately in Table~\ref{tab:teacher_registry} because their evaluation populations are not uniform.

\begin{table}[htbp]
\centering\footnotesize\setlength{\tabcolsep}{5pt}\renewcommand{\arraystretch}{1.16}
\caption{Main evaluation settings. Practice counts are probe / validation items per build. The additional outer selection budget is stated below.}
\label{tab:main_settings}
\begin{tabular}{@{}lllrr@{}}
\toprule
Setting & Student & Metric & Test items & Practice \\
\midrule
InfoSeek & Qwen3-VL-4B & Relaxed exact match & 6,000 & 150 / 300 \\
DocVQA & Gemma-4-E4B & ANLS & 3,568 & 150 / 300 \\
SlideVQA & Qwen3-VL-4B & Exact match & 2,215 & 150 / 300 \\
LiveVQA & Qwen3-VL-4B & Judged accuracy & 2,000 & 150 / 300 \\
BLINK reflectance & Qwen3-VL-4B & Accuracy & 134 & 44 / 45 per fold \\
BLINK correspondence & Qwen3-VL-4B & Accuracy & 441 & 147 / 147 per fold \\
BLINK depth & Qwen3-VL-4B & Accuracy & 124 & 41 / 42 per fold \\
\bottomrule
\end{tabular}
\end{table}

The overall and initial-to-selected results (Tables~\ref{tab:main} and \ref{tab:compilation_main}) use GPT-5.5, interface v2, the reported last-proposal reflection protocol (A), and the frozen strict-improvement gate. InfoSeek, LiveVQA and SlideVQA builds receive no separate student profile. The DocVQA builds (and every other Gemma DocVQA build in this paper) receive the Gemma profile block in the compile request, a deviation from the no-profile protocol that we record rather than rerun. Three independent teacher draws are run per setting, with BLINK retaining three cross-build folds within each replicate. Each build uses the probe/gate allocation shown above. Gate selection additionally consumed 300 disjoint selection items on each of InfoSeek and DocVQA, bringing development use on those tasks to 750 unique items. This outer selection budget is separate from the 450 items used within a build (Appendix~\ref{app:formal}). The earlier single interface-v2 builds (37.8, 84.1 and 38.1 on InfoSeek, DocVQA and LiveVQA) are fixed diagnostic references, identified in Table~\ref{tab:build_registry}. LiveVQA uses April and May 2025 news images and the Claude Sonnet 5 judge, which did not participate in compilation. Appendix~\ref{app:livevqa} separates judged accuracy, containment and the abstention-sensitive F-score.

\begin{table}[htbp]
\centering\small\setlength{\tabcolsep}{5pt}\renewcommand{\arraystretch}{1.14}
\caption{Human recipes used in Table~\ref{tab:main}. Full prompts are in Appendix~\ref{app:human}.}
\label{tab:recipe_registry}
\begin{tabularx}{\linewidth}{@{}lYr@{}}
\toprule
Setting & Human recipe & Score \\
\midrule
InfoSeek & Select a candidate name, then read retrieved Wikipedia passages & 33.8 \\
DocVQA & Supply page OCR in typed fields & 82.4 \\
SlideVQA & Read the three slides selected by the text ranker & 48.7 \\
LiveVQA & Select one candidate headline and read its article card & 25.5 \\
BLINK reflectance & Fixed Retinex verdict rule & 59.0 \\
BLINK correspondence & Fixed LightGlue and DINO rule & 64.2 \\
BLINK depth & Fixed depth-estimator rule & 94.4 \\
\bottomrule
\end{tabularx}
\end{table}

These recipes are fixed or development-tuned references rather than a matched-budget comparison with HC. The InfoSeek recipe uses public Wikipedia content, not the teacher-written fact store. The tuned InfoSeek placement policy over the shared teacher store scores 37.4 and is a separate reference. Likewise, the unoptimized DSPy DocVQA program scores 86.2 and is distinct from the typed-field recipe at 82.4. The student-routed controls use the resources supplied to HC with a plain instruction. Their exact interfaces are defined in Appendix~\ref{app:free}.

\begin{table}[!htb]
\centering\footnotesize\setlength{\tabcolsep}{5pt}\renewcommand{\arraystretch}{1.16}
\caption{\textbf{Which HC result belongs to which comparison?} Scores identify the reference harness, not a new comparison across rows. Formal means use three replicates. All other numerical entries identify single harnesses. Practice counts include probe and validation items.}
\label{tab:build_registry}
\begin{tabularx}{\linewidth}{@{}p{.34\linewidth}rrY@{}}
\toprule
Comparison & InfoSeek & DocVQA & Reference and practice budget \\
\midrule
Formal results (Tables~\ref{tab:main}--\ref{tab:compilation_main}) & 37.7 & 85.0 & Interface v2, reflection A; three replicates, 450 items each \\
Decision interventions (Table~\ref{tab:onedec_main}) & 37.8 & 84.1 & Current runtime; one harness, 450 items (InfoSeek, DocVQA, LiveVQA); SlideVQA and BLINK use the formal replicate-1 artifacts. The BLINK integrate entries are the student-routed arm of Table~\ref{tab:main} \\
Generated-token comparison (Table~\ref{tab:tokens_main}) and compute baselines (Table~\ref{tab:tts}) & 37.3 & 85.7 & Formal replicate-1 harnesses re-executed with token logging; score and tokens from the same execution \\
One-shot construction (Table~\ref{tab:oneshot}) & 39.7 & 84.9 & Legacy runs; 1,200 / 999 items, respectively \\
Fixed-harness comparison with tuning (Table~\ref{tab:combined_main}) & -- & 85.4 & Identical artifact hashes on both students: SlideVQA = the formal replicate-1 harness; DocVQA = a current-protocol build on the 999 adapter-training items \\
Current transfer source (Table~\ref{tab:predict}) & 37.8 & -- & Current runtime; 450 items \\
Runtime-cost records (Table~\ref{tab:cost}) & 39.7 & 85.7 & Historical runs with recorded deployment costs \\
\bottomrule
\end{tabularx}
\end{table}

The LiveVQA decision interventions use the single current-runtime harness at 38.1. The InfoSeek selection diagnostics use a separate 1,200-item harness and its rewritten store (Appendix~\ref{app:matrix}). Per-replicate formal scores are in Appendix~\ref{app:formal}.

The two InfoSeek transfer analyses use different source harnesses. The recompilation comparison fixes a 450-item source, while the selection diagnostics use the earlier 1,200-item source. Appendix~\ref{app:matrix} separates them. Table~\ref{tab:combined_main} combines a 450-item harness with a 1,300-label LoRA on SlideVQA and a 999-item harness with a 999-label LoRA on DocVQA. These are component comparisons, not a common total adaptation budget across tasks.

\begin{table}[htbp]
\centering\small\setlength{\tabcolsep}{5pt}\renewcommand{\arraystretch}{1.15}
\caption{Teacher reference scores. Different populations prevent treating this column as a common upper bound.}
\label{tab:teacher_registry}
\begin{tabularx}{\linewidth}{@{}lrY@{}}
\toprule
Setting & GPT-5.5 & Evaluation population and input \\
\midrule
InfoSeek & 50.0 & 300-item practice subset \\
DocVQA & 93.3 & Full test set, 2,048-pixel images \\
SlideVQA & 64.7 & 150-question practice pool with the linker's top-three slides \\
LiveVQA & 58.3 & 300-question practice pool with the same candidate passages \\
BLINK reflectance & 79.9 & Same items as the student and harness \\
BLINK correspondence & 78.7 & Same items as the student and harness \\
BLINK depth & 89.5 & Same items as the student and harness \\
\bottomrule
\end{tabularx}
\end{table}

\FloatBarrier

The seven primary evaluation settings cover knowledge retrieval (InfoSeek~\citep{infoseek,echosight}), document reading (DocVQA~\citep{docvqa}), three perceptual settings from BLINK~\citep{blink}, news-image question answering (LiveVQA~\citep{livevqa}), and multi-page reading (SlideVQA~\citep{slidevqa}). InfoSeek uses 3,000 practice and 6,000 test items. DocVQA reports ANLS on 3,568 test items. LiveVQA and SlideVQA splits and scorers are specified in Apps.~\ref{app:livevqa} and \ref{app:slidevqa}. Unless stated otherwise, a build uses 150 probe items, 300 disjoint validation items and three reflections. The probe supplies labelled execution traces, while validation scores select the retained harness and are visible in the round history. Final test items are not supplied to the compiler. BLINK uses three cross-build folds, with each item evaluated only by a harness built on the other folds. Practice counts include both probe and validation items. Accepted trajectory counts are reported separately from item budgets (App.~\ref{app:ladder}).

\paragraph{Why these BLINK tasks.} An exploratory survey covered all 14 BLINK tasks. We retain five tasks grouped into three settings (Table~\ref{tab:blink_selection}) because they expose distinct relationships between available measurements and student decisions: weak photometric evidence, geometric matching with semantic residuals, and a strong depth channel. This is a study of these interfaces, not a claim of coverage across all BLINK capabilities. A large teacher--student gap alone was not an inclusion rule: depth has a modest teacher gap but a useful external estimator, whereas IQ Test showed no suitable carrier among the resources screened. The latter remains a diagnostic in App.~\ref{app:reasoning}.
\begin{table}[tbp]
\centering\small
\caption{BLINK task selection: the five selected tasks form three evaluation settings.}
\label{tab:blink_selection}
\begin{tabular}{p{.21\linewidth}rp{.62\linewidth}}
\toprule
Setting & Items & What the comparison tests \\
\midrule
Relative reflectance & 134 & Whether local photometric measurements can support a useful verdict despite incomplete illumination invariance. \\
Correspondence & 441 & Visual (172), semantic (139), and functional (130) correspondence share a matching interface but differ in what geometric signals can resolve. \\
Relative depth & 124 & Whether a depth estimator's measurements are better used through code than interpreted by the student. \\
\bottomrule
\end{tabular}
\end{table}

\paragraph{Statistical reporting.} Binary correctness contrasts use paired exact McNemar tests on the same items. ANLS score contrasts use paired item bootstrap intervals. Analyses that threshold ANLS at 0.5 are identified explicitly. Intervals and nonsignificant contrasts are reported where available. Per-row paired differences retain their original precision before the displayed scores are rounded. Repeated builds are shown individually: variation across teacher draws is distinct from uncertainty over test items. Different prompt contracts, stores, image resolutions and validation budgets are identified rather than pooled as identical runs.

\paragraph{Averages.} Average rows weight the named settings or students equally on their 0--100 score scales rather than pooling accuracy. Formal means and paired differences retain precision before display rounding, so averaging the printed one-decimal cells can differ by 0.1 point from a reported mean. Historical tables retain their source-build averages. We do not average confidence intervals or $p$-values.

\paragraph{Tools.} BLINK reflectance uses IIW-derived images~\citep{iiw}, while correspondence includes SPair-71k~\citep{spair} and FunKPoint~\citep{funkpoint}. Available signals come from DINOv2~\citep{dinov2}, LightGlue~\citep{lightglue}, Depth-Anything-V2~\citep{depthanything}, and the task-specific measurements below. OWLv2~\citep{owlv2} is used in the exploratory policy evaluation (App.~\ref{app:vla}), and weight comparisons use LoRA~\citep{lora}.

\subsection{Harness languages}
\label{app:dsl}
\paragraph{Harness components.} For task $\tau$, let $D_p$ and $D_t$ be labelled practice and held-out test splits. Frozen student $S$ uses $H=(P,T,R,M,C)$: prompts/output contract $P$, executable tools $T$, routing of inputs and tool outputs $R$, retrievable content $M$, and bounded-candidate construction $C$. Teacher $L$ edits these offline and is unavailable at deployment.

\paragraph{Knowledge language (InfoSeek).} Keys: \texttt{prompt} (system, before/after question, card frame with \texttt{\{cards\}}, answer suffix), \texttt{memory} ($k$, \texttt{sim\_floor}, \texttt{max\_card\_chars}, \texttt{store} selector), \texttt{decode.max\_new\_tokens}, \texttt{discriminate\_prompt} (with \texttt{\{options\}} and a ``0. None of them'' option), \texttt{name\_prompt}, \texttt{router\_py} (\texttt{def select(item)} $\to$ mode \texttt{answer}/\texttt{discriminate}/\texttt{name}, cards to show, note), \texttt{parse\_py}, and a store rebuild request honoured once per build (1,380 cards, written from entity names alone, without showing questions or answers to the teacher). Discriminate runs as two passes in separate contexts: the student picks a number, then answers with the chosen card, or without a card if it chose ``none''.
\paragraph{Document language (DocVQA).} \texttt{prompt} (system, before/after, \texttt{tool\_frame} with \texttt{\{tool\_text\}}, suffix), \texttt{tool} (\texttt{ocr} on/off, \texttt{max\_chars}), \texttt{decode}, \texttt{router\_py} (\texttt{def select(item)} $\to$ tool text or none, note), \texttt{parse\_py}.
\paragraph{Perception language (BLINK).} \texttt{prompt}, \texttt{tool.signals} on/off, \texttt{decode}, \texttt{answer\_mode} \texttt{student}$|$\texttt{verdict\_if\_present}, \texttt{router\_py} (\texttt{def select(item)} $\to$ tool text, verdict letter, note), \texttt{parse\_py}. Reflectance signals comprise marker coordinates, RGB at radius 3/8, chromaticity, HSV, luminance, context luminance in 15/40-px annuli, Retinex log-reflectance at $\sigma$ 5/15/40. Correspondence provides DINOv2 patch cosine (3$\times$3 and 1$\times$1) between REF and each candidate, best-match location and its distance to candidates, SuperPoint+LightGlue projection of REF and its distances, match count. Depth provides Depth-Anything-V2 Small and Base at radius 4/10, local std, percentile, vertical position, per-model A$-$B difference.

\Needspace{35\baselineskip}
\subsection{Student profile protocol}
\label{app:profile}
    \begin{table}[!h]
\centering\footnotesize\setlength{\tabcolsep}{4pt}\renewcommand{\arraystretch}{1.08}
\caption{\textbf{Student profiles, one measurement per row.} Rates and precision are percentages, and $\Delta$ denotes a task-score change in points. Greedy decoding on 300 practice items unless the protocol states otherwise. These contract-specific measurements are not a common capability scale.}
\label{tab:profile}
\begin{tabularx}{\linewidth}{@{}Yrrrrr@{}}
\toprule
Measurement & \shortstack{Qwen3-VL\\2B} & \shortstack{Qwen3-VL\\4B} & \shortstack{Qwen3-VL\\8B} & \shortstack{Gemma-4\\E4B} & \shortstack{InternVL3.5\\4B} \\
\midrule
\multicolumn{6}{@{}l}{\emph{Requesting resources and following rules}} \\
Memory activation rate & 0.0 & 82.3 & 89.3 & 97.3 & 58.7 \\
Generated entity-name precision & 0.0 & 30.4 & 36.2 & 15.1 & 0.6 \\
Rule compliance, turn 1 & 36 & 100 & 98 & 100 & 22 \\
Rule compliance, turn 6 & 32 & 100 & 98 & 90 & 18 \\
Correct choice, 1-tool menu & 19 & 66 & 75 & 46 & 45 \\
Correct choice, 3-tool menu & 10 & 38 & 52 & 37 & 48 \\
Correct choice, 7-tool menu & 6 & 30 & 37 & 27 & 39 \\
Unnecessary calls, 7-tool menu & 87 & 98 & 91 & 36 & 46 \\
\addlinespace[3pt]
\multicolumn{6}{@{}l}{\emph{Evidence presentation}} \\
Text demonstrations, $\Delta$ & $-7.0$ & $-7.3$ & $+1.0$ & $+1.7$ & $-2.7$ \\
Image demonstrations, $\Delta$ & $-0.7$ & $-2.7$ & $+2.3$ & 0.0 & $+2.0$ \\
Response framing, $\Delta$ & $-1.3$ & $-8.0$ & $-3.3$ & $+0.3$ & $+0.7$ \\
One-card input, $\Delta$ & $+1.7$ & $+5.3$ & $+5.0$ & $+11.7$ & $+5.3$ \\
Three-card input, $\Delta$ & $-10.7$ & $-2.7$ & $+7.7$ & $+2.7$ & $+3.7$ \\
Card capture fraction (\%) & 38 & 37 & 37 & 59 & 59 \\
Candidate precision when picking & 57.0 & 66.0 & 61.2 & 49.3 & 50.8 \\
\addlinespace[3pt]
\multicolumn{6}{@{}l}{\emph{Output format, retry behavior and OCR offers}} \\
Valid JSON, 1 field & 100 & 100 & 100 & 99.7 & 99 \\
Valid JSON, 6 fields & 100 & 99.7 & 99 & 99 & 27.7 \\
Identical resend after failure & 22.5 & 0.5 & 0.5 & 0.0 & 3.4 \\
Direct answer after failure & 61.9 & 99.5 & 93.8 & 100 & 71.5 \\
OCR request rate, DocVQA & -- & 1.0 & -- & 0.0 & -- \\
OCR request rate, ChartQA & -- & 4.0 & -- & 0.4 & -- \\
\bottomrule
\end{tabularx}
\end{table}

Table~\ref{tab:profile} separates resource requests, evidence presentation and response behavior. The rule-compliance test repeats a one-line rule at turns 1 and 6. Menu tests vary the number of offered tools, and JSON tests vary the required fields. OCR request rates use separate DocVQA and ChartQA offer contracts. A dash denotes an unreported measurement. Activation rates depend on the contract: Gemma activates memory under one lookup interface but does not request OCR under an offer interface. The profile separates syntax compliance, argument precision, selection and response framing. No single score is treated as a general reasoning-capability scale. The compiler-input ablation (App.~\ref{app:ablate}) tests whether these measurements help when supplied as sensitivities or as a menu of executable decisions.

\subsection{Current compiler interface and a formal execution example}
\label{app:prompts}
This section specifies the runtime used by the formal results and follows one selected InfoSeek harness from its generated fields to actual student requests. Runtime interface v2 and reflection protocol A are separate choices: the interface limits what a program may observe and return, while reflection determines which proposal the teacher edits. Protocol B changes reflection while retaining the same runtime (Appendix~\ref{app:reflectv2}).

\paragraph{The runtime contract.} Routers read observation-derived inputs: question text and the top three CLIP candidates on InfoSeek and LiveVQA, or question text and OCR on DocVQA. Benchmark answer-type and condition annotations are not exposed. Any question-type hint must be inferred from the question itself. (One sentence of the DocVQA manual still names the removed question-type field, but the runtime never passes it and the two artifacts whose code reads it take a dead branch.) The \texttt{parse(text)} hook may return only a contiguous span of the raw student output after whitespace and markdown normalization. Otherwise, the runtime uses its default extractor and reports the violation count to the teacher. Proposal-time linting rejects control-code string literals that equal or contain a multi-word store entity name. These constraints permit evidence selection and answer extraction without granting the router reference-answer metadata or the parser permission to compute a new answer. Table~\ref{tab:dslslots} lists the task-specific inputs and editable decisions.

\paragraph{What the compiler receives and returns.} The teacher receives a runtime manual, task context and execution evidence, then returns a complete harness JSON. Table~\ref{tab:compiler_request} summarizes this interface, and Listings~\ref{lst:compile_req}--\ref{lst:reflect_req} give the requests rendered from frozen inputs. Under reflection A, the object to edit is the latest proposal, including a rejected proposal. Validation independently retains the best eligible artifact. The teacher therefore follows a proposal lineage, while deployment follows the validation incumbent. This distinction is separate from runtime versioning.

\begin{table}[htbp]
\centering\small\setlength{\tabcolsep}{5pt}\renewcommand{\arraystretch}{1.12}
\caption{\textbf{Compiler inputs and outputs under the current interface.} The subsequent listings expand this documented contract with archived student requests and generated harness components.}
\label{tab:compiler_request}
\begin{tabularx}{\linewidth}{@{}p{.23\linewidth}Y@{}}
\toprule
Stage & Supplied information or required output \\
\midrule
Runtime manual & Allowed observations, JSON fields and defaults, router modes, code-hook signatures, and output constraints. \\
Initial compilation & Task description, bare-student probe score and resource examples. No separate student profile on InfoSeek, LiveVQA and SlideVQA. The DocVQA builds receive the Gemma profile block (a deviation from the stated protocol, recorded in Appendix~\ref{app:protocol}). \\
Reflection A & Latest proposal JSON and its traced probe report, plus round history with scores and the retained incumbent. The incumbent's JSON is not separately supplied as the edit target. \\
Probe evidence & Stage-level errors, 30 stratified InfoSeek failure traces, pick precision and abstention, with one additional trace query per round. Gold labels serve offline diagnosis only. \\
Teacher response & One complete harness JSON containing prompt, resource and decoding fields, plus executable router and parser source. \\
Selection & Strict improvement of the unrounded validation mean, with the bare-model floor. The selected artifact is frozen before test evaluation. \\
\bottomrule
\end{tabularx}
\end{table}

\begin{table}[htbp]
\centering\scriptsize\setlength{\tabcolsep}{3pt}\renewcommand{\arraystretch}{1.1}
\caption{Division of work under the current runtime interfaces. External control lists operations available to the selected programs. Student work occurs on paths that invoke the frozen model. Individual formal artifacts can use different policies (Appendix~\ref{app:rcontrols}). The compiler fills these fields subject to the input and output constraints above.}
\label{tab:dslslots}
\begin{tabular}{@{}p{1.5cm}p{3.2cm}p{2.7cm}p{2.85cm}p{2.85cm}@{}}
\toprule
Setting & Router inputs & Keys beyond \texttt{prompt}, \texttt{decode} & External control & Student work \\
\midrule
InfoSeek & Question text; CLIP top-3 entity names, similarities and cards from the fixed inventory & \texttt{memory}, \texttt{discriminate\_\allowbreak prompt}, \texttt{name\_prompt}, \texttt{parse\_py} & when to show a card, when to invoke the bounded pick, which card lines to show & bounded choice among $\le 3$ named candidates when routed to it (with the escape), then reading the card and answering \\
LiveVQA & Question text; CLIP top-3 news headlines, similarities and article cards & same keys; no store rebuild & retrieve and present the article evidence; card rewriting & read the passages and answer, with an explicit headline pick on paths that request one \\
DocVQA & Question text and noisy page OCR & \texttt{tool} (\texttt{ocr}, \texttt{max\_chars}), \texttt{parse\_py} & OCR provision and the answer contract & read the page and the supplied OCR, then answer \\
SlideVQA & MiniLM and CLIP slide rankings, per-slide OCR, deck size & \texttt{pages} ($k$, ranker, whole deck), \texttt{tool}, \texttt{parse\_py}; router may return a verdict & evidence-page ranking and selection & read and combine the selected slides, then answer \\
BLINK (3) & the bed's pixel, matcher or depth measurements at the marked points (\S\ref{app:dsl}) & \texttt{tool.signals}, \texttt{answer\_mode}, \texttt{parse\_py} & compute a verdict from the measurements and decide when to defer & answer when the rule defers; in the suggestion variants, read the verdict as advice \\
LIBERO-plus & instruction, nearest training sentence with similarity, detector scores of the instruction's noun phrases, end-effector state, steps since last motion & \texttt{tools} (detector, \texttt{check\_every}, restore), \texttt{persist} (\texttt{max\_steps}, stall) & instruction override and the exposed termination controls (abort, stop) & generate actions from the observations and the current instruction \\
\bottomrule
\end{tabular}
\end{table}

\paragraph{Shared content and its scope.} The InfoSeek inventory is fixed before compilation: its 1,380 entities cover all 9,000 sampled items (3,000 practice and 6,000 test), so the gold entity is always present. Comparators share the inventory and linker. This is retrieval over a fixed inventory, not open-world recognition. Fact cards are written in batches of ten from entity names alone, with a 900-character limit and a JSON mapping from entity IDs to card strings. The card-writing instructions were produced after an earlier teacher had read practice reports. Thus the reused store is practice-informed even though individual card-generation calls contain no questions or reference answers (Appendix~\ref{app:infoseek}). This lineage applies to the shared content and does not make the current runtime an earlier interface.

\paragraph{A concrete formal artifact.} Table~\ref{tab:manifest_formal} identifies the six formal InfoSeek artifacts. Replicates 1--3 are independent teacher draws, not ranks. The listings below use replicate 1 under A: round 1, validation 41.0, test 37.3, one of the three artifacts averaged in Table~\ref{tab:main}. They do not stand for the separate 37.8 single-build intervention reference. The store selector and artifact name are not protocol identifiers. The manifest determines the runtime and reflection procedure. The loop archives teacher replies and probe reports but not the request payloads. The two requests behind this artifact were therefore rendered again from the frozen run inputs (the archived round-0 report, history, initial harness, bare probe score and store note) with the prompt-building code at the launch revision, which is byte-identical to the released code. No inference call was made. The archived compile reply parses to exactly the initial harness and the archived round-1 reflection reply to exactly the selected harness (content hashes in Table~\ref{tab:manifest_formal} and the release manifest), which ties the rendered requests to the executed chain. Listings~\ref{lst:compile_req} and \ref{lst:reflect_req} reproduce them with marked omissions, and Listing~\ref{lst:reflect_reply} the archived rationale of the accepted reply. Each call was a single user message to GPT-5.5 with a 16,000-token completion limit and provider-default sampling. Two stale strings in the manual (``600-item probe set'' for a 150-item probe and the profile header around the no-profile placeholder) are kept as sent.

\begin{table}[!htb]
\centering\footnotesize\setlength{\tabcolsep}{6pt}\renewcommand{\arraystretch}{1.1}
\caption{\textbf{Formal InfoSeek replicates.} Each pair starts from the same initial harness: A uses the reported last-proposal reflection, B the incumbent-anchored variant. All use Qwen3-VL-4B with greedy decoding, sampled GPT-5.5 proposals, the same store and practice split, the current runtime, and the frozen strict gate. Teacher cost is per arm. Round 0 denotes the initial harness. Full test records across settings are in Table~\ref{tab:formal_reps}.}
\label{tab:manifest_formal}
\begin{tabular}{@{}llrrrr@{}}
\toprule
Replicate & Protocol & Selected round & Validation & Test & Teacher \$ \\
\midrule
1 & A & 1 & 41.0 & 37.3 & 1.19 \\
1 & B & 3 & 40.7 & 36.1 & 0.54 \\
2 & A & 1 & 43.3 & 39.1 & 0.83 \\
2 & B & 0 & 40.0 & 36.9 & 0.57 \\
3 & A & 1 & 40.0 & 36.6 & 0.91 \\
3 & B & 2 & 37.0 & 34.6 & 0.71 \\
\bottomrule
\end{tabular}
\end{table}

\paragraph{Read the execution before the code.} Listing~\ref{lst:runtime} contains the text portions of two archived test requests, the student outputs and offline scoring annotations. Image payloads are omitted, and gold answers were not student inputs. The pizza question takes the two-pass path: choose a candidate name, then answer from its card in a fresh context. The mandarin-duck question takes the direct-card path. Listings~\ref{lst:hval}--\ref{lst:parse} give the selected JSON and its two code hooks. These are generated components of the same formal artifact, not teacher instructions.

\Needspace{8\baselineskip}
\begin{lstlisting}[style=prompt, caption={Archived student requests from formal InfoSeek replicate 1, protocol A. The first item uses a bounded candidate pick followed by a fresh answer context, while the second supplies the top card directly. The listing includes outputs and clearly separated offline scoring annotations.}, label={lst:runtime}]
ITEM infoseek_val_00000187 (test; image attached to every student call; interface v2: the router received only the question and the three CLIP candidates)
Question: What is the country of origin of this food?
Router decision: mode = discriminate, candidates = ["Pizza", "Marinara sauce", "Caprese salad"]

--- PASS 1: the bounded pick (student sees the image + this text) ---
Look at the image. Which option is the exact entity shown?
1. Pizza
2. Marinara sauce
3. Caprese salad
0. None of them / wrong kind
Answer with the number only.
Student output: '1'  -> chosen: 1

--- PASS 2: answer with the chosen card (fresh context; student sees the image + this text) ---
Use the image to identify the entity, then answer from the memory card if shown.
Question: 
What is the country of origin of this food?
Memory card:
- Entity: Pizza
Type: Italian baked flatbread dish.
Origin: Naples, Italy.
Main ingredients: Leavened wheat dough, tomato sauce, cheese, and toppings.
Classic style: Neapolitan pizza with tomatoes and mozzarella.
Cooking: Traditionally baked at high heat in a wood-fired oven.
Protected food: Pizza Napoletana has EU Traditional Speciality Guaranteed status.
Common toppings: Mozzarella, basil, pepperoni, vegetables, mushrooms, olives, meats, and seafood.
Global role: One of the world's most popular fast foods.
Answer only the requested value as a single word, number, or short phrase. For numeric questions, give one number, not a range or units.

Student output: 'Italy'
parse_py -> 'Italy'   (gold: 'Italy'; scored correct)

============================================================

ITEM infoseek_val_00000076 (test; image attached to every student call; interface v2: the router received only the question and the three CLIP candidates)
Question: What is the closest parent taxonomy of this bird?
Router decision: mode = answer, note = 'Taxonomy: answer the Genus field; if absent, use the first word of the scientific name.'

--- the single student call (image + this text) ---
Use the image to identify the entity, then answer from the memory card if shown.
Question: 
What is the closest parent taxonomy of this bird?
Memory card:
- Entity: Mandarin duck
Scientific name: Aix galericulata.
Family: Anatidae.
Genus: Aix (from scientific name).
Taxonomy: answer the Genus field; if absent, use the first word of the scientific name.
Answer only the requested value as a single word, number, or short phrase. For numeric questions, give one number, not a range or units.

Student output: 'Aix'
parse_py -> 'Aix'   (gold: 'Aix'; scored correct)
\end{lstlisting}

\Needspace{6\baselineskip}
\begin{lstlisting}[style=prompt, caption={Selected harness JSON from formal InfoSeek replicate 1, A ($H_{\mathrm{val}}$, round 1, test 37.3). The two code hooks are displayed in the following listings.}, label={lst:hval}]
{
 "name": "v2_typeguard_guided_parse",
 "rationale": "The report shows wrong linked cards are much worse than no card, especially when the question names a broad category such as plant, building, organization, or material but CLIP candidates are visibly the wrong kind; this version adds a conservative type-mismatch guard that falls back to no card. Several linked-correct failures came from the student answering an entity/type instead of the requested taxonomy/numeric/geometric value, so the router adds short per-question guidance and derives a Genus line from scientific names for taxonomy questions. Numeric outputs are also parsed more tightly when the student gives a single number with units.",
 "prompt": {
  "format": "plain",
  "system": null,
  "before_question": "Use the image to identify the entity, then answer from the memory card if shown.\nQuestion: ",
  "after_question": "",
  "card_frame": "Memory card:\n{cards}",
  "answer_suffix": "Answer only the requested value as a single word, number, or short phrase. For numeric questions, give one number, not a range or units."
 },
 "memory": {
  "k": 1,
  "sim_floor": 0.0,
  "max_card_chars": 1000,
  "store": "v2"
 },
 "decode": {
  "max_new_tokens": 32,
  "samples": 1,
  "temperature": 0.0,
  "top_p": 1.0,
  "aggregate": "vote"
 },
 "discriminate_prompt": "Look at the image. Which option is the exact entity shown?\n{options}\n0. None of them / wrong kind\nAnswer with the number only.",
 "name_prompt": "What is the specific name of the entity shown in the image? Answer with the name only.",
 "name_sim_floor": 0.5,
 "router_py": "<see Listing router>",
 "parse_py": "<see Listing parse>"
}
\end{lstlisting}

\Needspace{6\baselineskip}
\begin{lstlisting}[style=prompt, language=Python, caption={The same artifact's router. It uses question text and CLIP candidates to choose direct card input, a bounded pick, or no card. It filters card lines under a character budget, may derive a genus line, and attaches question-text hints. Appendix~\ref{app:rcontrols} tests these mechanisms separately.}, label={lst:router}]
import re

STOP = set('the a an of in on at to for from by with and or is are was were this that these those its his her their it entity shown image photo picture what which who whom whose when where how many much old tall long high deep wide large big small does do did has have had belong located made built created founded opened established'.split())

GROUPS = [
    (('family','familia'), ['family']),
    (('genus','genera'), ['genus']),
    (('order',), ['order']),
    (('class',), ['class']),
    (('phylum','division'), ['phylum','division']),
    (('kingdom',), ['kingdom']),
    (('taxonomy','taxon','parent','upper'), ['genus','scientific name','taxonomy','family']),
    (('species','binomial','scientific'), ['species','scientific name','binomial']),
    (('country','nation'), ['country','location','located','place']),
    (('city','town','municipality'), ['city','town','municipality','location','located']),
    (('state','province','region','county','district','prefecture'), ['state','province','region','county','district','prefecture','location']),
    (('where','located','location'), ['location','located','country','city','region','site','address']),
    (('native','endemic','range','distribution'), ['native','endemic','range','distribution']),
    (('elevation','altitude'), ['elevation','altitude']),
    (('height','tall'), ['height','tall']),
    (('length','long'), ['length','long']),
    (('width','wide'), ['width','wide']),
    (('area','size'), ['area','size']),
    (('depth','deep'), ['depth','deep']),
    (('diameter',), ['diameter']),
    (('weight','mass'), ['weight','mass']),
    (('speed','velocity'), ['speed','velocity']),
    (('density',), ['density']),
    (('decomposition','melting','boiling'), ['decomposition','melting','boiling','temperature']),
    (('population','inhabitants'), ['population','inhabitants']),
    (('capacity','seats'), ['capacity','seating','seats']),
    (('founded','established','formed'), ['founded','established','formed']),
    (('built','constructed','construction'), ['built','constructed','construction']),
    (('opened','inaugurated'), ['opened','inaugurated']),
    (('completed','finished'), ['completed','finished']),
    (('created','painted','written','composed','designed'), ['created','creator','artist','author','composer','designer','date']),
    (('born','birth'), ['born','birth']),
    (('died','death'), ['died','death']),
    (('discovered','discovery'), ['discovered','discovery']),
    (('dissolved','closed'), ['dissolved','closed']),
    (('architect','architecture'), ['architect','designed by']),
    (('artist','painter','sculptor'), ['artist','painter','sculptor','creator']),
    (('author','writer'), ['author','writer']),
    (('composer',), ['composer']),
    (('designer','designed'), ['designer','designed by']),
    (('manufacturer','maker','company','produced'), ['manufacturer','maker','company','produced by']),
    (('founder','founded by'), ['founder','founded by']),
    (('operator','operates','managed'), ['operator','operated by','managed by','manager']),
    (('diet','food','source','prey','eats'), ['diet','food','prey','feeds','source']),
    (('parent peak',), ['parent peak','peak','mountain']),
    (('crossed','crosses','over'), ['crosses','crossed','river','waterway']),
    (('material','made of'), ['material','made of']),
    (('style','architectural'), ['style','architectural style']),
    (('genre',), ['genre','type','category']),
    (('type','kind','category'), ['type','category','entity']),
    (('vertex figure','facet','face'), ['vertex figure','facet','face','side']),
    (('named after','etymology','name origin'), ['named after','etymology','name origin','name meaning']),
    (('religion','denomination'), ['religion','denomination']),
    (('currency',), ['currency']),
    (('language',), ['language']),
    (('capital',), ['capital']),
    (('isbn','license','iata','icao'), ['isbn','license','iata','icao'])
]

def q_tokens(q):
    toks = re.findall(r'[a-z0-9]+', q.lower())
    return [t for t in toks if len(t) > 2 and t not in STOP]

def wanted_terms(q):
    ql = q.lower()
    w = set()
    for keys, vals in GROUPS:
        if any(k in ql for k in keys):
            w.update(vals)
    if re.search(r'\bwhen\b|\byear\b|\bdate\b', ql):
        w.update(['date','year','founded','built','opened','created','completed','established','born','died','mentioned'])
    if re.search(r'\bwho\b|\bwhom\b|\bwhose\b', ql):
        w.update(['creator','architect','artist','author','designer','manufacturer','founder','composer','operator','by'])
    if re.search(r'\bhow\s+(many|much|tall|high|long|deep|wide|large|big|old|fast)\b', ql):
        w.update(['height','elevation','length','width','depth','area','population','capacity','age','built','opened','founded','weight','speed'])
    return w

def is_taxonomy_q(q):
    ql = q.lower()
    return bool(re.search(r'closest\s+(parent|upper)\s+taxonomy|upper\s+taxonomy|parent\s+taxonomy|\bgenus\b', ql))

def augment_lines(lines, question):
    out = list(lines)
    if is_taxonomy_q(question) and not any(re.match(r'(?i)\s*genus\s*:', ln) for ln in out):
        for ln in out:
            m = re.search(r'(?i)^\s*(?:scientific name|binomial)\s*:\s*([A-Z][A-Za-z-]+)\s+[a-z][A-Za-z-]+', ln)
            if m:
                out.append('Genus: ' + m.group(1) + ' (from scientific name).')
                break
    return out

def compress(card, question, limit=850):
    if not card:
        return ''
    raw_lines = [ln.strip() for ln in card.splitlines() if ln.strip()]
    if not raw_lines:
        return card[:limit]
    raw_lines = augment_lines(raw_lines, question)
    toks = q_tokens(question)
    want = wanted_terms(question)
    kept = []
    used = set()
    def add(i):
        if i not in used and 0 <= i < len(raw_lines):
            kept.append((i, raw_lines[i])); used.add(i)
    add(0)
    scored = []
    ql = question.lower()
    numeric_q = bool(re.search(r'\b(when|year|date|how|many|much|tall|high|long|deep|wide|elevation|height|length|area|population|capacity|weight|mass|speed|velocity|density|temperature)\b', ql))
    for i, ln in enumerate(raw_lines[1:], 1):
        l = ln.lower()
        label = l.split(':', 1)[0]
        s = 0
        for t in toks:
            if t in l:
                s += 1
                if t in label:
                    s += 2
        for term in want:
            if term in l or term in label:
                s += 5
        if re.search(r'\d', ln) and numeric_q:
            s += 2
        if is_taxonomy_q(question) and label in ('genus','scientific name','binomial'):
            s += 8
        if s > 0:
            scored.append((-s, i, ln))
    scored.sort()
    for _, i, _ in scored[:7]:
        add(i)
    if len(kept) == 1:
        for i in range(1, min(len(raw_lines), 8)):
            add(i)
    kept.sort()
    out = []
    total = 0
    for _, ln in kept:
        extra = len(ln) + (1 if out else 0)
        if total + extra > limit:
            break
        out.append(ln); total += extra
    return '\n'.join(out) if out else card[:limit]

def category_patterns(q):
    ql = q.lower()
    pats = []
    if re.search(r'\bplant\b', ql):
        pats += [r'\bplant\b', r'\btree\b', r'\bshrub\b', r'\bflower', r'\bgrass\b', r'\bbamboo\b', r'\bcactus\b', r'\borchid\b', r'botanical', r'\bflora\b']
    if re.search(r'\binsect\b', ql):
        pats += [r'\binsect\b', r'\bbeetle\b', r'\bdragonfly\b', r'\bbutterfly\b', r'\bmoth\b', r'\bwasp\b', r'\bbee\b']
    if re.search(r'\bbird\b', ql):
        pats += [r'\bbird\b', r'\bavian\b']
    if re.search(r'\banimal\b', ql):
        pats += [r'\banimal\b', r'\bmammal\b', r'\bbird\b', r'\bfish\b', r'\breptile\b', r'\bamphibian\b', r'\binsect\b', r'\bcanid\b', r'\bfelid\b', r'\bbovid\b']
    if re.search(r'\bbuilding\b', ql):
        pats += [r'\bbuilding\b', r'\bchurch\b', r'\bcathedral\b', r'\bmonument\b', r'\bmemorial\b', r'\bcastle\b', r'\bmosque\b', r'\btemple\b', r'\bsynagogue\b', r'\bpalace\b', r'\btower\b', r'\bmuseum\b', r'\bfortress\b']
    if re.search(r'\bbridge\b', ql):
        pats += [r'\bbridge\b']
    if re.search(r'\bmountain\b|\bpeak\b', ql):
        pats += [r'\bmountain\b', r'\bpeak\b', r'\bsummit\b', r'\bmassif\b', r'\brange\b']
    if re.search(r'\blake\b', ql):
        pats += [r'\blake\b', r'\breservoir\b']
    if re.search(r'\borganization\b|\bbusiness group\b', ql):
        pats += [r'\borganization\b', r'\bcompany\b', r'\bairline\b', r'\bbusiness\b', r'\bgroup\b', r'\binstitution\b', r'\buniversity\b', r'\bteam\b', r'\bclub\b']
    if re.search(r'\bfood\b', ql):
        pats += [r'\bfood\b', r'\bdish\b', r'\bcocktail\b', r'\bdrink\b', r'\bbeverage\b', r'\bcuisine\b']
    if re.search(r'\bmaterial\b', ql):
        pats += [r'\bmaterial\b', r'\bcompound\b', r'\bchemical\b', r'\bsubstance\b', r'\bmineral\b']
    return pats

def card_matches_category(c, pats):
    if not pats:
        return True
    hay = ((c.get('name') or '') + '\n' + (c.get('card') or '')).lower()
    return any(re.search(p, hay) for p in pats)

def guidance(q):
    ql = q.lower()
    notes = []
    if is_taxonomy_q(q):
        notes.append('Taxonomy: answer the Genus field; if absent, use the first word of the scientific name.')
    if re.search(r'\b(when|year|date|how|many|much|tall|high|long|deep|wide|elevation|height|length|area|population|capacity|weight|mass|speed|velocity|density|temperature)\b', ql):
        notes.append('Number: give one requested number only, not a range or units.')
    if 'named after' in ql:
        notes.append('Named-after: answer the eponym/root name, not the whole entity.')
    if 'vertex figure' in ql or 'facet' in ql:
        notes.append('Answer the requested part or figure, not the object name.')
    return ' '.join(notes)

def select(item):
    clips = item.get('clip') or []
    q = item.get('question') or ''
    if not clips:
        return {'mode': 'answer', 'cards': [], 'note': guidance(q)}
    s1 = float(clips[0].get('sim') or 0.0)
    s2 = float(clips[1].get('sim') or 0.0) if len(clips) > 1 else -1.0
    gap12 = s1 - s2
    pats = category_patterns(q)
    any_type_match = any(card_matches_category(c, pats) for c in clips[:3]) if pats else True
    if pats and not any_type_match and s1 < 0.335:
        return {'mode': 'answer', 'cards': [], 'entity_ids': [], 'note': guidance(q)}
    confident = (len(clips) == 1) or (s1 >= 0.38) or (gap12 >= 0.045) or (s1 >= 0.31 and gap12 >= 0.022)
    if confident:
        return {
            'mode': 'answer',
            'cards': [compress(clips[0].get('card') or '', q, 850)],
            'entity_ids': [clips[0].get('entity_id')],
            'note': guidance(q)
        }
    names = []
    for c in clips[:3]:
        nm = c.get('name') or ''
        if nm:
            names.append(nm)
    return {'mode': 'discriminate', 'candidates': names, 'note': guidance(q)}
\end{lstlisting}

\Needspace{6\baselineskip}
\begin{lstlisting}[style=prompt, caption={Compile request of formal InfoSeek replicate 1 (the call that produced the initial harness), rendered from the frozen inputs with the launch-revision code. Its status is \emph{reconstructed from frozen inputs}, not an archived payload. The runtime manual is the first block.}, label={lst:compile_req}]
# RECONSTRUCTED FROM FROZEN INPUTS (not an archived payload): the single user message of the compile call of formal InfoSeek
# replicate 101 (run eval_results/p7_handbook2/formal/infoseek/rep101/H0), rendered by hd_teacher.compile_prompt at code revision 85b260c
# (prompt modules byte-identical to the launch revision) from the frozen inputs; no system message; model gpt-5.5, max_completion_tokens 16000,
# no reasoning_effort set. Transformations for print: non-ASCII folded to ASCII. Source sha256 5e01b55c63431894...

You are compiling an INFERENCE-TIME HARNESS for a small frozen vision-language model (the student). The harness is a JSON object; the runtime executes it. The student is called with an image and a text prompt and must answer a knowledge question about the entity in the image (InfoSeek: e.g. "What is the elevation of this mountain?", "Which family does this plant belong to?"). Answers are short strings or numbers; scoring is the official InfoSeek relaxed exact match (numbers must fall in the gold range).

At inference NO large model is available. The only deterministic signals the runtime can compute per item are:
- clip: the top-3 entity names from a 1,380-entity store (the gold entity is always in the store), each with a CLIP ViT-B-32 image->name cosine similarity (typically 0.2-0.4; top-1 is the gold entity about 52% of the time, top-3 contains it 68%), and that entity's memory card text.
- The question text.
Cards: the store's current cards are mostly just the entity NAME (median 17 characters); only 138 of 1,380 are full fact cards. You may request a regenerated store ("memory.store": "v2", written by you offline with your own instructions; its cost is charged to this build) - see "rebuild_store" below.

HARNESS JSON (all keys optional; defaults shown):
{
 "name": "short id", "rationale": "why this harness, 1-3 sentences",
 "prompt": {
   "format": "plain",                  # or "typed_fields": the DSPy-adapter serialisation (system message declaring typed input/output fields, [[ ## field ## ]] markers, answer field + completed marker; answer_suffix becomes the objective sentence; the output is parsed from the answer field unless parse_py is set). It is a fixed contract, not an optimiser.
   "system": null,                      # a system message
   "before_question": "",               # text placed before the question
   "after_question": "",                # text placed after the question / cards
   "card_frame": "Retrieved from your memory store (may or may not describe the right entity):\n{cards}",   # must contain {cards}
   "answer_suffix": "Answer the question with a single word, number, or short phrase."
 },
 "memory": {"k": 1, "sim_floor": 0.0, "max_card_chars": 600, "store": "v1"},   # k in 0..3 = how many CLIP top cards are shown by default (subject to the floor)
 "decode": {"max_new_tokens": 64},
 "discriminate_prompt": "Which of these is the entity shown in the image?\n{options}\n0. None of them\nAnswer with the number only.",   # used when a router returns mode "discriminate" (student picks 1..n or 0 = no card); must contain {options}
 "name_prompt": "What is the specific name of the entity ... Answer with the name only.",   # used when a router returns mode "name" (student names the entity; MiniLM links the name into the store)
 "name_sim_floor": 0.5,
 "router_py": null,    # OPTIONAL python source defining  def select(item) -> dict  ; runs in a sandbox (stdlib only, no network, no files)
 "parse_py": null      # OPTIONAL python source defining  def parse(text) -> str   ; EXTRACTION ONLY: it must return a contiguous span of the student's raw output (after whitespace/markdown normalisation); a value that is not such a span is replaced by the default extractor for that item and the count is reported to you; no unit conversion, no interval-to-number, no rewriting of the answer (the official scorer does its own normalisation)
}

router select(item) receives item = {"uid", "question", "clip": [{"name","sim","card","entity_id"}, x3 sorted by sim]} (anything about the question's type must be computed from its text) and returns
{"mode": "answer" | "discriminate" | "name",
 "cards": [card strings to show, <=3],      # for mode answer; you may rewrite/compress a card text (e.g. keep only lines relevant to the question)
 "entity_ids": [ids of the entities those cards came from, same order],   # optional; lets the round report attribute outcomes to the right card
 "candidates": [names, <=3],                # for mode discriminate (their cards are looked up by name)
 "note": "optional extra text shown after the cards"}
Degrees of freedom left to the student: mode answer = 0 decisions; discriminate = 1 (choose among <=3); name = 1 (generate a name). If a student profile is given below, it tells you which of these this student can execute.

rebuild_store: to regenerate the memory cards, add  "rebuild_store": {"instructions": "<how each card should be written; e.g. what facts to prioritise, length limit>"}  and set "memory.store": "v2". The runtime will have YOU (the teacher, offline) write one card per entity from the entity name alone, following your instructions; the student never sees anything else. Do this only if you believe richer cards help this student (the profile reports how much injected text it can absorb). The store can be rebuilt ONCE per build (the instructions you give first are the ones that count; later rebuild requests are ignored and the report will say so).

Constraints: Control code must generalise: it may not name specific entities, items, or practice answers (a string literal equal to a store entity name is rejected and you will be asked again); a rule supported by fewer than about ten probe items is treated as unsupported. The harness must run with zero large-model calls at inference; keep prompts short (the student is sensitive to extra text); subtractive edits (k=0, no system text, fewer modes) are welcome and the bare-student score is always reported so you can see whether the harness is helping at all.

Reply with ONE JSON object for the harness at the top level (the keys listed above directly in the object - do not wrap it in "harness"/"spec"; unknown keys are ignored). You may precede it with brief reasoning. Python hook sources must be JSON string values (escape newlines as \n).

## Task
InfoSeek (knowledge-intensive visual question answering). Each item is a photo of an entity (landmark, building, species, artwork, vehicle, food...) and a question about a fact of that entity (dates, dimensions, locations, taxonomy, counts, materials...). The student must identify the entity and recall or read the fact. Scoring: official InfoSeek relaxed exact match; numeric answers must fall within the gold range. The practice split has labels; the teacher (you) sees practice traces only. The test split is evaluated once at the end and is never shown to you.

## Student profile (measured on practice data; feed into your design)
(no profile measured)

Full profile JSON:
{
 "note": "no student profile available for this build"
}

## Store examples (3 of 1,380 cards)
- Laurelia novae-zelandiae: Laurelia novae-zelandiae is an endemic New Zealand evergreen forest tree, commonly called pukatea.
Native to North, South and Stewart Islands.
Grows up to about 35 m tall.
Trunks can exceed 1 m in diameter and are often buttressed.
Common in wet lowland forest and swamp margins.
Family: Atherospermataceae.
- Byland Abbey: Byland Abbey is a ruined Cistercian monastery in North Yorkshire, England.
Founded in 1135.
Established at its final site near Coxwold in 1177.
Dissolved under Henry VIII in 1538.
Its abbey church was about 100 m long.
Managed today by English Heritage.
- Via dei Fori Imperiali: Via dei Fori Imperiali is a major road in central Rome, Italy.
Runs between Piazza Venezia and the Colosseum.
About 850 m long.
Built in 1924-1932 under Benito Mussolini.
Opened as Via dell'Impero on 9 April 1932.
Crosses the archaeological area of the Imperial Fora.

## Bare student accuracy on the 600-item probe set (no harness): 19.33

Write the round-0 harness now.

NOTE: a regenerated store 'v2' ALREADY EXISTS for this build (1,380 fact cards written earlier by GPT-5.5 with these instructions: "Write one compact encyclopedic fact card for the given entity name. Length limit: 900 characters. Use plain text, one fact per line, with stable labels when possible. First line must be `Entity: <canonical name>`. Include aliases/common names if important. Prioritize facts commonly asked in InfoSeek: entity type/category; country/city/region/location; dates/years founded, built, opened, created, born/died, discovered, dissolved; creators such as architect, artist, author, designer, manufacturer, founder; exact numeric facts with units such as elevation, height, length, width, area, depth, diam"). Set "memory.store": "v2" to use it; a rebuild_store request will be ignored (the existing v2 is kept).
\end{lstlisting}

\Needspace{6\baselineskip}
\begin{lstlisting}[style=prompt, caption={Round-1 reflection request under protocol A of the same replicate (the call that produced the selected harness), rendered from the archived round-0 report and history. The manual, the parent's two code hooks and 28 of the 30 failure traces are elided as marked.}, label={lst:reflect_req}]
# RECONSTRUCTED FROM FROZEN INPUTS (not an archived payload): the single user message of the round-1 reflection call under protocol A of formal
# InfoSeek replicate 101 (run eval_results/p7_handbook2/formal/infoseek/rep101/A), rendered by hd_teacher.reflect_prompt at code revision 85b260c
# from the frozen inputs (state.json history, the archived report_r0.md, harness_r0.json, bare probe score); same model and decoding as the compile call.
# Print transformations: the leading runtime manual (identical to the compile request) is replaced by one line; the parent's two code hooks and
# 28 of the 30 failure traces are elided as marked; non-ASCII folded. Source sha256 06c4f7145e3f2d96... (32546 chars).

[RUNTIME MANUAL: identical to the first block of the compile request, omitted]

You are in round 1 of the harness-distillation loop. History so far (bare student = 19.33 on the probe set; best validated harness so far = 'v2_confident_top1_else_discrim' at validation 40.0):
- round 0: 'v2_confident_top1_else_discrim' probe 36.67 | validation 40.0 | ACCEPTED (new best)

## The harness that produced this report (edit THIS object; keep the exact schema and key names)
{
 "name": "v2_confident_top1_else_discrim",
 "rationale": "Use the existing regenerated fact-card store because InfoSeek mostly requires entity-specific facts that the frozen student is unlikely to know. Trust CLIP top-1 only when it is comparatively confident; otherwise ask the student to visually discriminate among the top CLIP candidates, then answer from the selected card. Cards are question-compressed for confident top-1 cases to keep prompts short.",
 "prompt": {
  "format": "plain",
  "system": null,
  "before_question": "Use the image to identify the entity, then answer from the memory card.\nQuestion: ",
  "after_question": "",
  "card_frame": "Memory card:\n{cards}",
  "answer_suffix": "Answer with a single word, number, or short phrase."
 },
 "memory": {
  "k": 1,
  "sim_floor": 0.0,
  "max_card_chars": 1000,
  "store": "v2"
 },
 "decode": {
  "max_new_tokens": 32,
  "samples": 1,
  "temperature": 0.0,
  "top_p": 1.0,
  "aggregate": "vote"
 },
 "discriminate_prompt": "Look at the image. Which option is the exact entity shown?\n{options}\n0. None of them\nAnswer with the number only.",
 "name_prompt": "What is the specific name of the entity shown in the image? Answer with the name only.",
 "name_sim_floor": 0.5,
 "router_py": "[... router_py source, 141 lines, omitted; identical to the archived harness_r0.json ...]",
 "parse_py": "import re\n\ndef parse(text):\n    if text is None:\n        return ''\n    s = str(text).strip()\n    if not s:\n        return s\n    m = re.search(r'(?is)\\b(?:answer|ans)\\s*[:\\-]\\s*(.+)', s)\n    if m:\n        s = m.group(1).strip()\n    s = s.split('\\n')[0].strip()\n    s = re.sub(r'^[-*\u2022\\s]+', '', s).strip()\n    s = s.strip('`*_\"\\' ')\n    return s\n"
}

Below is the report of the LAST round's harness on the 600-item probe set, with raw student traces.

### Round result for harness 'v2_confident_top1_else_discrim'
Probe accuracy: 36.67 (n=150); bare student on the same items: 19.33
Fired (card shown): 87.3% | unparsed: 0.0% | mean output chars: 11.4 | student seconds: 19.6
Accuracy by DoF condition (linked = whether the card shown was the gold entity's; computed from practice labels, never shown to the student): {"correct": {"acc": 46.94, "n": 98}, "none": {"acc": 31.58, "n": 19}, "wrong": {"acc": 9.09, "n": 33}}
By split/qtype: {"unseen_entity/num": {"acc": 25.0, "n": 16}, "unseen_entity/str": {"acc": 44.07, "n": 59}, "unseen_question/num": {"acc": 37.5, "n": 16}, "unseen_question/str": {"acc": 32.2, "n": 59}}
By mode: {"answer": {"acc": 36.0, "n": 50}, "discriminate": {"acc": 37.0, "n": 100}}
Discriminate pass: {"n": 100, "picked_none": 19.0, "linked_correct_when_picked": 65.4}

#### 30 failure traces (stratified by linked condition; 95 failures total)
- uid=infoseek_val_00066064 split=unseen_entity qtype=num mode=answer linked=correct correct=0
  Q: What is the weight of a female of this animal in kilogram?
  gold: 32.5
  CLIP top-3 (name, sim): [('Sea otter', 0.315), ('Beaver', 0.284), ("Hoffmann's two-toed sloth", 0.279)]
  cards shown: ['Entity: Sea otter\nWeight: About 14-45 kg (31-99 lb).\nNoted for: Densest fur of any animal; often uses rocks as tools.']
  student output: '14-45 kg' -> parsed: '14-45 kg'

- uid=infoseek_val_00072264 split=unseen_entity qtype=str mode=discriminate linked=correct correct=0
  Q: What is the vertex figure of this object?
  gold: line segment
  CLIP top-3 (name, sim): [('Hexagon', 0.282), ('Jigsaw puzzle', 0.267), ('Cornettes de Bise', 0.251)]
  cards shown: ['Entity: Hexagon\nType: Polygon in Euclidean geometry.\nDefinition: Six-sided polygon with six vertices and six interior angles.\nRegular hexagon: All sides and ang']
  candidates: ['Hexagon', 'Jigsaw puzzle', 'Cornettes de Bise'] chosen=1 pass1='1'
  student output: 'Hexagon' -> parsed: 'Hexagon'

[... 28 further traces of this report omitted; the archived report is report_r0.md in the run directory ...]

Decide the next harness. You may keep, edit, or simplify the current one; removing components is as legitimate as adding them. If you first want to see more traces, reply ONLY with {"query": {...}} using keys among correct (0/1), linked (none/correct/wrong), qtype (num/str), split, mode, sim_min, sim_max, contains - you get one query per round, up to 30 traces. Otherwise reply with the complete new harness JSON (the full object, not a diff), starting with a 'rationale' that names what in the report motivated each change.

NOTE: a regenerated store 'v2' ALREADY EXISTS for this build (1,380 fact cards written earlier by GPT-5.5 with these instructions: "Write one compact encyclopedic fact card for the given entity name. Length limit: 900 characters. Use plain text, one fact per line, with stable labels when possible. First line must be `Entity: <canonical name>`. Include aliases/common names if important. Prioritize facts commonly asked in InfoSeek: entity type/category; country/city/region/location; dates/years founded, built, opened, created, born/died, discovered, dissolved; creators such as architect, artist, author, designer, manufacturer, founder; exact numeric facts with units such as elevation, height, length, width, area, depth, diam"). Set "memory.store": "v2" to use it; a rebuild_store request will be ignored (the existing v2 is kept).
\end{lstlisting}

\Needspace{6\baselineskip}
\begin{lstlisting}[style=prompt, caption={Archived teacher reply to Listing~\ref{lst:reflect_req}: the rationale field. The remaining fields are Listings~\ref{lst:hval}--\ref{lst:parse}.}, label={lst:reflect_reply}]
# ARCHIVED TEACHER REPLY (verbatim excerpt): run rep101/A, teacher_reflect_r1_0.txt; the complete reply parses to the selected round-1 harness
# (content hash 5c8837f3e04a; Listings hval/router/parse). Reply sha256 ba6b2c7dab9f4c7c... Shown: the rationale field; the rest is the JSON of the listings.

"rationale": "The report shows wrong linked cards are much worse than no card, especially when the question names a broad category such as plant, building, organization, or material but CLIP candidates are visibly the wrong kind; this version adds a conservative type-mismatch guard that falls back to no card. Several linked-correct failures came from the student answering an entity/type instead of the requested taxonomy/numeric/geometric value, so the router adds short per-question guidance and derives a Genus line from scientific names for taxonomy questions. Numeric outputs are also parsed more tightly when the student gives a single number with units."
[... remaining JSON fields = Listing hval; router_py = Listing router; parse_py = Listing parse ...]
\end{lstlisting}

\Needspace{6\baselineskip}
\begin{lstlisting}[style=prompt, language=Python, caption={The same artifact's extraction-only parser. It removes answer markers and selects text without generating a new numeric value. The contract-versus-extractor experiment in Appendix~\ref{app:rcontrols} covers this artifact: replaying the default extractor on its archived outputs changes no test score.}, label={lst:parse}]
import re

def parse(text):
    if text is None:
        return ''
    s = str(text).strip()
    if not s:
        return s
    m = re.search(r'(?is)\b(?:answer|ans)\s*[:\-]\s*(.+)', s)
    if m:
        s = m.group(1).strip()
    s = s.split('\n')[0].strip()
    s = re.sub(r'^[-*\u2022\s]+', '', s).strip()
    s = s.strip('`*_"\' ')
    num_pat = r'[-+]?\d{1,3}(?:,\d{3})*(?:\.\d+)?|[-+]?\d+(?:\.\d+)?'
    nums = list(re.finditer(num_pat, s))
    if len(nums) == 1:
        before = s[:nums[0].start()].strip().lower()
        after = s[nums[0].end():].strip().lower()
        allowed = set('about approx approximately around circa ca c kg kilogram kilograms g gram grams mg lb lbs pound pounds m meter meters metre metres km kilometer kilometers kilometre kilometres cm centimeter centimeters centimetre centimetres mm ft foot feet mi mile miles square cubic sq population people inhabitants seats percent percentage degrees degree celsius c ?c'.split())
        def ok_side(x):
            if not x:
                return True
            x = x.replace('.', ' ').replace('?', ' ?')
            toks = re.findall(r'[a-z?]+', x)
            return toks and all(t in allowed for t in toks)
        if ok_side(before) and ok_side(after) and not re.search(r'[---]\s*' + num_pat, s):
            return nums[0].group(0)
    return s
\end{lstlisting}

\subsection{What the formal routers do, and frozen-artifact controls}
\label{app:rcontrols}
\paragraph{Mechanism inventory.} The six formal InfoSeek artifacts (three builds and two protocols, Table~\ref{tab:manifest_formal}) were inspected before any control was chosen (Table~\ref{tab:rinv}). None inserts a \texttt{TARGET ANSWER} value line. Three derive a genus line from the scientific name on taxonomy questions, three attach question-type notes, all six rewrite the retrieved card by keyword-scored line selection under a character budget, and routing ranges from similarity thresholds with a category abort (replicate 1, A) through similarity-only routing to no routing at all (replicate 3: the top three cards are always shown). A control that targets a mechanism the artifact does not have is reported as not applicable rather than simulated.

\begin{table}[!htb]
\centering\footnotesize\setlength{\tabcolsep}{4pt}\renewcommand{\arraystretch}{1.12}
\caption{\textbf{Mechanisms present in the formal InfoSeek routers.} All six use keyword-scored card-line selection. The budget is characters per card. A = reported reflection. B = incumbent-anchored variant. ``Genus'' means a line derived from the scientific name, and ``hints'' means question-type notes.}
\label{tab:rinv}
\begin{tabularx}{\linewidth}{@{}llccrY@{}}
\toprule
Replicate & Protocol & Genus & Hints & Card budget & Routing \\
\midrule
1 & A & Yes & Yes & 850 & Similarity thresholds + category abort \\
1 & B & No & No & 850 & Similarity thresholds only \\
2 & A & No & Yes & 850 & Similarity thresholds only \\
2 & B & No & No & 850 & Similarity thresholds only \\
3 & A & Yes & Yes & 560 & Always show the top three cards \\
3 & B & Yes & No & 560 & Always show the top three cards \\
\bottomrule
\end{tabularx}
\end{table}

\paragraph{Controls.} Each control edits the frozen artifact by replacing one named function (the appended override lines are released) and re-runs the student on the 6,000 test items. Everything else, including the store, the prompts and the parse hook, is byte-identical. R0 replays the artifact unchanged. R1 removes the derived line, R2 the question-type notes, and R3 both. R4 shows the stored card's original lines in original order under the same character budget. R5 disables the category abort so that routing depends on similarity alone. Entries are paired differences against R0 on the same items with fixed/broken counts ($^*$: exact McNemar $p<0.05$). Removing a component from a selected artifact is not the same as compiling without it.

\begin{table}[htbp]
\centering\scriptsize\setlength{\tabcolsep}{3pt}
\caption{Frozen-artifact controls on the formal InfoSeek artifacts (6,000 test items, paired against the R0 replay). R0 aggregate scores are within 0.1 point of the archived scores. The small replay differences are shown explicitly.}
\label{tab:rcontrols}
\begin{tabular}{@{}llrrrrrr@{}}
\toprule
Replicate & Protocol & archived / R0 & R1 derived line & R2 hints & R3 both & R4 original cards & R5 similarity routing \\
\midrule
1 & A & 37.30 / 37.30 & $+0.00$ (0/0) & $-2.23$ (37/171$^*$) & $-2.38$ (37/180$^*$) & $+0.72$ (65/22$^*$) & $+0.68$ (66/25$^*$) \\
1 & B & 36.12 / 36.13 & n/a & n/a & n/a & $+0.75$ (77/32$^*$) & n/a \\
2 & A & 39.13 / 39.12 & n/a & $-0.13$ (6/14) & n/a & $+0.27$ (21/5$^*$) & n/a \\
2 & B & 36.93 / 36.93 & n/a & n/a & n/a & $+0.37$ (28/6$^*$) & n/a \\
3 & A & 36.62 / 36.60 & $-0.10$ (11/17) & $-1.18$ (12/83$^*$) & $-1.55$ (11/104$^*$) & $+2.87$ (315/143$^*$) & n/a \\
3 & B & 34.63 / 34.72 & $-0.27$ (2/18$^*$) & n/a & n/a & $+1.67$ (299/199$^*$) & n/a \\
\bottomrule
\end{tabular}
\end{table}

\paragraph{Reading.} The keyword-scored card rewrite, which every formal router implements, costs between 0.3 and 2.9 points on test on every one of the six artifacts (R4 is positive and significant everywhere, with the largest effect on replicate 3, whose routers compress each of three cards to 560 characters): the teacher's line selection drops lines the student needed more often than it removes distractors. The category abort of replicate 1 (A) likewise costs 0.7 points. The derived genus line changes at most 0.3 points. The question-type notes are the one mechanism that helps: removing them costs 2.2 points on replicate 1 (A) and 1.2 on replicate 3 (A), and nothing measurable on replicate 2 (A), whose notes are shorter. These are selection gaps of validation-chosen artifacts on held-out items, not evidence that the mechanisms could not be useful. The inference is limited to these frozen artifacts and their named interventions.

\paragraph{Output contract and extractor.} For the six formal artifacts (three per protocol) the contract sentences of the prompt (those instructing the output form: ``answer only'', ``single word, number or short phrase'' and the like, selected by a frozen rule released with the code) are separated from the extraction hook (Table~\ref{tab:ocontrols}): O11 is the archived run, O10 replays the fixed default extractor on the archived raw outputs, O01 removes the contract sentences and re-generates with the artifact's hook, O00 applies the default extractor to those outputs. The rule requires an output-form sentence to mention the answer or reply and an output restriction. Frozen before any A evaluation, it selects the same sentences on the B artifacts as the earlier rule. Replaying the default extractor changes the score of no artifact by more than 0.1 points while the contract is present, so the extraction hooks do no semantic work. Removing the contract sentences costs 1.8 to 6.1 points on the A artifacts with the artifact's hook (O01) and 1.8 to 4.6 with the default extractor (O00), and 1.7 to 2.6 / 1.3 to 3.2 on the B artifacts: the student's outputs become longer and the extractor then matters by up to 1.5 points. The contract accounts for the measured output-interface benefit on both protocols. This does not directly decompose the historical 4.3-point intervention on the diagnostic harness in Section~\ref{sec:carry}.

\begin{table}[htbp]
\centering\scriptsize\setlength{\tabcolsep}{4pt}
\caption{Contract-versus-extractor decomposition on the six formal InfoSeek artifacts (test accuracy, with paired differences against O11 in parentheses). Every paired difference of 1.7 points or more is significant at $p<10^{-5}$.}
\label{tab:ocontrols}
\begin{tabular}{@{}llrrrr@{}}
\toprule
Replicate & Protocol & O11 selected / hook & O10 selected / default & O01 removed / hook & O00 removed / default \\
\midrule
1 & A & 37.30 & 37.30 ($+0.00$) & 35.52 ($-1.78$) & 35.52 ($-1.78$) \\
2 & A & 39.13 & 39.12 ($-0.02$) & 34.38 ($-4.75$) & 35.15 ($-3.98$) \\
3 & A & 36.62 & 36.62 ($+0.00$) & 30.55 ($-6.07$) & 32.03 ($-4.58$) \\
1 & B & 36.12 & 36.12 ($+0.00$) & 34.05 ($-2.07$) & 32.88 ($-3.23$) \\
2 & B & 36.93 & 36.87 ($-0.07$) & 34.37 ($-2.57$) & 33.90 ($-3.03$) \\
3 & B & 34.63 & 34.63 ($+0.00$) & 32.92 ($-1.72$) & 33.38 ($-1.25$) \\
\bottomrule
\end{tabular}
\end{table}

\FloatBarrier

\section{Compiler ablations and build stability}
\label{app:compiler}
\subsection{Teacher, profile and feedback ablations}
\label{app:ablate}
\begin{table}[htbp]
\centering\small\setlength{\tabcolsep}{5pt}\renewcommand{\arraystretch}{1.15}
\caption{Compiler-input ablations (historical: earlier runtime and manual, budgets in Appendix~\ref{app:ladder}). The matched score-only comparison under the current protocol is in the text.}
\label{tab:ablate}
\begin{tabularx}{\linewidth}{@{}Yrr@{}}
\toprule
Configuration & InfoSeek & DocVQA \\
\midrule
DSL default: name card / plain OCR & 23.6 & 80.1 \\
Store-only / OCR-only control & 33.5 & 80.1 \\
Sensitivity profile (3 / 2 builds) & 33.7 to 34.0 & 84.9 / 85.7 \\
Profile withheld (3 / 1 builds) & 35.6 / 39.7 / 40.3 & 86.0 \\
Measurements-only profile & 34.0 & n/a \\
Decision-menu profile & 38.8 & n/a \\
Score-only report & 30.2 & 72.9 \\
Selected harness with router removed & n/a & 85.1 / 86.4 \\
Prompt-only loop & 18.7 & 80.4 \\
Student as teacher, text-only & 23.6 & 85.0 \\
gpt-5.4-mini / local 30B teacher & 22.5 / 23.5 & 85.3 / 85.1 \\
\bottomrule
\end{tabularx}
\end{table}

\paragraph{Traces diagnose failures that scores alone do not.} In the earlier ablation, a score-only report reached 30.2 on InfoSeek and 72.9 on DocVQA. The latter build removed OCR after a parse failure it could not diagnose. A matched comparison under the current protocol starts from the formal replicate-1 initial harness with the same gate and changes only the report: the score-only arm selects 37.0 on InfoSeek (traced run: 37.3) and 74.4 on DocVQA (traced: 85.8), where it keeps a round-1 proposal that the traced loop had surpassed by repairing the output path. Traced feedback repairs broken parse hooks and identifies harmful components, although on InfoSeek the two reports led to nearly the same artifact. These ablations support using execution evidence, while the one-shot results below show that reflection is not necessary on every task. The one-shot comparator also removes guidance from its manual, so its difference from HC does not isolate iteration. The same-build $H_0$-to-selected differences are the direct evidence of what the loop's trajectory added, and a rationale in a reflect reply is a hypothesis about the failure, not a causal attribution of the score change to the named component.

\paragraph{Profile content matters.} Sensitivity profiles produce InfoSeek builds at 33.7--34.0 that retain no student selection. No-profile builds keep a bounded pick, while a decision-menu profile also retains it (38.8, from 38.15 at round 0). The comparisons involve a small number of builds and different budgets. They support a distinction between profile formats, not a universal benefit from withholding measurements. The earlier manual included two facts from the Qwen3-VL-4B profile battery in every condition. ``Profile withheld'' removes the measured battery and its summary but leaves those facts, so the comparison does not isolate access to all profile information.

\paragraph{Teacher capacity is task-dependent.} On DocVQA, GPT-5.5, gpt-5.4-mini, the local 30B teacher and the student itself reach similar scores. On InfoSeek, the smaller teachers do not elect to regenerate the fact store and remain near the name-card default. Teacher-written content and feedback-driven control are therefore separate sources of improvement.

\subsection{One-shot construction}
\label{app:oneshot}
As a second reference, the teacher writes a harness in one GPT-5.5 call per setting (USD 1.44 across the original eight-setting batch) from the runtime's manual with every guidance sentence removed by a fixed rule (references to the student profile, degrees of freedom, calibration facts, and ``prefer'' / ``should'' sentences, totalling 2--8 logged removals per setting), plus the task description. The runtime validates the harness, which is then tested once. Tab.~\ref{tab:oneshot} pairs it against the compiled harness and the human recipe: performance varies with the interface and the decisions that must be fitted to the student.
\begin{table}[htbp]
\centering\footnotesize\setlength{\tabcolsep}{5pt}\renewcommand{\arraystretch}{1.15}
\caption{Independent one-shot construction versus iterative HC (historical: one teacher call under the earlier manual with guidance removed, paired against the earlier single HC builds listed in Table~\ref{tab:build_registry}, without a rerun under the current runtime). Average weights the seven settings equally.}
\label{tab:oneshot}
\begin{tabularx}{\linewidth}{@{}Yrrl@{}}
\toprule
Setting & One-shot & HC & One-shot minus HC \\
\midrule
BLINK reflectance & 65.7 & 64.9 & $+0.7$ [$-7.5$, $+9.0$], n.s. \\
BLINK correspondence & 62.8 & 67.6 & $-4.8$ [$-9.1$, $-0.7$], $p=.035$ \\
BLINK depth & 95.2 & 95.2 & Identical verdicts \\
DocVQA & 85.5 & 84.9 & $+0.6$ [$-0.3$, $+1.4$], n.s. \\
InfoSeek & 25.0 & 39.7 & $-14.7$ [$-15.8$, $-13.5$] \\
LiveVQA & 30.6 & 37.1 & $-6.5$, $p=4\times10^{-14}$ \\
SlideVQA & 48.3 & 56.7 & $-8.4$ [$-10.2$, $-6.5$] \\
\midrule
Average & 59.0 & 63.7 & $-4.7$ \\
\bottomrule
\end{tabularx}
\end{table}
One call matches the selected build on depth and is statistically tied on reflectance and DocVQA, but falls below the compared loop on InfoSeek, correspondence, LiveVQA and SlideVQA. In particular, the InfoSeek one-shot result (25.0) is below the public-content human recipe (33.8). The manual supplies tool descriptions and an editable interface, but the changed guidance and independent initial draw prevent this comparison from isolating iteration.

\subsection{Build variation and validation selection}
\label{app:builds}
\begin{figure}[!htb]
    \centering
    \includegraphics[width=\linewidth]{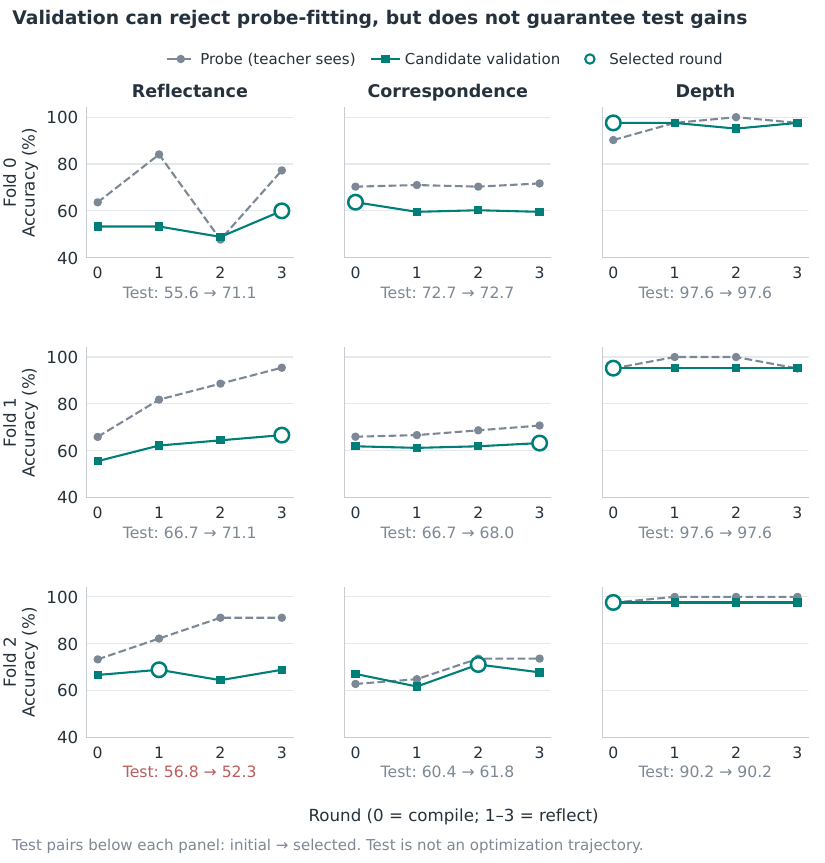}
    \caption{\textbf{Probe improvements can fail to generalize.} Seed-0 BLINK trajectories, with tasks in columns and folds in rows. Dashed lines show teacher-visible probe scores, solid lines candidate validation scores, and rings gate-selected rounds. Held-out test scores are printed below each panel as initial $\to$ selected, not used for selection. Reflectance fold 2 loses test accuracy despite its gate improvement. These are the archived per-fold trajectories replotted at readable scale.}
    \label{fig:loop}
\end{figure}

The formal initial-to-selected comparisons are reported in Appendix~\ref{app:formal}. Validation selection can still choose a proposal that performs worse on held-out items. Figure~\ref{fig:loop} illustrates this with per-fold BLINK trajectories: probe, validation and test play different roles, and only validation chooses the retained round.

\subsection{Initial-build safeguard and execution feedback}
\label{app:brokenround}
\label{app:feedback}
If round 0 scores below bare on the probe, the teacher receives a traced diagnosis and may recompile it at most twice before validation selection. Bare also remains a validation candidate. The safeguard prevents a below-bare candidate from becoming the only available reference but does not guarantee improvement on unseen data.

Execution reports identify the stage at which an item fails, including linking, candidate selection, evidence use and output extraction. The teacher can revise the corresponding component, but several components may change together. A reflection rationale is therefore a proposed explanation, not a causal attribution. The formal initial-to-selected comparisons measure the resulting build-level changes, while the frozen-artifact interventions test individual mechanisms (Appendices~\ref{app:formal} and \ref{app:rcontrols}).

\subsection{Formal replicates, the frozen gate and the reflection-protocol comparison}
\label{app:formal}
\paragraph{Data roles.} Content-task builds use disjoint practice subsets: a probe set $P$ (150 items) whose traces the teacher reads and a gate set $G$ (300) that selects among proposals. On InfoSeek and DocVQA, an additional practice selection set $S$ (300) was used once to choose the gate rule, for 750 unique development items per task rather than 450. The official test set $T$ is separate from the practice pool. Neither $S$ nor $T$ is supplied to the teacher. The formal runs reuse the benchmark test splits examined in earlier experiments. They measure repeated builds on those splits rather than a new prospective test population. BLINK retains its three cross-build folds.

\paragraph{Protocols and gate selection.} All formal builds share runtime interface v2 and a common round-0 harness $H_0$ per replicate. A edits the last proposal, while B edits the incumbent (Appendix~\ref{app:reflectv2}). Both use strict improvement of the unrounded mean on $G$, with the bare floor. Before the formal builds, five development builds per procedure on each of InfoSeek and DocVQA compared A, B, and B with a one-point acceptance margin, using only $S$ for this choice. The predeclared adoption rule for the margin required a mean gain of at least one point over B across the two tasks, no loss on either task, and an improvement in at least three of five builds. The margin failed all three conditions, so the strict gate was frozen. This selection consumed the additional data stated above and \$22.67 of teacher calls. It does not establish that a strict gate prevents validation overfitting.

\begin{table}[!htb]
    \centering\small\setlength{\tabcolsep}{6pt}\renewcommand{\arraystretch}{1.1}
    \caption{\textbf{Build-level gains and the reflection comparison.} Each initial-to-selected comparison uses the same build and validation-only selection. Gains are ordered by replicates 1--3. Protocol B starts from the same initial harnesses and uses the same gate. Average weights the seven settings equally. Individual gains use unrounded scores.}
    \label{tab:compilation_detail}
    \begin{tabular}{lrrlr}
        \toprule
        Setting & Initial $H_0$ & Selected $H_{\mathrm{val}}$ & Gain per build & Selected, protocol B \\
        \midrule
        InfoSeek & 33.5 & 37.7 & $+1.5$, $+2.2$, $+8.8$ & 35.9 \\
        DocVQA & 78.4 & 85.0 & $+11.5$, $+1.5$, $+7.0$ & 79.8 \\
        SlideVQA & 54.4 & 54.6 & $+1.5$, $+1.1$, $-1.8$ & 55.7 \\
        LiveVQA & 34.1 & 38.6 & $+5.6$, $+3.9$, $+4.1$ & 37.3 \\
        BLINK reflectance & 61.7 & 62.9 & $+3.0$, $+0.7$, $+0.0$ & 61.2 \\
        BLINK correspondence & 65.2 & 65.7 & $+0.5$, $+0.7$, $+0.2$ & 65.4 \\
        BLINK depth & 95.4 & 95.4 & $+0.0$, $+0.0$, $+0.0$ & 95.4 \\
        \midrule
        Average & 60.4 & 62.9 & $+2.5$ & 61.5 \\
        \bottomrule
    \end{tabular}
\end{table}

\paragraph{Formal runs.} Three replicates (1--3) per setting, each with its own $H_0$ and two arms from it: A and the frozen variant (protocol B with the strict gate). The replicates share the seed-0 $P$ and $G$ items. The three initial harnesses of a setting differ because the teacher is sampled. $H_0$ and the selected harness are evaluated on identical test items. The selected round comes from $G$ only. LiveVQA is scored by the fixed Sonnet-5 judge. Table~\ref{tab:formal_reps} lists every build, and Table~\ref{tab:compilation_main} gives the means. Formal teacher spend was \$56.88 (\$1.39 for the LiveVQA judge).

\begin{table}[!htb]
\centering\scriptsize\setlength{\tabcolsep}{3pt}
\caption{Formal builds: test scores of the shared $H_0$, the selected harness under the reported protocol (A) and under the incumbent-anchored variant (B), selected rounds (BLINK: per fold), and teacher cost in USD per arm. The $H_0$ column is the arm-A re-evaluation. The arm-B re-evaluation of the same $H_0$ agreed with it on 98.1--100\% of items (DocVQA lowest), an observed replay agreement under batched inference rather than a bound on future variability.}
\label{tab:formal_reps}
\begin{tabular}{@{}llrrrllrr@{}}
\toprule
Setting & Replicate & $H_0$ & selected (A) & selected (B) & round, A & round, B (no-ops) & \$ A & \$ B \\
\midrule
InfoSeek & 1 & 35.8 & 37.3 & 36.1 & 1 & 3 (0) & 1.19 & 0.54 \\
InfoSeek & 2 & 36.9 & 39.1 & 36.9 & 1 & 0 (0) & 0.83 & 0.57 \\
InfoSeek & 3 & 27.8 & 36.6 & 34.6 & 1 & 2 (0) & 0.91 & 0.71 \\
DocVQA & 1 & 74.3 & 85.8 & 79.1 & 1 & 3 (0) & 0.60 & 0.49 \\
DocVQA & 2 & 83.1 & 84.6 & 83.9 & 1 & 1 (0) & 0.66 & 0.54 \\
DocVQA & 3 & 77.7 & 84.6 & 76.4 & 3 & 3 (0) & 0.62 & 0.41 \\
LiveVQA & 1 & 33.4 & 39.0 & 37.6 & 3 & 1 (0) & 0.76 & 0.54 \\
LiveVQA & 2 & 33.0 & 37.0 & 36.1 & 2 & 1 (0) & 0.81 & 0.48 \\
LiveVQA & 3 & 35.9 & 40.0 & 38.1 & 2 & 3 (0) & 0.89 & 0.64 \\
SlideVQA & 1 & 56.5 & 58.1 & 58.8 & 1 & 3 (0) & 0.81 & 0.79 \\
SlideVQA & 2 & 53.0 & 54.1 & 54.0 & 3 & 1 (0) & 0.83 & 0.80 \\
SlideVQA & 3 & 53.6 & 51.8 & 54.3 & 1 & 1 (0) & 0.91 & 0.67 \\
BLINK reflectance & 1 & 61.2 & 64.2 & 61.2 & 0/1/1 & 0/0/2 (0) & 2.09 & 2.00 \\
BLINK reflectance & 2 & 61.9 & 62.7 & 60.4 & 2/2/2 & 3/0/1 (0) & 2.24 & 1.68 \\
BLINK reflectance & 3 & 61.9 & 61.9 & 61.9 & 0/0/0 & 1/3/1 (0) & 1.97 & 1.73 \\
BLINK correspondence & 1 & 66.7 & 67.1 & 67.3 & 0/0/1 & 0/0/2 (0) & 2.39 & 2.27 \\
BLINK correspondence & 2 & 64.2 & 64.9 & 63.9 & 1/0/2 & 2/1/3 (0) & 2.76 & 1.91 \\
BLINK correspondence & 3 & 64.9 & 65.1 & 64.9 & 0/1/0 & 0/1/3 (0) & 2.63 & 1.86 \\
BLINK depth & 1 & 96.0 & 96.0 & 96.0 & 0/0/0 & 0/0/0 (1) & 1.29 & 1.39 \\
BLINK depth & 2 & 95.2 & 95.2 & 95.2 & 0/0/0 & 0/0/0 (0) & 1.50 & 1.74 \\
BLINK depth & 3 & 95.2 & 95.2 & 95.2 & 0/0/0 & 0/0/0 (1) & 1.44 & 1.47 \\
\bottomrule
\end{tabular}
\end{table}

\paragraph{Paired statistics.} On InfoSeek, DocVQA and LiveVQA every A gain over $H_0$ is significant per build (InfoSeek 240/152, 229/97 and 706/178 fixed/broken items, McNemar $p\leq 1.1\times10^{-5}$; LiveVQA 166/54, 159/81 and 145/63, $p<10^{-6}$; DocVQA paired-bootstrap 95\% intervals $[+10.3,+12.6]$, $[+0.8,+2.2]$ and $[+5.9,+8.0]$). SlideVQA gains $+1.5$ ($p=0.04$) and $+1.1$ ($p=0.16$) in two builds and loses $1.8$ in the third ($p=0.016$). The BLINK paired changes are not statistically significant (reflectance 15/11, 15/14 and 0/0; correspondence 4/2, 11/8 and 1/0; depth identical verdicts in every fold). Protocol B against A on the same items: InfoSeek $-1.2$, $-2.2$, $-2.0$ ($p\leq 0.002$ each); DocVQA $-6.7$ and $-8.2$ (intervals exclude zero) and $-0.7$ (interval $[-1.4,+0.1]$); LiveVQA $-1.3$ ($p=0.04$), $-0.8$ ($p=0.25$), $-1.9$ ($p=0.002$); SlideVQA $+0.8$ ($p=0.25$), $-0.1$, $+2.5$ ($p<10^{-3}$); no statistically significant paired difference on BLINK. Protocol B selected the initial harness in InfoSeek replicate 2 and, as did A, in every depth fold. It returned two no-op replies in total and regressed below $H_0$ on test in three builds (DocVQA replicate 3, reflectance replicate 2, correspondence replicate 2) against one for A (SlideVQA replicate 3).

\paragraph{Reading.} Protocol B removes the rejected-parent lineage and produces one-component edits, as intended, but under the tested budgets it is the empirically weaker procedure on five of seven settings and better only on SlideVQA. The paper therefore reports protocol A as the method and B as a matched variant. The three-build means are an improvement under the tested budgets and settings. They are not evidence that any single proposal mechanism caused a gain, and the one SlideVQA regression shows that a validation-accepted edit can lose on test.

\subsection{Reflection A and the incumbent-anchored variant B}
\label{app:reflectv2}
The formal comparison changes the proposal procedure while keeping runtime interface v2, the initial harness, practice items and strict gate fixed. Table~\ref{tab:reflection_contract} makes the distinction explicit. B supplies the incumbent artifact and paired evidence, requests one coherent mechanism change, and allows a no-op. Its effects therefore cannot be attributed to prompt wording alone. A remains the reported procedure, and B is a matched alternative, not the meaning of ``runtime v2''.

\begin{table}[htbp]
\centering\small\setlength{\tabcolsep}{5pt}\renewcommand{\arraystretch}{1.12}
\caption{Reflection procedures compared on the same runtime. B is also called reflection protocol v2 in the experiment records. This name is independent of runtime interface v2.}
\label{tab:reflection_contract}
\begin{tabularx}{\linewidth}{@{}p{.19\linewidth}YY@{}}
\toprule
Component & A: last proposal & B: accepted incumbent \\
\midrule
Edit target & Latest proposal, accepted or rejected & Current incumbent, supplied as full JSON with its hash \\
Probe report & Latest proposal's report; paired traces against the preceding proposal & Incumbent's report; latest rejection shown separately with a compact paired report against the incumbent \\
History & Scores and accepted incumbent identified in the round history & Parent/proposal/incumbent hashes, changed components, scores and gate decision \\
Edit instruction & Return a revised complete harness & One coherent mechanism change, evidence and denominator, and a possible regression; unchanged incumbent is allowed \\
Stopping & Up to three reflection rounds & Up to three rounds; stop after two consecutive no-ops \\
Selection & Strict validation improvement, with bare floor & Same gate \\
\bottomrule
\end{tabularx}
\end{table}

The acceptance rule is unchanged: neither procedure uses test outcomes to accept or reverse a rejection. Editing the incumbent changes the parent of the next proposal, not the statistical resolution of the validation set. The formal outcomes in Table~\ref{tab:formal_reps}, including test regressions, are the evidence for comparing these procedures.

\FloatBarrier

\section{Decision diagnostics and cross-student transfer}
\label{app:decisions}
\subsection{Student-routed and single-decision controls}
\label{app:free}
The student-routed control exposes the same resources with a plain instruction: three candidate cards on InfoSeek and LiveVQA, page OCR on DocVQA, rounded raw signals on BLINK, and the whole deck plus OCR of five ranked slides on SlideVQA. It removes compiled routing, verdicts, parsers and output contracts. Main scores are in Table~\ref{tab:main}, while Table~\ref{tab:free} retains the paired uncertainty rather than repeating all baseline columns.
\begin{table}[!htb]
\centering\small\setlength{\tabcolsep}{7pt}\renewcommand{\arraystretch}{1.12}
\caption{\textbf{Cost of removing compiled control.} Differences are student-routed minus HC on matched test items, with negative values favoring HC. The single diagnostic harnesses differ from the formal means (Appendix~\ref{app:protocol}). A dash means that the interval or exact $p$-value was not reported. The average weights settings equally.}
\label{tab:free}
\begin{tabular}{@{}lrrr@{}}
\toprule
Setting & Score change & 95\% interval & Paired $p$-value \\
\midrule
InfoSeek & $-8.7$ & -- & $<0.001$ \\
DocVQA & $-3.6$ & $[-4.8,-2.5]$ & -- \\
SlideVQA & $-13.6$ & $[-15.6,-11.6]$ & $<0.001$ \\
LiveVQA & $-1.2$ & -- & $0.06$ \\
BLINK reflectance & $-14.2$ & -- & $0.013$ \\
BLINK correspondence & $-2.3$ & -- & $0.27$ \\
BLINK depth & $-22.6$ & -- & $<0.001$ \\
\midrule
Average & $-9.5$ & -- & -- \\
\bottomrule
\end{tabular}
\par\smallskip
\begin{minipage}{\linewidth}\footnotesize
References: InfoSeek, DocVQA and LiveVQA against the single current-interface diagnostic builds; SlideVQA and BLINK against the formal replicate-1 artifacts (fold artifacts pooled). Paired item counts (newly correct / newly incorrect after removing control): InfoSeek 220 / 740; LiveVQA 70 / 95; SlideVQA 122 / 424; reflectance 17 / 36; correspondence 29 / 39; depth 3 / 31. Their exact McNemar $p$-values are $2\times10^{-66}$ and 0.06. Confidence intervals and paired tests are distinct summaries, not interchangeable estimates.
\end{minipage}
\end{table}

\paragraph{One interface at a time.}
Table~\ref{tab:onedec_main} reports score changes on the single-build reference artifacts. Table~\ref{tab:intervention_definitions} specifies the operational change in each column. The 4B requests a card on 58\% of InfoSeek items and slides on 71\% of SlideVQA items. Gemma requests OCR on 5\% of DocVQA items and instead answers the underlying question on 92\%. BLINK request rates range from 80\% to 100\%. These rates describe the tested contracts rather than general tool-use ability.

\begin{table}[htbp]
\centering\small\setlength{\tabcolsep}{5pt}\renewcommand{\arraystretch}{1.18}
\caption{Definitions of the decision-interface interventions.}
\label{tab:intervention_definitions}
\begin{tabularx}{\linewidth}{@{}lYY@{}}
\toprule
Decision & Intervention & Interpretation \\
\midrule
Decide & Supply the resource only after a yes/no request & Tests activation under that request contract \\
Select & Show all candidates without the explicit pick or ranker & Changes the selection interface, not only the decision owner \\
Generate & Use a student-generated entity name for lookup & Tests open argument generation on InfoSeek \\
Integrate & Use raw content instead of rewritten content. On BLINK, use raw signals instead of the processed rule output & The intervention differs between content and perception tasks \\
Persist & Remove the answer contract and parser together (a bundled output interface). On BLINK, show the verdict as a suggestion & Tests the output contract and abstention control, or verdict emission; separate formal contract/extractor controls are in App.~\ref{app:rcontrols} \\
\bottomrule
\end{tabularx}
\end{table}

The interventions are not independent and their effects are not additive. On BLINK, the raw-signal integration arm is also the all-back arm. Each ablation is a single unselected variant compared with a validation-selected harness. Nonsignificance does not establish equivalence across all prompts. In particular, InfoSeek's selected harness already retains a student pick. Its select ablation removes the explicit two-pass choice and presents candidate content together. It should not be described simply as handing a code decision back to the student.

The all-back comparisons are recorded in Table~\ref{tab:free}. These are fixed-harness interventions, not additional compiler draws.

\subsection{Decision ownership and student-specific policies}
\label{app:dofmap}
\begin{figure}[!htbp]
    \centering
    \includegraphics[width=0.9\linewidth,trim=0 11 0 0,clip]{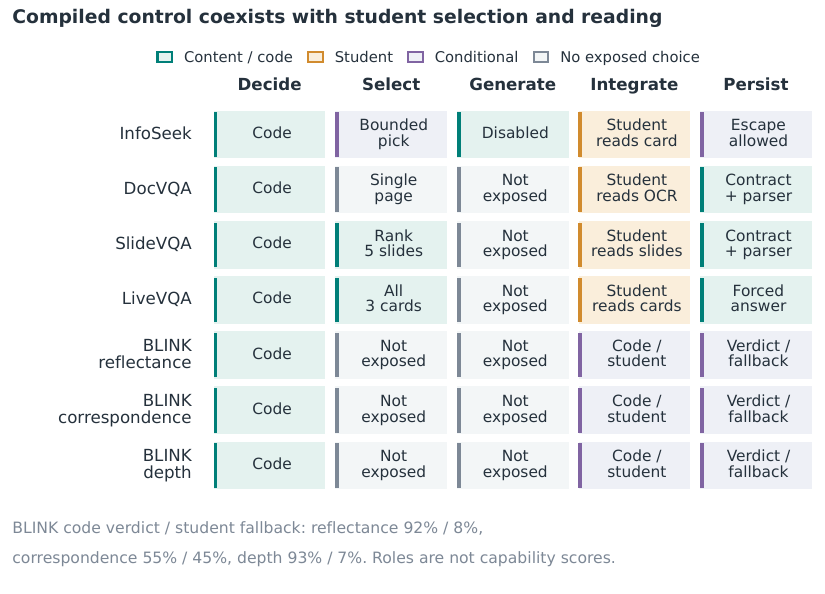}
    \caption{\textbf{Decision ownership in historical diagnostic programs.} Rates are program-specific, not formal means or capability scores. Interventions: Table~\ref{tab:onedec_main}.}
    \label{fig:beddof}
    \vspace{3pt}
    \includegraphics[width=\linewidth]{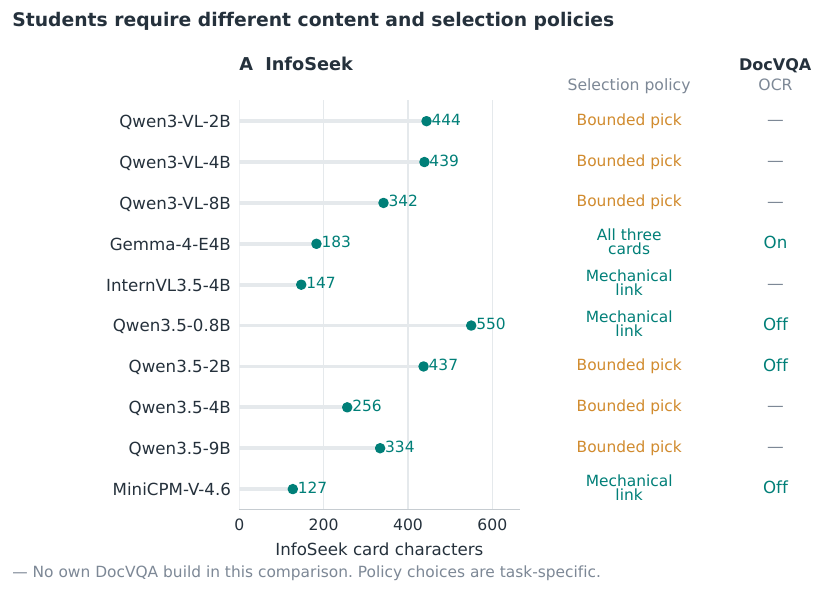}
    \caption{\textbf{Student-specific policies within each task.} InfoSeek shows card widths and selection methods for ten own builds. DocVQA shows whether OCR is kept in four own builds (dash: not evaluated). The 4B uses its restart-rule build. The 8B's second build scores 40.3, compared with 23.2 for its first and 34.7 for its transplant.}
    \label{fig:studentdof}
\end{figure}

Figures~\ref{fig:beddof} and \ref{fig:studentdof} describe historical single-build programs, not a capability scale. Comparisons stay within tasks: reading text, choosing an entry and answering after a rule defers are different events. InfoSeek retains bounded picks and content tasks retain reading, while BLINK source rules answer 55--93\% of items. LiveVQA and SlideVQA retain similar broad roles but have different control gains, so counting compiled decisions does not predict their value.
\FloatBarrier
\subsection{Cross-student transfer and recompilation}
\label{app:matrix}
Table~\ref{tab:predict} transplants the single interface-v2 InfoSeek diagnostic harness to nine students and compares it with an own build for each under the same protocol. Under the cleaned interface the own build beats the transplant on three of nine students and loses on five, with an average difference of $-0.4$. The gains concentrate on students that cannot execute the source program (Qwen3.5-0.8B, $+11.3$) or read cards differently (Qwen3-VL-2B, Gemma). The mid-sized Qwen3.5 and InternVL students do better with the transplanted harness than with their own single-draw build. Tables~\ref{tab:predict_full} and \ref{tab:fired} diagnose the earlier 1,200-item discriminate harness and distinguish its rewritten store from the reference store. These columns must not be treated as repeated measurements of one identical harness. Additional students receive each source harness unchanged. Some diagnostic measurements were collected after transfer evaluation. The correlations characterize transfer behavior rather than constitute an independent prospective prediction test.
\begin{table}[!htb]
\centering\footnotesize\setlength{\tabcolsep}{4pt}\renewcommand{\arraystretch}{1.16}
\caption{InfoSeek transfer and recompilation at 450 practice items under interface v2. The transplant is the earlier single interface-v2 Qwen3-VL-4B harness (37.8 on its source student) evaluated unchanged. Own builds use the same protocol (restart rule, bare floor, no profile). Two-sided exact McNemar tests compare the own build with the transplant on 6,000 matched test items. Values below 0.001 are shown as $<0.001$. These per-comparison tests do not measure variation across teacher draws. Average weights the nine receiving students equally. Each own build is a single teacher draw.}
\label{tab:predict}
\begin{tabular}{@{}lrrrrr@{}}
\toprule
Student & Bare & Transplant & Own build & Difference & Paired $p$-value \\
\midrule
Qwen3-VL-2B & 13.2 & 30.1 & 32.1 & $+2.0$ & $<0.001$ \\
Qwen3-VL-8B & 20.3 & 39.7 & 38.0 & $-1.8$ & $<0.001$ \\
Gemma-4-E4B & 6.7 & 30.8 & 33.0 & $+2.2$ & $<0.001$ \\
InternVL3.5-4B & 9.6 & 30.9 & 26.2 & $-4.7$ & $<0.001$ \\
\midrule
Qwen3.5-0.8B & 6.8 & 16.1 & 27.3 & $+11.3$ & $<0.001$ \\
Qwen3.5-2B & 9.0 & 30.7 & 26.0 & $-4.7$ & $<0.001$ \\
Qwen3.5-4B & 17.9 & 35.4 & 31.1 & $-4.4$ & $<0.001$ \\
Qwen3.5-9B & 20.5 & 38.7 & 38.0 & $-0.8$ & $0.11$ \\
MiniCPM-V-4.6 & 10.0 & 26.1 & 23.7 & $-2.3$ & $<0.001$ \\
\midrule
Average (9 recipients) & 12.7 & 30.9 & 30.6 & $-0.4$ & -- \\
\bottomrule
\end{tabular}
\end{table}
\begin{table}[!htb]
\centering\scriptsize\setlength{\tabcolsep}{4pt}
\caption{InfoSeek decision diagnostics for the 1,200-item source harness. Pick/net precision is measured on a 300-item probe (CLIP top-1 approximately 52\%). ``0'' is abstention on routed test items. $S$ uses the reference store, and routed gain is measured against $S'$ using the harness's own store.}
\label{tab:predict_full}
\begin{tabular}{lrrrrr}
\toprule
Student & Pick / net & ``0'' & Routed gain & $H-S$ ($p$) & $S-$bare \\
\midrule
Qwen3-VL-2B & 57.0 / 55.3 & 1\% & $+3.8$ & $+2.2$ ($10^{-5}$) & $+17.1$ \\
Qwen3-VL-4B & 66.0 / 58.0 & 17\% & $+7.5$ & $+6.3$ ($5\!\times\!10^{-36}$) & $+15.5$ \\
Qwen3-VL-8B & 61.2 / 55.7 & 12\% & $+6.6$ & $+6.1$ ($5\!\times\!10^{-31}$) & $+13.8$ \\
Gemma-4-E4B & 49.3 / 47.0 & 8\% & $+1.9$ & $+3.1$ ($3\!\times\!10^{-10}$) & $+22.3$ \\
InternVL3.5-4B & 50.8 / 48.3 & 21\% & $+0.5$ & $+2.6$ ($3\!\times\!10^{-6}$) & $+19.9$ \\
Qwen3.5-0.8B & 39.8 / 39.7 & 0\% & $+0.7$ & $+3.2$ ($3.5\!\times\!10^{-10}$) & $+11.8$ \\
Qwen3.5-2B & 87.7 / 48.7 & 85\% & $-3.9$ & $-0.4$ ($0.42$) & $+18.4$ \\
Qwen3.5-4B & 75.0 / 51.7 & 51\% & $+0.6$ & $+2.1$ ($1.8\!\times\!10^{-4}$) & $+13.9$ \\
Qwen3.5-9B & 76.9 / 55.3 & 32\% & $+5.8$ & $+5.7$ ($1.6\!\times\!10^{-29}$) & $+12.0$ \\
MiniCPM-V-4.6 & 53.8 / 49.0 & 19\% & $+0.1$ & $+0.2$ ($0.75$) & $+16.9$ \\
\bottomrule
\end{tabular}
\end{table}
\begin{table}[h]
    \centering
    \tablestyle
    \caption{Store-only ($S$) vs discriminate harness ($H_B$) on the same test items, split by whether the router fires (60\%, CLIP margin small) or not (40\%). ``0'' = share of fired items on which the student takes the escape option.}
    \label{tab:fired}
    \begin{tabular}{lrllll}
        \toprule
        student & ``0'' & linked-correct on fired $S\to H_B$ & acc.\ on fired $S\to H_B$ & acc.\ not fired $S\to H_B$ & $H_B - S$ \\
        \midrule
        Qwen3-VL-2B & 1\% & 32.5$\to$39.6 & 20.8$\to$25.1 ($+4.3$) & 44.7$\to$43.8 ($-0.9$) & $+2.2$ \\
        Qwen3-VL-4B & 17\% & 32.5$\to$42.8 & 23.4$\to$31.9 ($+8.5$) & 48.7$\to$51.7 ($+3.0$) & $+6.3$ \\
        Qwen3-VL-8B & 12\% & 32.5$\to$43.7 & 24.2$\to$31.8 ($+7.6$) & 49.1$\to$52.9 ($+3.8$) & $+6.1$ \\
        Gemma-4-E4B & 8\% & 32.5$\to$33.0 & 20.0$\to$22.4 ($+2.4$) & 42.8$\to$47.0 ($+4.2$) & $+3.1$ \\
        InternVL3.5-4B & 21\% & 32.5$\to$31.4 & 20.8$\to$22.1 ($+1.3$) & 42.7$\to$47.2 ($+4.5$) & $+2.6$ \\
        Qwen3.5-0.8B & 0\% & 32.5$\to$25.1 & 12.9$\to$14.3 ($+1.4$) & 27.1$\to$33.0 ($+5.9$) & $+3.2$ \\
        Qwen3.5-2B & 85\% & 32.5$\to$11.6 & 19.6$\to$14.7 ($-4.9$) & 39.2$\to$45.5 ($+6.3$) & $-0.4$ \\
        Qwen3.5-4B & 51\% & 32.5$\to$26.4 & 22.7$\to$23.7 ($+1.0$) & 45.6$\to$49.5 ($+3.9$) & $+2.1$ \\
        Qwen3.5-9B & 32\% & 32.5$\to$37.1 & 23.8$\to$29.9 ($+6.1$) & 45.6$\to$50.7 ($+5.1$) & $+5.7$ \\
        MiniCPM-V-4.6 & 19\% & 32.5$\to$33.5 & 18.6$\to$18.5 ($-0.1$) & 39.7$\to$40.3 ($+0.6$) & $+0.2$ \\
        \bottomrule
    \end{tabular}
\end{table}
The routed decision has distinct failure modes. Qwen3.5-2B takes the escape option on 85\% of routed items. MiniCPM's linking precision is close to the linker's, leaving little selection benefit. Whole-harness gains also include answer-side edits and, where applicable, a changed store. The with-sensitivity-profile harness is within two points of store-only on the five students for which it was evaluated.

\begin{table}[tbp]
\centering\scriptsize\setlength{\tabcolsep}{3pt}
\caption{Cross-task transfers on held-out test items. DocVQA uses the Gemma-compiled OCR harness, and BLINK uses the 4B's fold-specific harnesses. Correspondence bare scores in parentheses use 96 rather than 16 decode tokens. Dashes denote unreported comparisons.}
\label{tab:newstudents}
\begin{tabular}{lrrlll}
\toprule
Student & DocVQA bare & OCR $\Delta$ & Reflectance & Correspondence & Depth \\
\midrule
Qwen3-VL-2B & 91.4 & $-3.7$ & 29.9$\to$61.9 & 51.3$\to$58.1 & 74.2$\to$93.5 \\
Qwen3-VL-4B & 94.9 & $-0.7$ & 50.8$\to$64.9 & 54.9$\to$67.6 & 82.3$\to$95.2 \\
Qwen3-VL-8B & 95.3 & $-0.7$ & -- & -- & -- \\
Gemma-4-E4B & 72.7 & $+12.2$ & 46.3$\to$64.2 & 39.5$\to$62.6 & 64.5$\to$92.7 \\
InternVL3.5-4B & -- & -- & -- & -- & -- \\
Qwen3.5-0.8B & 84.9 & $-10.3$ & 28.4$\to$62.7 & 0.0 (17.9)$\to$57.4 & 67.7$\to$93.5 \\
Qwen3.5-2B & 91.5 & $-20.5$ & 19.4$\to$61.9 & 0.0 (19.3)$\to$55.6 & 69.3$\to$93.5 \\
Qwen3.5-4B & 94.7 & $-3.9$ & 50.0$\to$63.4 & 63.3$\to$66.2 & 81.5$\to$95.2 \\
Qwen3.5-9B & 95.4 & $-1.7$ & 55.2$\to$64.2 & 64.2$\to$69.2 & 87.9$\to$94.3 \\
MiniCPM-V-4.6 & 86.1 & $-4.5$ & 38.8$\to$62.7 & 37.9$\to$60.1 & 76.6$\to$95.2 \\
\bottomrule
\end{tabular}
\end{table}
The correspondence rule answers the same 243/441 items at 83.1\% precision for every receiving student. On deferred items, short decode budgets can prevent weaker students from returning an answer letter. A separate build for Qwen3-VL-2B returns a verdict on every item and reaches 63.3, versus 58.1 for the transplant.

\paragraph{Negative OCR transfer and subtractive edits.} With the earlier interface-v1 Gemma harness transplanted, Qwen3.5-2B echoed the injected OCR block on 364/3,568 items. On clean short answers the 0.8B and 2B lost about ten points, and 73--82\% of their new errors copied strings from the OCR. The transplant of the formal replicate-1 Gemma harness in Table~\ref{tab:fresh} harms the same students (8.8, 28.8 and 15.1 points below bare). In fresh builds under the current protocol, traced feedback turns OCR off on the two Qwen3.5 students, and the gate's bare floor retains the bare model for MiniCPM-V-4.6 after two restarts (Table~\ref{tab:fresh}). No fresh build is significantly above bare. The earlier fresh builds, which predate the floor, reached the same near-bare outcome by teacher edits alone and are kept in the release.
\begin{table}[htbp]
\centering\footnotesize\setlength{\tabcolsep}{5pt}\renewcommand{\arraystretch}{1.16}
\caption{Fresh DocVQA builds under the current protocol (interface v2, reflection A, strict gate with the bare floor and up to two round-0 restarts). The transplant is the formal replicate-1 Gemma harness evaluated unchanged. The two Qwen3.5 builds turn OCR off. For MiniCPM-V-4.6, every proposal was rejected and the gate retained the bare model (after two restarts, with the third initial harness breaking the parser). Selected scores are not significantly different from bare. Average weights the three students equally.}
\label{tab:fresh}
\begin{tabular}{@{}lrrrrrr@{}}
\toprule
Student & Bare & Transplant & Initial & Selected & $p$ vs bare & Teacher \$ \\
\midrule
Qwen3.5-2B & 91.5 & 82.7 & 91.5 & 91.6 & -- & 0.81 \\
Qwen3.5-0.8B & 84.9 & 56.1 & 85.7 & 85.8 & -- & 0.46 \\
MiniCPM-V-4.6 & 86.1 & 71.0 & 13.7 & 86.3 (bare retained) & -- & 1.04 \\
\midrule
Average & 87.5 & 69.9 & 63.6 & 87.9 & -- & 0.77 \\
\bottomrule
\end{tabular}
\end{table}

The builds use 150 probe items, 300 validation items, three reflections and GPT-5.5 without a profile. Scores are ANLS on 3,568 test items. The selected Qwen3.5-2B harness is round 3 with a parse hook, and the 0.8B keeps an image-first prompt in round 1. MiniCPM retains the bare model under the current gate. The small difference between its reported bare and selected scores reflects re-evaluation of that retained model, not acceptance of a parse-fix harness.

\FloatBarrier
\Needspace{13\baselineskip}
\subsection{Evidence for decision constraints and remaining reasoning gaps}
\label{app:reasoning}
\label{app:prior}
Section~\ref{sec:decisioncriteria} introduces the channel edge and decision-retention criterion. Here we collect the supporting observations and distinguish current results from exploratory and earlier diagnostics.

The observations below motivate limits on the tested interfaces. They do not identify an irreducible ``reasoning fraction'' of the teacher's advantage. If $r=\max_c\Delta_c/(\mathrm{acc}_L-\mathrm{acc}_S)$ is the fraction recovered by the channels tested in a pilot, $1-r$ may also reflect missing tools, imperfect retrieval, incompatible output contracts or limited search. IQ Test was screened but did not receive a comparable HC build, so it is not a measured failure of all possible harnesses.
\begingroup
\footnotesize\setlength{\tabcolsep}{5pt}\renewcommand{\arraystretch}{1.12}
\begin{longtable}{@{}p{.23\linewidth}p{.34\linewidth}p{.34\linewidth}@{}}
\caption{Evidence summary for decision constraints and reasoning limits. Current results and exploratory diagnostics are separated from earlier measurements, which used different splits or runtimes and are not pooled with the main HC comparisons.}\label{tab:reasoning}\label{tab:dof}\\
\toprule
Evidence / setting & Observation & Supported interpretation \\
\midrule
\endfirsthead
\caption[]{Evidence summary for decision constraints and reasoning limits (continued).}\\
\toprule
Evidence / setting & Observation & Supported interpretation \\
\midrule
\endhead
\midrule
\multicolumn{3}{r}{\emph{Continued on the next page}}\\
\endfoot
\bottomrule
\endlastfoot
BLINK diagnostic builds & HC trails the teacher by 14.9 points on reflectance, 20.1 on semantic and 12.3 on functional correspondence; depth is 5.6 higher (n.s.). & Available measurements support some decisions but do not close every task gap (Fig.~\ref{fig:blink}). \\
Current compute control & CoT with voting reaches HC on reflectance and correspondence, with much higher generated-token counts (Table~\ref{tab:token_efficiency}). & Some gains can also be obtained by increasing the student's reasoning compute. \\
Exploratory MathVista / IQ Test & MathVista: best tested tool adds 4.2 against a 19.2 teacher gap. IQ Test: teacher 51.3 versus 4B 24.0 on 150 items; no suitable carrier identified. & A teacher advantage alone does not establish a usable external channel. Unrecovered gap is not an isolated measure of reasoning. \\
Earlier InfoSeek lookup & Student-routed lookup: $+1.6$ versus $+5.9$ for mechanical linking; bounded candidate selection: $+6.9$. & Open argument generation and bounded selection impose different demands. \\
Earlier code-tool diagnostic & MathVerse: self-routed interpreter adds 10.2 for the 4B but loses 3.6 for InternVL; execution depends on the generated code. & A computation tool still requires valid, grounded arguments from its executor. \\
Earlier advice-block diagnostic & On WeMath, a failure-derived standing block costs the 4B 12.8 points; its benefit over generic advice is not significant. & Added guidance can interfere with reasoning under a particular model and prompt contract. \\
\end{longtable}
\endgroup
The earlier code-tool diagnostic used 2,615 MathVerse items, and the WeMath comparison used 1,160 items, with advice derived from 500 practice failures. These controls motivated the taxonomy, while the main single-decision ablations provide the direct evidence for the current compiled harnesses. The practical boundary is whether useful evidence or computation can be expressed through the available interface and reliably consumed by the student. The experiments do not establish that reasoning is inherently uncompilable.

\FloatBarrier

\section{Additional VQA results}
\label{app:vqa_results}
\subsection{InfoSeek: content, placement and store lineage}
\label{app:infoseek}
\begin{figure}[!htb]
    \centering
    \includegraphics[width=\linewidth]{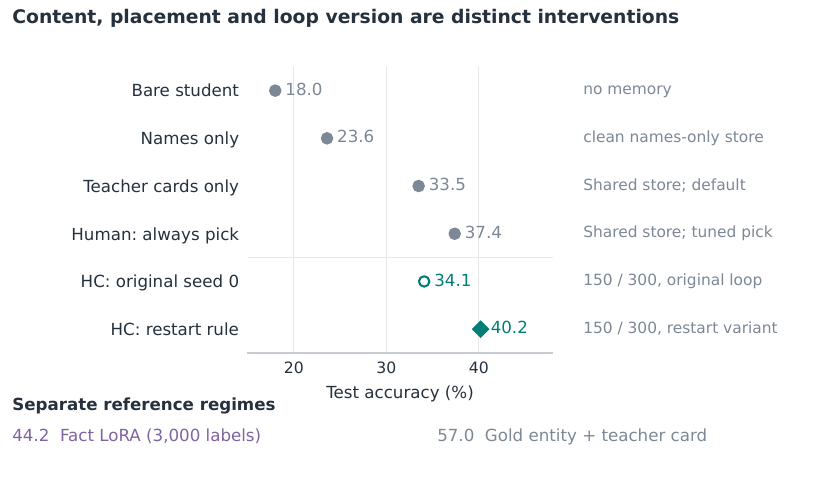}
    \caption{\textbf{Content, placement and compiler version are separate comparisons.} InfoSeek $\times$ Qwen3-VL-4B, test $n=6{,}000$. Names-only uses the cleaned store, while teacher cards and tuned human placement use the reference store. The two 150/300 builds use the same seed, splits and store, with different loop procedures. The restart-rule draw started sound without triggering a restart. Points are named arms, not an additive decomposition. The weight-trained and gold-linked references are separate regimes.}
    \label{fig:main}
\end{figure}
Store lineage: build 1 (1,200 items) rewrote all 1,380 cards in its round 0 (median card 432 characters against 496). Builds 2 and 3, the \dof{}-menu build, the tuned human and GEPA reuse the reference store. The transplant matrix uses build 1's harness with its own store and the reference store for the store-only arm, so the matrix's $H_B - S$ includes the store difference, measured for every student by a store-only run on the rewritten store ($S'$): $+0.4$ (Qwen3-VL-2B), $+0.9$ (4B), $+1.1$ (8B), $+0.5$ (Gemma), $+1.5$ (InternVL), $+1.0$, $-0.9$, $+0.6$, $+0.2$ (Qwen3.5 0.8B/2B/4B/9B), $-0.2$ (MiniCPM-V-4.6). Table~\ref{tab:predict_full} distinguishes routed gains against $S'$ from whole-harness gains against $S$. Per linking condition, accuracy when the shown card is the gold entity's rises from 36.4 (name card) to 49--52 ($H_0$) to 55--57 (best) and to 59.9 under the discriminate harness. On wrongly linked items it stays 9--11.

\begin{table}[tbp]
    \centering\scriptsize\setlength{\tabcolsep}{3pt}
    \caption{InfoSeek $\times$ Qwen3-VL-4B: the bounded pick decomposed. $S$ uses the reference store, and $S'$ uses build 1's rewritten store. Gains are relative to $S'$ for build 1 and $S$ otherwise. Parentheses report linked-correct rates before $\to$ after the pick. ``Rest'' means items answered without the pick.}
    \label{tab:decomp}
    \begin{tabular}{llrlrr}
        \toprule
        Harness & Store & Routed & Routed gain (linking) & Rest gain & Total \\
        \midrule
        Store-only $S$ / $S'$ & Ref. / rewritten & -- & -- & -- & 33.5 / 34.3 \\
        Tuned human pick & Ref. & 100\% & -- & -- & 37.4 \\
        HC build 1 (1,200 items) & Rewritten & 60\% & $+7.5$ (32.5 $\to$ 42.8) & $+2.3$ & 39.7 \\
        HC build 2 & Ref. & 68\% & $+8.0$ (36.7 $\to$ 46.2) & $+4.4$ & 40.3 \\
        HC build 3 & Ref. & 61\% & $+4.2$ (34.0 $\to$ 44.6) & $-1.1$ & 35.6 \\
        HC, decision-menu profile & Ref. & 76\% & $+6.5$ (38.8 $\to$ 48.4) & $+1.8$ & 38.8 \\
        HC, sensitivity (3 builds) & Own / ref. & 0\% & No pick retained & $+0.2$ to $+0.5$ & 33.7--34.0 \\
        \bottomrule
    \end{tabular}
\end{table}

\subsection{DocVQA: consolidated baseline comparison}
\label{app:docvqa}
\begin{table}[htbp]
\centering\small\setlength{\tabcolsep}{6pt}\renewcommand{\arraystretch}{1.14}
\caption{DocVQA baselines with Gemma-4-E4B. Scores are ANLS on 3,568 test items.}
\label{tab:docvqa}
\begin{tabularx}{\linewidth}{@{}Ylr@{}}
\toprule
Configuration & Practice items & ANLS \\
\midrule
Bare & None & 72.7 \\
Human: plain OCR, 4,000 characters & Tuned on 500 & 80.1 \\
Human: typed OCR field & Tuned on 500 & 82.4 \\
Unoptimized DSPy typed program & None & 86.2 \\
One-shot teacher harness & None & 85.5 \\
HC, 50 probe / 50 validation & 100 & 84.4 \\
HC, 150 probe / 300 validation & 450 & 85.6 \\
HC, 500 probe / 500 validation & 999 unique & 84.9 \\
Answer LoRA & 999 & 84.7 \\
Same LoRA with the 999-item HC & 999 & 84.7 \\
Teacher, GPT-5.5 at 2,048 pixels & Not applicable & 93.3 \\
\bottomrule
\end{tabularx}
\end{table}
The earlier GPT-5.5 builds use always-on OCR, a short output contract and a parse hook. Question-specific special cases are rejected by validation. Compiler-input effects are reported in Table~\ref{tab:ablate}, and typed-field serialization is isolated in App.~\ref{app:typed}. Plain chain-of-thought scores 75.2. Adding it to plain OCR scores 78.8 against 80.1 for plain OCR alone.

\subsection{BLINK: verdict coverage and task-dependent residuals}
\label{app:blink}
\begin{figure}[!htb]
    \centering
    \includegraphics[width=\linewidth]{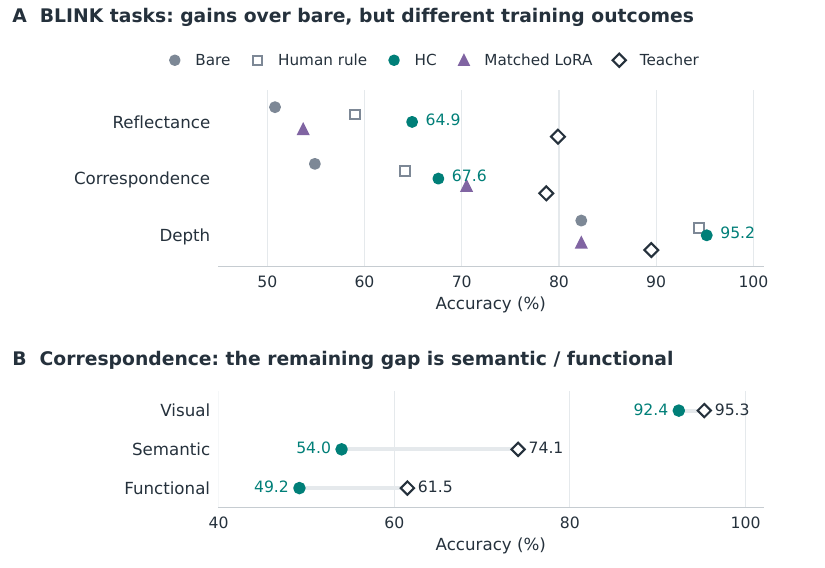}
    \caption{\textbf{Perception gains and remaining teacher gaps differ by task.} (A) Bare, human rule, \hc{}, matched-label LoRA and GPT-5.5 at medium effort on the same items. \hc{} and LoRA pool held-out folds. LoRA uses 89 / 291 / 82 labels per fold. (B) The correspondence subsets, not additional independent benchmarks. Semantic and functional gaps remain large, and no \hc{} build significantly exceeds its fixed human rule.}
    \label{fig:blink}
\end{figure}
\begin{table}[tbp]
\centering\small\setlength{\tabcolsep}{4pt}
\caption{BLINK verdict and fallback subsets (three held-out folds pooled). Paired scores compare the harness with bare on the same subset. A dash denotes no fallback items.}
\label{tab:blinkpart}
\begin{tabular}{llrrrr}
\toprule
Setting & Harness & Verdict share & Verdict / bare & Fallback / bare & Total \\
\midrule
Reflectance & HC seed 0 & 92\% & 65.9 / 50.4 & 54.5 / 54.5 & 64.9 \\
 & HC seed 1 & 100\% & 61.9 / 50.7 & -- & 61.9 \\
 & Human & 84\% & 60.2 / 50.4 & 52.4 / 52.4 & 59.0 \\
Correspondence & HC seed 0 & 55\% & 83.1 / 65.0 & 48.5 / 42.4 & 67.6 \\
 & HC seed 1 & 65\% & 76.0 / 59.9 & 47.4 / 45.5 & 66.0 \\
 & Human & 93\% & 65.0 / 55.0 & 53.3 / 53.3 & 64.2 \\
Depth & HC seed 0 & 93\% & 96.5 / 82.6 & 77.8 / 77.8 & 95.2 \\
 & HC seed 1 & 94\% & 96.6 / 82.1 & 85.7 / 85.7 & 96.0 \\
 & Human & 94\% & 94.9 / 82.1 & 85.7 / 85.7 & 94.4 \\
\bottomrule
\end{tabular}
\end{table}
The teacher and compiled harness are evaluated on identical items. On visual / semantic / functional correspondence, HC scores 92.4 / 54.0 / 49.2 against the teacher's 95.3 / 74.1 / 61.5. The remaining differences therefore vary substantially within the family. No compiled build significantly exceeds its task's fixed human rule.

\paragraph{Tool access is not enough.} In the exploratory tool comparison, the 4B requests signals on 95--100\% of depth, counting and spatial items, but on only 2\% of visual-correspondence items. Tool-only accuracy is below the student on four of the five screened tasks. These contract-specific observations motivate the controlled activation interventions in Table~\ref{tab:onedec_main} but are not included in the main cross-build average.

\paragraph{Build variation.} Per-fold test scores for seed 0 / seed 1 are: reflectance 71.1, 71.1, 52.3 / 62.2, 64.4, 59.1; correspondence 72.7, 68.0, 61.8 / 64.7, 61.2, 72.2; depth 97.6, 97.6, 90.2 / 97.6, 95.1, 95.1. The selected round is round 0 in all six depth folds. On reflectance fold 2, validation prefers a round scoring 52.3 on test over the initial 56.8 (Fig.~\ref{fig:loop}).

\subsection{LiveVQA: retrieval, contracts and scoring}
\label{app:livevqa}

\textbf{Bed.} LiveVQA-2025 news sources (6,691 instances, April--May 2025), the first L1 (what is shown) and first L2 (a fact from the story) question per instance. The practice split contains 2,000 instances (probe 150, gate 300, teacher pool 150), and the test split contains 1,000 instances, or 2,000 questions.

\paragraph{Retrieval channel.} CLIP ViT-L/14 links each image to headlines across all articles (gold at rank 1 on 41\%, in the top three on 55\%). Each candidate's card contains its two passages most similar to the question.

\paragraph{Why the human recipe trails student routing.} The main-table human recipe (25.5) uses the student to pick one of the three candidate headlines before reading its article card, whereas the student-routed control (36.9) shows all three cards directly. The former creates an early information bottleneck: if the selected headline is wrong, the answer stage cannot inspect the other candidate bodies. The latter lets the student judge relevance while reading those bodies. This is a plausible explanation for the 11.4-point gap, consistent with the benefit of retaining all three cards in HC. It is not an isolated selection ablation: the arms also differ in prompting and output handling, and the available records do not apportion the gap between these factors. ``Human recipe'' identifies a manually designed policy, not a human oracle or an optimized upper bound.

\paragraph{Scoring and judge audit.} Open answers are judged correct / incorrect / not attempted against the gold list, accuracy = correct over all items. The judge whose numbers the paper reports is Claude Sonnet 5, a model from a different provider that took no part in compiling. The same prompt run with GPT-5.5, the teacher's model (validated on 30 hand-labelled items, 28 agreements), is used as a scoring audit, alongside the LLM-free normalized containment metric used by the loop for probe and validation scores. The two judges agree on 96--99.8\% of items per arm (Cohen's $\kappa$ 0.85--0.98), Sonnet is the more lenient by 0.2--1.9 points on every arm, harness and non-harness alike, and the positive HC-versus-human and HC-versus-top-1 conclusions hold under all three metrics. The judge is stricter than containment on the bare and recipe arms and more lenient on the harness arms under both judges, so this is a property of answer form (cards yield names in forms a string match misses), not a preference of the teacher's model for its own harness. The harness prompt has no abstention option, so the bare of Tab.~\ref{tab:main} uses that prompt. The vqa\_eval prompt with ``answer Unknown'' is a control (Sonnet accuracies 6.7 / 10.8 / 13.7 / 1.1 in the table's student order). GPT-5.5 with the top-three passages scores 58.3 under the Sonnet judge on a separate 300-question practice pool.
\begin{table}[tbp]
\centering\small\setlength{\tabcolsep}{4pt}
\caption{LiveVQA test accuracy ($n=2,000$), judged by Claude Sonnet 5. Human selects one headline, top-1 injects the mechanically linked article, and gold supplies the gold article. HC beats human ($p<10^{-35}$) and top-1 ($p<10^{-5}$) for each student. Average weights the four students equally. All four HC rows are single builds under interface v2 with reflection A (the Qwen3-VL-4B row is the diagnostic reference, not the three-build main-table mean, and the other three were recompiled under the current protocol for this version, replacing earlier-interface builds). The 9B's validation gate, which optimises containment, selected a round-3 harness that the judge scores 1.8 below its initial harness.}
\label{tab:livevqa}
\begin{tabular}{lrrrrrr}
\toprule
Student & Bare & Human & $H_0$ & $H_{\mathrm{val}}$ & Top-1 & Gold \\
\midrule
Qwen3-VL-4B (interface v2) & 14.7 & 25.5 & 37.5 & 38.1 & 27.4 & 51.6 \\
Qwen3.5-4B & 14.1 & 16.6 & 35.2 & 35.2 & 24.3 & 44.5 \\
Qwen3.5-9B & 16.9 & 22.5 & 37.9 & 36.1 & 29.9 & 55.1 \\
MiniCPM-V-4.6 & 7.1 & 13.8 & 22.5 & 30.1 & 5.0 & 8.3 \\
\midrule
Average & 13.2 & 19.6 & 33.3 & 34.9 & 21.7 & 39.9 \\
\bottomrule
\end{tabular}
\end{table}
\textbf{What the builds did.} Kept harnesses (no profile, three rounds, under \$1.5 of teacher calls each, interface v2): all three cards shown with no student pick, a short-answer contract, a parse hook. The Qwen3.5-4B kept its initial harness (every proposal rejected on the gate), while the 9B and MiniCPM-V-4.6 selected round-3 harnesses. The earlier interface-v1 builds of these three students (35.4 / 33.9 / 26.1 judged) are kept in the release. The 9B's v1 build had added a forced headline pick and repaired a below-bare round 0 by traced reflection.

\textbf{Why the student-routed arm is close to $H_{\mathrm{val}}$ here.} The selected 4B harness (interface v2) retains reading and evidence selection with the student: it shows all three cards with their bodies restructured with task-specific notes, and keeps a parser. Select and integrate stay with the student because the reflection found no selection rule that beat the student reading all three (the retain criterion). The student-routed arm offers the same three cards with a plain instruction, so both require the student to read the supplied evidence and differ only in presentation and output contract. The paired difference is 1.2 points (70 fixed, 95 broken, $p=.06$), and handing the output contract back to the student changes 4 items ($+0.2$). In this diagnostic artifact, the small contract-removal effect and the content interface are consistent with most of the observed advantage arising before output extraction. They do not decompose every formal build.

\paragraph{Abstention changes the metric tradeoff.} The remaining systematic difference is abstention: the v2 harness leaves 166 items not attempted against the student-routed arm's 627, and the LiveVQA F-score, which charges an abstention half a wrong answer, favours the student-routed arm (43.7 against 39.8). The gate optimised the probe's containment score, under which a wrong guess costs no more than an abstention. A gate on the task's own F-score would select differently. Where the compiler did remove a decision (a verdict on depth or reflectance, a page policy on SlideVQA, a bounded pick on InfoSeek) the student-routed arm gives up 5--22 points (App.~\ref{app:free}). Validation is 3--7 points above test for every arm including bare, so only within-test contrasts are reported. MiniCPM's numbers under the ``Unknown'' prompt (0.9 bare, 4.8 with the top-1 passage, 8.2 with the gold article) show a student whose failure is the abstention contract rather than the content: the same passages in the harness prompt give 20.2 containment.

\paragraph{Evaluation cost.} Teacher edge \$26 (44 answers empty at a 4k reasoning budget were re-asked at 12k), GPT-5.5 judge \$25 over 22 arms with a cross-arm verdict cache, Sonnet 5 judge \$16 over 35 arms, builds \$4.

\subsection{SlideVQA: page selection and image budget}
\label{app:slidevqa}

\textbf{Bed.} SlideVQA (NTT release): 20-slide decks and 14,484 questions. We use the validation split (1,652) as the practice pool (probe 150, gate 300, teacher pool 150) and the test split (2,215) once. Reasoning types are derived from the annotations: numerical (has an arithmetic expression, 194), multi-hop (more than one evidence slide, 503), single-hop (1,518).

\paragraph{Available evidence.} Tesseract OCR per slide (330 characters on average), a MiniLM question-to-OCR ranking and a CLIP ViT-L/14 question-to-slide ranking (all gold slides in the top three on 75--80\% of single-hop and 43--46\% of multi-hop questions). CLIP is two to five points above the OCR ranker.

\paragraph{Scoring and references.} SQuAD-style exact match and token F1. Bare = the whole deck as twenty 448-px images. Human = find-then-read: the MiniLM top-3 slides at 1,024 px with their OCR and an \texttt{Answer:} field.
\begin{table}[htbp]
\centering\footnotesize\setlength{\tabcolsep}{7pt}\renewcommand{\arraystretch}{1.15}
\caption{SlideVQA adaptation across students. Exact match on 2,215 test items. Average uses the same four receiving students in every column, excluding the source Qwen3-VL-4B.}
\label{tab:slidevqa}
\begin{tabular}{@{}lrrrrr@{}}
\toprule
Student & Bare & Human & Initial HC & Selected HC & Transplant \\
\midrule
Qwen3-VL-4B & 44.7 & 48.7 & 41.9 & 56.7 & n/a \\
Qwen3.5-4B & 52.6 & 38.0 & 48.2 & 61.4 & 50.7 \\
Qwen3.5-9B & 51.4 & 47.4 & 56.4 & 62.1 & 63.3 \\
MiniCPM-V-4.6 & 34.8 & 33.0 & 28.4 & 38.8 & 29.4 \\
Gemma-4-E4B & 43.2 & 35.8 & 38.5 & 48.4 & 43.0 \\
\midrule
Average (4 recipients) & 45.5 & 38.6 & 42.9 & 52.7 & 46.6 \\
\bottomrule
\end{tabular}
\end{table}

Bare uses the full deck at 448 pixels. The transplant is the Qwen3-VL-4B harness without changes. Selected HC exceeds the human recipe at $p<10^{-8}$ and bare at $p<10^{-3}$ on every row, using paired exact McNemar tests.

\begin{table}[htbp]
\centering\small\setlength{\tabcolsep}{9pt}\renewcommand{\arraystretch}{1.15}
\caption{SlideVQA input controls. Gold-slide controls marked $^{a}$ use the earlier human contract.}
\label{tab:slidevqa_inputs}
\begin{tabular}{@{}lrrr@{}}
\toprule
Student & Bare, 224 pixels & Top-three images & Gold slides \\
\midrule
Qwen3-VL-4B & 32.6 & 48.7 & 58.6 \\
Qwen3.5-4B & 36.7 & 33.6 & 37.0$^{a}$ \\
Qwen3.5-9B & n/a & 31.2 & 55.1$^{a}$ \\
MiniCPM-V-4.6 & n/a & 20.1 & 26.6$^{a}$ \\
Gemma-4-E4B & 34.6 & 39.0 & 47.6 \\
\bottomrule
\end{tabular}
\end{table}

$^{a}$ These controls change both slide selection and the answer contract relative to the final recipe. GPT-5.5 scores 64.7 with the linker's top-three slides and 78.7 with gold slides on the 150-question practice pool. The 4B's OCR-only arm scores 27.2 and its top-one-slide arm 37.6. Evaluating the labelled arithmetic expressions gives 93.8 on the numerical subset.
\textbf{What the builds did.} Round 0 sat below the student's bare on validation for four of five builds (the teachers reached for OCR and five slides, which the images-only control shows is negative for the 4B) and the traced reflections repaired every one: $H_{\mathrm{val}}-H_0$ is $+14.7$ / $+13.1$ / $+5.7$ / $+10.4$ on test. Every kept harness fuses the MiniLM, CLIP and a lexical ranking into five slides at 1,024 px, shows OCR snippets, sets a short contract and a parse hook. None retains a student pick or a student second hop. OCR is neutral for the 4B (images-only equals the recipe), worth $+4$ to $+16$ for the three Qwen3.5/MiniCPM students, and $-3.2$ for Gemma-4-E4B, whose bare whole-deck reading (43.2) also beats every selection recipe. Its build still adds $+5.2$ over bare by raising the gold-slide recall of the shown pages from 68\% to 87\% and lands at the gold-page control (47.6). The 4B's harness transfers whole to the 9B ($+11.9$, equal to the 9B's own build) and harms Qwen3.5-4B and MiniCPM, whose own builds keep shorter contracts. The numerical subset moves from 16.0 to 41.2 for the 4B with no compiled arithmetic.

Branch-support diagnostics are reported in App.~\ref{app:audit}.

\FloatBarrier

\section{Comparisons with alternative adaptation methods}
\label{app:baselines}
\subsection{Fixed and tuned human harnesses}
\label{app:human}
Human recipes are written from the task description and available tools. Thresholds are fixed in advance or tuned on development items. The same authors construct these recipes and run the compiler. BLINK rules were written before those builds finished, whereas the InfoSeek placement grid was evaluated after the teacher builds. Fixed public-content recipes and tuned placement over teacher-written cards are distinct comparators.
\textbf{Human policies on the teacher's store and tuned grids.} InfoSeek (teacher's store, validation 600, test 6,000): a grid over margin routers $a\in\{0.30,0.33,0.36\}\times b\in\{0.01,0.02,0.04\}$, ``never'', ``always'' and a MiniLM reranker; best on validation = always discriminate (41.33) $\to$ test 37.37; the teacher's no-profile builds 42.67/41.67 $\to$ 39.73/40.27 (smaller validation$\to$test drop). DocVQA (validation 500): OCR length $\{1000,2000,4000\}\times$ suffix $\{$plain, exact-span + \texttt{Answer:} + parse$\}$; best = 4,000 plain (81.83) $\to$ test 80.11; every exact-span variant was worse (77--79). BLINK: fixed a-priori rules (Retinex $\sigma$15 with $|\Delta|<0.08\to$ same; LightGlue projection within 8\% of the diagonal else DINO argmax; Depth-Anything-V2-Small at radius 4), verdict-only: 59.0 / 64.2 / 94.4.
\begin{table}[h]
\centering\scriptsize
\caption{Human heuristics: settings, routing and verbatim prompt text. LiveVQA scores use the Sonnet judge.}
\label{tab:humanspec}
\resizebox{\linewidth}{!}{%
\begin{tabular}{p{2.1cm}p{4.1cm}p{6.0cm}r}
\toprule
bed & setting & prompt text (verbatim) & test \\
\midrule
InfoSeek, \emph{recognise then read} & stage 1: the three CLIP candidate names as options, store of name cards; stage 2: the two Wikipedia passages of the picked entity most similar to the question (MiniLM), fallback to the CLIP top-1 entity on abstention; 64 decode tokens & stage 1: ``Which of these is the entity shown in the image?\textbackslash n\{options\}\textbackslash n0. None of them\textbackslash nAnswer with the number only.''; stage 2: ``Retrieved from Wikipedia (about the entity the retriever believes is shown; may be the wrong entity):'' + passages + ``Answer the question with a single word, number, or short phrase.'' & 33.8 \\
DocVQA, \emph{typed OCR field} & page image + question + 4,000 characters of Tesseract OCR in the DSPy typed-fields serialisation; generic field extractor; 96 decode tokens & context frame ``Context (OCR text of the page; may contain errors):\textbackslash n\{tool\_text\}''; objective ``Answer the question about the document image with a single word, number, or short phrase, copied from the page.'' & 82.4 \\
BLINK reflectance & Retinex log-reflectance ($\sigma{=}15$) at A and B; $|\Delta|<0.08 \to$ ``same'', else the lower one is darker; verdict answers the item, else the student & tool text shown to the student on deferred items: ``Local reflectance estimate (brightness relative to surroundings): A = \%.2f, B = \%.2f'' & 59.0 \\
BLINK correspondence & LightGlue projects REF into image 2; if the nearest candidate is within 8\% of the diagonal, pick it; else the candidate with the highest DINOv2 $3{\times}3$ cosine to REF; defer if no signal & ``Keypoint matcher projects REF next to point \%s'' / ``Feature similarity to REF: A 0.xx, B 0.xx, \ldots'' & 64.2 \\
BLINK depth & Depth-Anything-V2-Small at radius 4; larger = closer; defer if a marker is missing & ``Depth estimator: relative closeness at A = \%.2f, at B = \%.2f (larger = closer)'' & 94.4 \\
LiveVQA, \emph{pick a headline, read it} & the three CLIP-linked article titles as options, the picked article's card ($k{=}1$) & the InfoSeek stage-1 prompt over article titles; card frame ``Retrieved from your memory store (may or may not describe the right entity):\textbackslash n\{cards\}'' & 25.5 \\
SlideVQA, \emph{find three slides, read them} & MiniLM top-3 slides at 1,024 px + 1,500 characters of their OCR; 64 decode tokens & ``Answer with a single word, number, or short phrase copied from the slides. Respond in exactly this format:\textbackslash nAnswer: <answer>'' & 48.7 \\
\bottomrule
\end{tabular}}
\end{table}

\subsection{Test-time scaling}
\label{app:tts}

\begin{table}[h]
\centering\small
\caption{Compute-only baselines (test, same items). The $H_{\mathrm{val}}$ column is the formal replicate-1 harness of each setting re-executed with token logging (fold artifacts pooled on BLINK). Bare scores are in Table~\ref{tab:main}. Vote = majority over 5 sampled direct answers (9 on BLINK). CoT+vote uses reasoning-first rollouts. TTAug = 16 augmented views with token-level averaging \citep{ttaug}, paired against its own single-view control on the same backend ($\Delta$ shown). Generated-token costs for these same comparison runs are in Table~\ref{tab:token_efficiency}. Average weights all six settings equally for vote, CoT+vote and HC. TTAug is omitted because InfoSeek is missing.}
\label{tab:tts}
\resizebox{\linewidth}{!}{%
\begin{tabular}{lrrrr}
\toprule
bed $\times$ student & vote & CoT+vote & TTAug ($\Delta$ vs own ctrl) & $H_{\mathrm{val}}$ \\
\midrule
DocVQA $\times$ Gemma-4-E4B & 72.5 & 80.4 & 74.8 ($+1.6$) & \bfs{85.7} \\
SlideVQA $\times$ 4B & 44.6 & 51.1 & 46.8 ($+2.1$, $N{=}8$) & \bfs{58.1} \\
InfoSeek $\times$ 4B & 17.6 & 24.6 & -- & \bfs{37.3} \\
BLINK reflectance $\times$ 4B & 50.8 & \bfs{64.9} & 51.5 ($-1.5$) & 64.2 \\
BLINK correspondence $\times$ 4B & 54.0 & \bfs{68.5} & 56.9 ($+2.7$) & 67.4 \\
BLINK depth $\times$ 4B & 82.3 & 85.5 & 86.3 ($+4.0$, n.s.) & \bfs{96.0} \\
\midrule
Average (6 settings) & 53.6 & 62.5 & -- & \bfs{68.1} \\
\bottomrule
\end{tabular}}
\end{table}

\paragraph{Why these comparators.} The human recipes require no teacher compilation calls, although both human and compiled harnesses still incur student and tool inference costs. A third option is to spend the \emph{student's} compute at inference with no teacher and no labels. Two recent VLM methods motivate these controls. AVIS \citep{avis} combines visual-token pruning with self-consistency rollouts in chain-of-thought mode (temperature 0.7, top-$p$ 0.9, majority vote), with a learned predictor choosing the number of rollouts $K$ per query. We evaluate only fixed-$K$ self-consistency with these sampling settings, not AVIS's learned allocation policy or visual-token pruning. Its reported policy-training procedure uses 5,000 calibration questions with 10 rollouts each, beyond our practice-data budgets. Kaya et al.\ \citep{ttaug} decode $N$ augmented copies of the input jointly, averaging the next-token distributions at every step (TTAug). We run TTAug at $N=16$ where feasible ($N=8$ for SlideVQA), using three random image transforms from the paper's high-strength list plus keyboard-typo / word-split / word-drop text augmentation with the original appended after ``In other words'', and the sample-and-rank variant (highest cumulative log-probability). Every SC arm is a decode option of the harness runtimes (\texttt{decode.samples}), so bare, human and $H_{\mathrm{val}}$ specs are sampled unchanged, and every candidate rollout is stored. TTAug runs on HF transformers, so each augmented arm is paired with its own single-view control on that backend (0.4 above vLLM for Gemma, 2.5 below for the 4B at 1,024 px). Per-item TTAdapt (weight updates per test item) is out of scope. Vote normalization preserves multiple-choice answer letters.

\paragraph{Results (Table~\ref{tab:tts}).} A vote over sampled direct answers moves no setting by more than 0.9. A vote on top of any harness, human or compiled, moves nothing (DocVQA typed-fields recipe 85.6 vs 85.7, $H_{\mathrm{val}}$ 84.8 vs 84.9; SlideVQA 56.4 vs 56.7; InfoSeek 39.73 vs 39.73; BLINK $\pm 0.7$). Against the formal replicate-1 harnesses on the same items, chain-of-thought with a vote is not significantly different on reflectance ($+0.7$, 33 fixed / 32 broken) or correspondence ($+1.1$, $p=0.73$), and it stays short on depth ($-10.5$, $p=0.002$), DocVQA ($-5.2$, 95\% interval $[-6.4,-4.1]$), SlideVQA ($-7.0$, $p<10^{-9}$) and InfoSeek ($-12.7$, $p<10^{-80}$; $+6.6$ over bare, so reasoning does surface some entity knowledge, below the public-content human recipe); on the earlier single builds it had won semantic correspondence (64.8 vs 54.0) and lost visual (85.5 vs 92.4, the geometric tool's cell), and it costs the strong 4B reader 2.2 on DocVQA. TTAug against its own control: DocVQA $\times$ Gemma $+1.6$ ($p=2\times10^{-4}$; text-only $+1.5$, image-only $+0.5$ n.s., low strength $+1.8$), on the human OCR recipe $+1.1$; SlideVQA $+2.1$ at $N=8$; correspondence $+2.7$ ($p=0.04$ text-only); reflectance $-1.5$ and depth $+4.0$, both n.s.\ at $n \le 134$. The matched generated-token comparison is consolidated in Table~\ref{tab:token_efficiency}. TTAug additionally requires $N$ prefills per item (5 GPU-hours per 3,568-item pass on one 3090). These prefill costs are outside the generated-token metric. Offline build costs are reported separately in App.~\ref{app:cost}.

\subsection{Retrieval and prompt optimization}
\label{app:external}

\begin{table}[h]
    \centering
    \small
    \setlength{\tabcolsep}{5pt}
    \caption{External baselines (test). GEPA uses DSPy 3.3.1, a multimodal proposer, 3,000 metric calls per run and GPT-5.5 reflection. Arrows show the unoptimised seed $\to$ the selected program.}
    \label{tab:external}
    \begin{tabular}{lrr}
        \toprule
        arm & InfoSeek $\times$ 4B & DocVQA $\times$ Gemma \\
        \midrule
        RAG-wiki & 30.9 & -- \\
        RAG-wiki (gold entity) & 48.8 & -- \\
        GEPA-prompt & 17.9 $\to$ 19.3 & 82.6 $\to$ 82.9 \\
        GEPA-content (fixed store / OCR) & 32.9 $\to$ 33.3 & \bfs{86.2} $\to$ 86.2 \\
        GEPA-structured (two modules) & 34.9 $\to$ 36.0 & -- \\
        \bottomrule
    \end{tabular}
\end{table}
\paragraph{RAG.} Knowledge base = the English Wikipedia article of each store entity (1,380 articles obtained from Wikidata sitelinks). RAG-wiki: CLIP top-1 $\to$ top-2 350-character passages by MiniLM question similarity, injected result-framed (30.87). RAG-wiki-lead (first 600 characters): 29.63. RAG-wiki-gold (gold entity): 48.75. RAG-web (DuckDuckGo snippets, 300-item stratified subset, 225 with results): 22.0 vs store-only 30.0 and \hc{} 38.3 on the same items.
\paragraph{GEPA.} DSPy 3.3.1, \texttt{dspy.GEPA} with the multimodal instruction proposer (images sent as images), train = the 150-item probe, validation = 300 disjoint practice items, 3,000 metric calls per run, GPT-5.5 reflection under a \$4 cap (actual \$0.32--0.95 per run, 9--20 reflection calls). Programs: prompt (image + question), content (plus the fixed store card or OCR text), and structured (Discriminate $\to$ Answer, both instructions evolved, feedback naming candidates, gold entity and the choice). Every accepted candidate sat within the 300-item gate's standard error of the seed. Best-validation$\to$test drops of 2--4 points are the winner's curse. The InfoSeek instructions memorised probe facts (``Sigiriya \ldots{} rock height is about 180 m''). Budgets: GEPA 3,000 rollouts vs \hc{} 3,250--6,000 per build, with reflection spend one third to one half of \hc{}'s compile + reflect.

\subsection{GEPA over the same executable artifact}
\label{app:gepa}

The loop of \S\ref{sec:method} is one search procedure: a single lineage, three reflections on a 30-trace report, a keep-better gate. To ask whether the search matters, we replace it with GEPA's \citep{gepa} and hold the artifact and tools fixed: the candidate is the full harness JSON (prompts, memory settings, discriminate and name prompts, \texttt{router\_py}, \texttt{parse\_py}), evaluated by the same runtime and student. The store is fixed (the reference teacher-written store, test 33.45 with the default harness). The reflection set is HC's 150-item probe and the validation set its 300-item gate, drawn by the same code for the same seed. The proposer is GPT-5.5 at medium reasoning reading three items with their images per call. Pareto candidate selection, strict-improvement acceptance and full-validation scoring are GEPA's own (pinned commit; \texttt{optimize\_anything} with a custom proposer). Every paid call is reserved against a per-run cap of \$2.5 and a global cap of \$10 before it is sent. The runs stopped at \$2.18--2.25 after 10--12 calls and 1,860--3,080 example evaluations. A matched HC build (seed 0, no profile, fixed store, probe 150 / gate 300) cost \$1.13 in four calls and $\approx$1,350 evaluations.

\begin{table}[h]
\centering\small
\caption{Search-step ablation on InfoSeek $\times$ Qwen3-VL-4B, fixed store, no profile (test $n=6{,}000$). Parentheses in the paired row give McNemar $p$-values. Seed 1's HC build used a 600-item gate, so the matched difference averages seeds 0 and 2 only. Both arms ran under the earlier runtime interface (v1). The comparison is historical and was not repeated under the current admissibility checks, which GEPA's seed-2 program (hard-coded entity names) would fail.}
\label{tab:gepa}
\begin{tabular}{lrrrr}
\toprule
search & seed 0 & seed 1 & seed 2 & mean (SD) \\
\midrule
HC loop & 34.1 & 40.3 & 35.6 & 36.6 (3.3) \\
GEPA & 33.5 & 37.8 & 36.6 & 35.9 (2.2) \\
HC $-$ GEPA, paired & $+0.6$ (.24) & $+2.5$ & $-1.0$ (.037) & $-0.2$ (matched) \\
\bottomrule
\end{tabular}
\end{table}

The average difference on the two matched seeds is small (HC minus GEPA: $-0.2$ points) and the spread between builds of either exceeds the difference between them. GEPA found the edits HC finds (a bounded discriminate step on ambiguous links, relevance-filtered cards, question-aware card lines). One seed never improved on the default on the gate, three of its code edits collapsed to 23 there, and its seed-2 router hard-codes three castle names from the probe, the probe-fitting the SlideVQA audit documents (App.~\ref{app:audit}). These runs do not support HC search superiority: what carries the result is the executable artifact with a decision structure to edit, the store, the trace-level evidence and the gate. Both search procedures work in this setting, with GEPA costing about twice as much.

\subsection{Typed input and output fields}
\label{app:typed}

The DSPy comparator of Table~\ref{tab:docvqa} is the \emph{un-optimised} seed program: GEPA raises its validation score from 87.3 to 90.7 and its test score by $0.0$ (86.24 $\to$ 86.19). Without OCR, the test score moves from 82.6 to 82.9. Image size is not the reason either: the seed program scores 86.25 at full resolution, and 768-px images are worth $+1.0$ to bare and $+0.5$ to the typed-OCR recipe in our runtime. The effect is the DSPy chat adapter's serialisation: a system message declaring typed input and output fields and their layout, \texttt{[[ \#\# field \#\# ]]} markers around each input, and an instruction to answer starting at the answer field and ending with a completion marker. Reproduced verbatim as a harness in our runtime it scores 81.1 without OCR (bare 72.7) and 85.3 with OCR (typed recipe 82.4), 0.9 short of the DSPy program. It is therefore an \emph{integrate}-type decision, how the inputs and the output contract are serialised, not an optimised instruction, and we expose it as a runtime option (\texttt{prompt.format = typed\_fields}) that a harness may select on any task (App.~\ref{app:dsl}). Its effect elsewhere is student- and bed-specific (Tab.~\ref{tab:typed}): $+9.0$ / $+3.3$ for Gemma, $0.0$ to $-0.7$ for the 4B on InfoSeek, $-1.3$ for the 4B on DocVQA, $-6.0$ on reflectance. DSPy's optimiser, a search over the instruction text on 150 training items, adds $+3.4$ on validation and $0.0$ on test here.
\begin{table}[htbp]
\centering\footnotesize\setlength{\tabcolsep}{5pt}\renewcommand{\arraystretch}{1.13}
\caption{Typed fields with the same items and content. Only serialization changes.}
\label{tab:typed}
\begin{tabularx}{\linewidth}{@{}YYrr@{}}
\toprule
Setting & Input & Plain & Typed \\
\midrule
DocVQA, Gemma-4-E4B & Bare & 72.7 & 81.7 \\
 & OCR recipe & 82.4 & 85.7 \\
DocVQA, Qwen3-VL-4B & Bare & 94.9 & 93.6 \\
\midrule
InfoSeek, Qwen3-VL-4B & Bare & 18.0 & 18.0 \\
 & Name card & 23.6 & 22.9 \\
 & Store only & 33.5 & 33.4 \\
 & Pick recipe & 25.0 & 24.7 \\
\midrule
LiveVQA, Qwen3-VL-4B & Bare & 14.7 & 13.7 \\
 & Pick recipe & 25.5 & 24.9 \\
\midrule
BLINK, Qwen3-VL-4B & Reflectance, bare & 50.8 & 44.8 \\
 & Correspondence, bare & 54.9 & 57.1 \\
 & Depth, bare & 82.3 & 81.5 \\
\bottomrule
\end{tabularx}
\end{table}

\subsection{Student-driven code and tool-use controls}
\label{app:agents}

We implement two agent-style controls inspired by ViperGPT and ToolLLM. These are not reproductions of their original models and training pipelines. \textbf{ViperGPT} \citep{vipergpt} has a code model write one program per query from the question and a tool API, never seeing the image. \textbf{ToolLLM} \citep{toolllm} trains a 7B model on teacher-written API-call trajectories and searches at inference with depth-first backtracking (DFSDT). What transfers to our beds is the procedure, in which the student decides which tool to call, reads the result, calls again or answers, and may backtrack: the activation, selection and argument-generation decisions studied in Sec.~\ref{sec:problem}. One runtime gives the student the harness's own tools (InfoSeek: the three CLIP candidates with their store cards; DocVQA: the page OCR and a line search; BLINK: the precomputed measurements) plus a self-query, in two modes: the student writes a Python program over the API without seeing the image (\emph{ViperGPT-S}, with a self-query fallback if the program fails), or runs a ReAct loop with give-up-and-backtrack, depth $\le 4$ and $\le 8$ nodes (\emph{DFSDT-S}, without trajectory training, with a separate trained control in App.~\ref{app:trajectory}). Zero teacher calls. A large-teacher version using GPT-5.5 to write per-query programs is run on 150-item practice subsets under a hard cap.

\begin{table}[htbp]
\centering\footnotesize\setlength{\tabcolsep}{4pt}\renewcommand{\arraystretch}{1.15}
\caption{Student-driven agent controls with the same tools. Scores use the test splits. Average gives equal weight to the five settings with HC results. Only the three performance columns available on all five are averaged.}
\label{tab:agents}
\begin{tabularx}{\linewidth}{@{}Yrrrrr@{}}
\toprule
Setting and student & \multicolumn{2}{c}{ViperGPT-S} & DFSDT-S & Tool use & HC \\
\cmidrule(lr){2-3}
 & Prose & Example & & & \\
\midrule
BLINK reflectance, 4B & 44.0 & 31.3 & 44.8 & 0\% & 64.9 \\
BLINK depth, 4B & 83.9$^{a}$ & 78.2 & 81.5 & 0\% & 95.2 \\
BLINK correspondence, 4B & 42.6 & 51.5 & 58.1 & 0\% & 67.6 \\
DocVQA, Gemma-4-E4B & 75.0 & n/a & 75.2 & 81\% & 84.9 \\
DocVQA, Qwen3-VL-4B & 44.5 & n/a & 95.0 & 97\% & n/a \\
InfoSeek, Qwen3-VL-4B & 14.1 & n/a & 19.6 & 28\% & 39.7 \\
\midrule
Average (5 paired settings) & 51.9 & -- & 55.8 & -- & 70.5 \\
\bottomrule
\end{tabularx}
\end{table}

The example column adds one example return value to the BLINK API documentation. Tool use is the share of DFSDT-S items with a non-query tool call. The three BLINK rows use Qwen3-VL-4B. Bare and human-recipe references appear in Table~\ref{tab:main}. The additional Qwen3-VL-4B DocVQA reader scores 94.9 bare.
$^{a}$ Every program in this depth arm crashed on the signal structure and fell back to a self-query. The example-documentation arm has working programs.

\begin{table}[htbp]
\centering\small\setlength{\tabcolsep}{12pt}\renewcommand{\arraystretch}{1.12}
\caption{Teacher-written programs on covered practice items. These are not full-test comparisons.}
\label{tab:teacher_programs}
\begin{tabular}{@{}llrr@{}}
\toprule
Setting & Reasoning setting & Items & Score \\
\midrule
DocVQA & Medium & 32 & 81.2 \\
DocVQA & Low & 58 & 65.5 \\
InfoSeek & Reported pilot & 44 & 29.5 \\
\bottomrule
\end{tabular}
\end{table}

The DocVQA medium-effort run costs about \$0.05 per program under a \$1.5 cap. The low-effort run costs about \$0.02 per program under a \$1 cap. Each score is evaluated on the items that run covered.

Three things hold across the beds (Table~\ref{tab:agents}). Student-written programs remain below the compiled harness across the evaluated tasks: on BLINK the programs read the measurements and the rules they write are inverted or anti-correlated (depth inverts the documented sign convention). On DocVQA the strong 4B reader replaces its own reading by regular expressions over noisy OCR and loses fifty points, an example of an unhelpful channel under the criterion in Section~\ref{sec:decisioncriteria}. On InfoSeek, the programs string-match the cards. The search never activates a measurement tool on BLINK (0 of 699 items; its $+3$ on correspondence is self-queries, a chain of thought) and does call OCR on DocVQA (81--97\% of items) but integrates the fetched lines worse than the harness's injected text under a contract (Gemma 75.2 vs 84.9; the 4B ignores what it fetched): activation and integration are separate decisions, and the compiled harness fixes both. A large model writing the program per query reaches the store-only level on InfoSeek and, at medium reasoning, the OCR-recipe level on DocVQA on the items its programs cover (at low reasoning its DocVQA programs fall below bare), at a large-model call per query costing \$0.02--0.05 each. These controls show that access to the same tools is insufficient under the tested student-driven interfaces. They do not rule out other agent training or prompting schemes.

\FloatBarrier

\Needspace{44\baselineskip}
\section{Data efficiency and weight adaptation}
\label{app:weights}
\subsection{Practice budgets and the 100-item regime}
\label{app:ladder}

\begin{figure}[!htb]
    \centering
    \includegraphics[width=\linewidth]{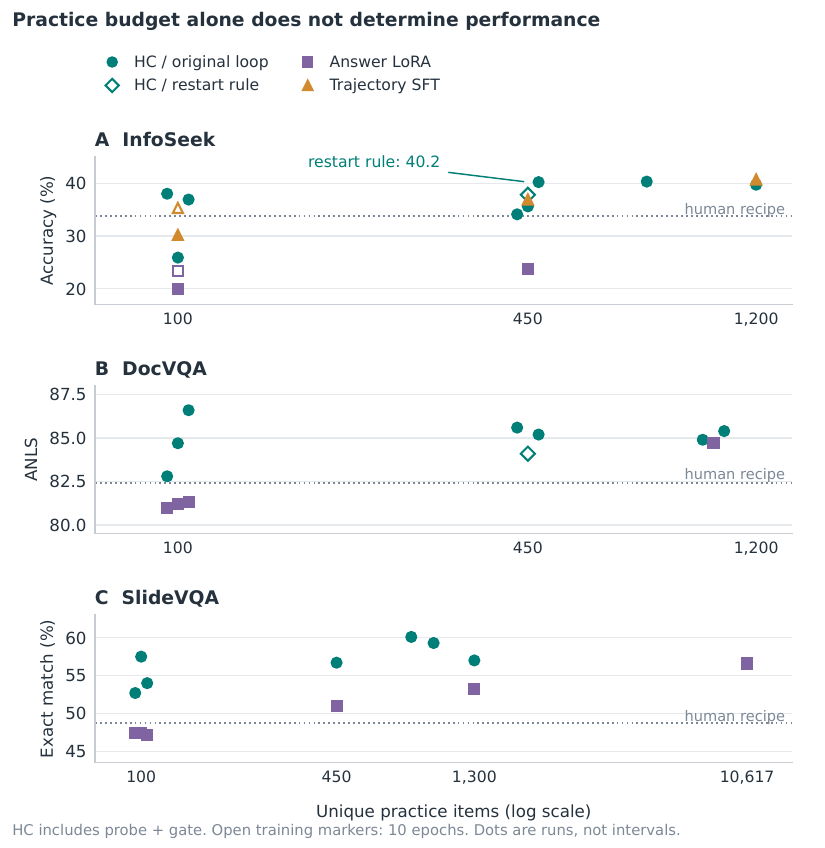}
    \caption{\textbf{Budget, build variation and loop version should not be conflated.} \hc{} dots are individual builds, with hollow diamonds marking the restart-rule procedure. Training points use the same unique practice items where matched, with 100-item training means where reported. Hollow training markers are 10-epoch variants. HC counts include probe and gate. The 100-item HC points use the current protocol, while the larger-budget points are the earlier diagnostic builds. The formal means are in Table~\ref{tab:main}. Dots are slightly offset horizontally to reveal repeated seeds. Different allocations and procedures are not connected into a scaling law. Full splits and accepted trajectory counts are in Tab.~\ref{tab:ladder}.}
    \label{fig:budget}
\end{figure}

\begin{table}[tbp]
\centering\scriptsize\setlength{\tabcolsep}{1.5pt}
\caption{Practice budgets and weight adaptation. HC counts probe / validation items. At 100 items, entries are three-seed means under the current protocol with the individual builds in parentheses (Fig.~\ref{fig:budget}). Epoch counts are shown when varied. Training counts are stated where they differ from the row budget. The DocVQA 500/500 setting contains 999 unique items.}
\label{tab:ladder}
\begin{tabular}{llrrr}
\toprule
Setting & HC probe / gate & HC & Answer-LoRA & Trajectory SFT \\
\midrule
InfoSeek & 50 / 50 & 33.6 (38.0, 25.9, 36.9) & 20.0 (3 ep) / 23.3 (10 ep) & 30.2 (3 ep) / 35.3 (10 ep) \\
 & 150 / 300, original & 34.1 / 35.6 & 23.7 & 36.9 \\
 & 150 / 600 & 40.3 & -- & -- \\
 & 600 / 600 & 39.7 & -- & 40.7 (1,199 accepted) \\
DocVQA & 50 / 50 & 84.7 (82.8, 84.7, 86.6) & 81.1 & -- \\
 & 150 / 300 & 85.6 & -- & -- \\
 & 500 / 500, interface v1 build & 84.9 & 84.7 & -- \\
 & 500 / 500, current build (Table~\ref{tab:combined_main}) & 85.4 & 84.7 & -- \\
SlideVQA & 50 / 50 & 54.7 (52.7, 57.5, 54.0) & 47.3 & -- \\
 & 150 / 300, single build & 56.7 & 51.0 & -- \\
 & 500 / 300 & 60.1 & -- & -- \\
 & 150 / 800 & 59.3 & -- & -- \\
 & 500 / 800 & 57.0 & 53.2 (1,300 labels) & -- \\
 & Full-train reference & -- & 56.6 (10,617 labels) & -- \\
BLINK reflectance & 44 / 45 per fold, single build & 64.9 & 53.7 (89 labels) & -- \\
BLINK correspondence & 147 / 147 per fold, single build & 67.6 & 70.5 (291 labels) & -- \\
BLINK depth & 41 / 42 per fold, single build & 95.2 & 82.3 (82 labels) & -- \\
LiveVQA (judged) & 150 / 300 & 37.1 & -- & 33.4 (449 accepted) \\
\bottomrule
\end{tabular}
\end{table}

\Needspace{9\baselineskip}
At 100 practice items, HC averages 33.6 / 84.7 / 54.7 on InfoSeek / DocVQA / SlideVQA under the current protocol (three builds per setting on the same 50/50 item lists as the adapters' training data, with individual builds in parentheses), above the corresponding answer-LoRA means. The comparison is not uniform across seeds: one InfoSeek build selected, on its 50-item gate, a round-1 harness that scores 25.9 on test against 26.8 for its initial harness. Increasing InfoSeek trajectory-SFT training from three to ten epochs raises its 100-item mean from 30.2 to 35.3, above HC's 33.6. The original 450-item HC builds and the restart-rule build have different outcomes despite using the same number of items. Budget alone is therefore not an explanation of the gap. Small validation sets also show substantial fluctuations between validation and test.

\subsection{Trajectory supervision at matched item budgets}
\label{app:trajectory}
For InfoSeek, GPT-5.5 writes ReAct trajectories using deterministic tools (candidate cards, lookup and finish) on the same 450 items as the original seed-0 HC build. Of 450 requests, 449 yield accepted trajectories (832 per-step rows at a teacher cost of \$4.26). Rank-8 LoRA training takes 51 GPU-minutes. The untrained control in this trajectory-specific runtime scores 30.2 and is distinct from the DFSDT-S interface in App.~\ref{app:agents}.

The trained agent scores 36.9, above answer-only LoRA on the same items (23.7) and the original HC builds (34.1 / 35.6). These values refer to the fixed harnesses used for this training comparison. At 1,200 practice items, 1,199 accepted trajectories produce 40.7, statistically tied with the 1,200-item HC build at 39.7 (paired difference $+0.9$, 95\% interval [$-0.1$, $+2.1$], $p=.10$). The 450-item result establishes an advantage over these particular source harnesses, not over every HC build at that budget.

The trained agent executes essentially a fixed program: store call then answer on 5,996/6,000 items at 450 trajectories and 5,999/6,000 at 1,200, with no substantive backtracking. These measurements support a comparison between external decision structure and decision structure learned in weights.

\subsection{Answer adaptation and task-specific training protocols}
\label{app:lora}

\textbf{DocVQA at matched labels.} An answer-only LoRA on Gemma-4-E4B (rank 8, 31 GPU-minutes, no teacher) trained on the 999 practice items the Gemma build consumed scores 84.7 ANLS on the 3,568 test items against 85.4 for a harness compiled under the current protocol on exactly those 999 items ($-0.8$ [$-1.9$, $+0.4$]), $+2.3$ over the typed human recipe ($p{=}7\!\times\!10^{-4}$) and $+12.0$ over bare. That same harness run on top of the adapter scores 86.4 ($+1.8$ [$+0.7$, $+2.8$] over the adapter, $+1.0$ [$+0.3$, $+1.7$] over the harness on the base student). The earlier interface-v1 harness (84.9) had added nothing to the adapter ($-0.03$). A thousand gold spans teach this student most of the format and the reading that the harness supplies as a contract and a channel. The current harness still adds a little on top of the weights.

\textbf{BLINK at matched labels.} For each BLINK bed an answer-LoRA (rank 8, 3 epochs, bare prompt, gold letter as target, and both images on correspondence) is trained on the items of two of the harness builds' three folds (89 / 291 / 82 labels) and scored on the held-out fold, the three folds pooled and paired against bare and the compiled harness on the same items. Reflectance 53.7 ($+3.0$ over bare, $6/2$, $p=0.29$; $-11.2$ below \hc{}, $18/33$, $p=0.049$) and depth 82.3 ($+0.0$, $8/8$; $-12.9$, $4/20$, $p=0.0015$): the measured gains over bare are not significant where the harness's edge is a rule over a measurement the student cannot make. Correspondence 70.5 ($+15.6$ over bare, $88/19$, $p=8.5\times 10^{-12}$; $+2.9$ over \hc{}, $53/40$, $p=0.21$): this is also the setting where the student's chain of thought with a vote also reaches $H_{\mathrm{val}}$ (App.~\ref{app:tts}) and the student-routed arm reads the raw matcher distances to 64.9: a skill the student carries, which labels or reasoning unlock and the harness supplies more cheaply, not a channel edge.

\textbf{LiveVQA trajectories.} The InfoSeek recipe on the news-image task: 449 accepted GPT-5.5 trajectories over the 4B build's 450 practice items (\$5.8), 842 per-step rows, LoRA rank 8, DFSDT at test with deterministic tools. Judged 33.4 (GPT-5.5 judge 32.5; containment 32.1): $+18.6$ over bare, $+7.7$ over the recipe, $-3.8$ below $H_{\mathrm{val}}$ on the same practice items ($p=5\times 10^{-5}$; $-3.2$ under the GPT-5.5 judge). Every trajectory is again the fixed program (one \texttt{clip\_candidates} call, then \texttt{finish}, with \texttt{read\_card} on 0 of 2,000 items), with one collapse the harness does not have: 126 items answered ``Netflix'' (6 correct).

\textbf{SlideVQA at matched labels.} LoRA (rank 8, LLM projections only) on the bare whole-deck input of the Qwen3-VL-4B, with the same items the harness builds consumed: 450 labels (the probe and gate) 51.0, 1,300 labels 53.2, the full 10,617-item train split 56.6, against bare 44.7 and the frozen formal replicate-1 harness 58.1. That same harness run on top of the 1,300-label adapter scores 63.7 ($+10.5$ over the adapter, $p<10^{-23}$; $+5.6$ over the harness on the base student). With the earlier round-2 harness of the 2026-09 build the full-train adapter had added nothing further (64.1 vs 63.9, $p=0.86$): the combination outperforms either route alone in this setup without identifying exactly which components the adapter learned. MMStar moves by $+0.7$ / $+1.1$ / $+3.9$ for the three adapters, so there is no drift penalty on this bed. Label use is asymmetric: the loop reads 150 items' labels as traces and the gate labels as one number per round, whereas the adapter uses the training labels directly.

\textbf{Reflectance label scaling (Fig.~\ref{fig:lora}).} A LoRA trained on IIW pairs in BLINK's format scores 52.2 with 44 labels, 56.0 / 50.0 with 147, 64.2 with 1,000, 72.4 with 5,000 and 80.6--81.3 with 20,000. \hc{} scores 64.9 using 44 probe plus 45 gate items per fold. The 89-label BLINK LoRA (53.7) uses a different training source and is shown separately.

Training data: Intrinsic Images in the Wild (5,230 photos, $\approx$875k pairwise judgments) synthesised into BLINK's format (white ring + black tag with the letter above each point, the BLINK question verbatim, majority judgment with weight $\geq 0.5$, and $\leq 4$ pairs per photo). We removed 132 photos whose perceptual hash matches a BLINK reflectance image (132 of 134 BLINK items covered), leaving a pool of 20,001 pairs. Recipe: rank 8, $\alpha$ 16, lr $5\!\times\!10^{-5}$, batch 1 $\times$ grad-accum 16, LM projections only, vision frozen, images $\leq$768 px. Training uses 44/147/1k at 3 epochs, 5k at 2, 20k at 1 ($\approx$6.5 s/step on one RTX 3090). The main comparison is in Section~\ref{sec:weights}. \hc{} on top of the 20k LoRA: transplanted harnesses 66.4, fresh 3-fold build 82.1 (teacher \$2.07), the teacher removing the tool in two folds and keeping a conservative router with 91\% precision on 22/45 items in the third. InfoSeek: the LoRA was trained on all 3,000 practice items, so for a fresh build a second LoRA was trained on 2,000 and probe/validation drawn from the held-out 1,000 (a distinct fresh-build protocol).

\Needspace{12\baselineskip}
\subsection{Label scaling and recompilation after tuning}
\label{app:weight_scaling}
\begin{figure}[!htb]
    \centering
    \includegraphics[width=\linewidth]{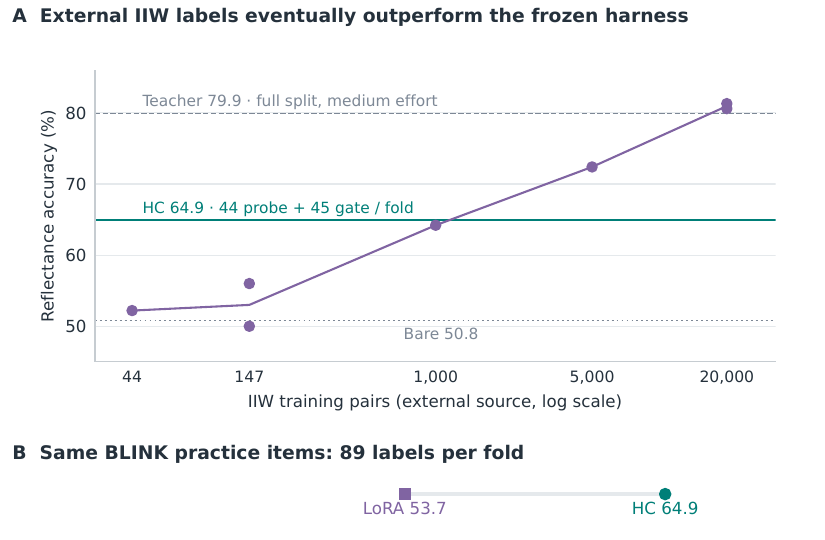}
    \caption{\textbf{Large external training sets can overtake a frozen harness.} (A) LoRA trained on contamination-audited IIW pairs. Dots show all available seeds, with a line through budget means only as a visual guide. \hc{} uses 44 probe plus 45 gate BLINK items per fold. The teacher reference is 79.9 on the full split at medium effort. (B) LoRA trained on the same 89 BLINK practice items as \hc{}, a distinct data regime.}
    \label{fig:lora}
\end{figure}

\begin{figure}[!htb]
    \centering
    \includegraphics[width=\linewidth]{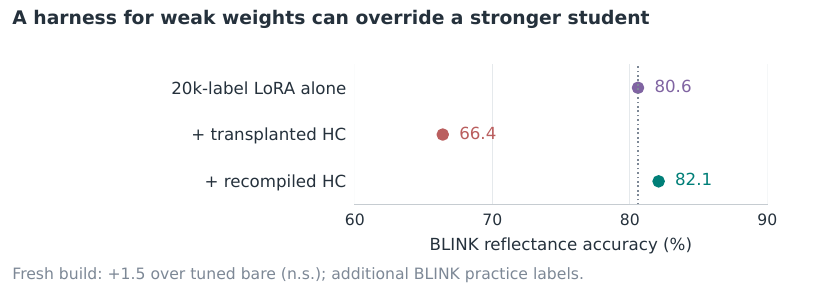}
    \caption{\textbf{Weight adaptation can reverse a harness component's value.} The old reflectance harness suppresses the 20k-label LoRA. A fresh cross-build removes the tool in two folds and uses a conservative router in the third. Recompilation consumes additional BLINK practice labels. Its $+1.5$ over the tuned bare student is not significant.}
    \label{fig:adapted}
\end{figure}

Figures~\ref{fig:lora} and \ref{fig:adapted} distinguish two questions: when more training data overtakes a fixed harness, and how that harness should change after the student is tuned.

\FloatBarrier

\section{Reliability audits and cost}
\label{app:reliability}
\subsection{Support, store and evaluation audits}
\label{app:audit}
Formal InfoSeek controls are in App.~\ref{app:rcontrols}. The support counts here concern the SlideVQA router.
For each round we count how often every router branch fires, and with what precision, on the probe, the validation set and test. The broad numeric verdicts of the 4B's round 1 fired on 11 probe items at 46\% and on 19 validation items at 16\%. The teacher removed them from the traces in round 2. The eight rules of its selected round 3 are keyed to phrases of specific probe decks (``exporter to Med'', ``credit-card handling fee''), fired on 3 probe items, 0 validation items and 0 of 2,215 test items. MiniCPM's rejected round 3 fired on 24 probe items at 96\% and on one validation item. A scan of the hooks for string literals that occur in probe items finds two to four per kept harness on this bed and up to thirteen in a LiveVQA round the gate rejected. They touch zero to two test items. Two test passes settle what the rules were worth: round 3 with its verdict rules and parser special cases removed scores 56.70 against the selected 56.66 (one item differs), and round 2, before the rules were added, scores 56.98 ($p=0.53$). The whole gain over $H_0$ is the ranker fusion, the contract and the parser of rounds 1--2. A subsequent runtime safeguard reports per-branch validation support to the teacher and drops branches with fewer than ten validation firings before test.

\paragraph{Store isolation.} A store audit found fact cards for 120/1,380 entities in an earlier name-card baseline. All names-only controls cited as such were rerun with a clean names-only store, changing by 0.3--0.6 points. Teacher fact-card generation uses entity names, not evaluation questions or answers. Store-only controls on the reference and rewritten stores separate content changes from routing changes (App.~\ref{app:infoseek}).

\paragraph{Evaluation safeguards.} Image inputs to multimodal proposers are sent as images. Reasoning-mode defaults and answer-letter normalization are fixed across compared arms. LoRA comparisons requiring a fresh compiler probe use practice items held out from that adapter's training. The LiveVQA judge comparison and the distinction between containment, accuracy and abstention-sensitive scores are documented in App.~\ref{app:livevqa}. Failed builds and unfavorable transfer results remain with their corresponding experiments rather than in a separate catalog.

\subsection{Exploratory VLA boundary case: LIBERO-plus}
\label{app:vla}
LIBERO-plus~\citep{liberoplus} evaluates robustness of language-conditioned robot policies under seven perturbation types. Our exploratory setup uses SmolVLA-0.45B and X-VLA-0.9B in LeRobot 0.6.1, with disjoint 208-episode probe and validation sets and a 600-episode test: 560 perturbed episodes plus 40 added mismatched-instruction episodes. On the latter, only an abort within the first quarter of the horizon counts as a correct refusal. The runtime exposes instruction canonicalization against the suite's ten training instructions (MiniLM), OWLv2 detections, motion/stall signals and a step budget. The teacher may write instruction overrides and stopping rules. X-VLA uses absolute end-effector control. No teacher runs online. These scores use our augmented evaluation, not the benchmark's standard leaderboard protocol.

\begin{table}[htbp]
\centering\small\setlength{\tabcolsep}{4pt}
\caption{Exploratory policy evaluation (600 episodes). ``Base $H_0$'' is the first SmolVLA harness, reused unchanged on the other checkpoints. It is a diagnostic artifact, distinct from the validation-selected own build.}
\label{tab:vla_main}
\begin{tabular}{lrrrr}
\toprule
Policy & Bare & Canonical only & Base $H_0$ & Own $H_{\mathrm{val}}$ \\
\midrule
SmolVLA-0.45B & 3.8 & 33.2 & 37.0 & 34.5 \\
SmolVLA, tuned on perturbations & 11.8 & 59.0 & 64.7 & 63.2 \\
X-VLA-0.9B & 59.7 & 60.7 & 63.3 & -- \\
\bottomrule
\end{tabular}
\end{table}

Most of the base SmolVLA gain is recovered by removing perturbation tags from the instruction. X-VLA does not show this collapse: canonicalization adds little, and the transplanted harness adds a few points through its external control settings. Providing its decisions as appended instruction text scores 60.2. The base SmolVLA validation-selected round is worse than its first candidate on test (34.5 versus 37.0, n.s.), and object detections do not distinguish mismatched instructions reliably.

\paragraph{Success, refusal and false aborts are separate outcomes.} The mixed score of Table~\ref{tab:vla_main} is $(560/600)\,S + (40/600)\,R$, where $S$ is the completion rate on the 560 valid perturbed episodes and $R$ the correct-refusal rate on the 40 mismatched ones. Refusal alone can contribute at most 6.7 points. Table~\ref{tab:vla_decomp} separates them from the archived per-episode records (no new rollouts): the selected or transplanted HC arms in Table~\ref{tab:vla_main} have at most six correct refusals, contributing at most 1.0 point to their mixed score. For base SmolVLA, those refusals account for 1.0 of the roughly 1.3-point gain over canonicalization. The remaining gain is valid-episode completion. The transplanted $H_0$ and the tuned selected harness gain through completion alone. The additional, unsuccessful tuned own-$H_0$ diagnostic in Table~\ref{tab:vla_decomp} has 13 correct refusals, contributing 2.2 points, and is not one of those selected arms. The one exception is instructive: the base-SmolVLA validation-selected round aborts 62 valid episodes (false aborts) and loses 3.7 points of completion against its own $H_0$ while gaining six refusals. Partial-progress stages are not logged on LIBERO and are not evaluated. These results use the project's augmented evaluation, not the official LIBERO-plus protocol.

\begin{table}[htbp]
\centering\scriptsize\setlength{\tabcolsep}{4pt}
\caption{LIBERO-plus outcomes decomposed from the archived records: completion on the 560 valid episodes, false aborts on valid episodes (all within the first quarter of the budget), correct refusals among the 40 mismatched instructions (abort within the first quarter), and the mixed score of Table~\ref{tab:vla_main}. Human recipe v2 refuses by detection and pays for it with 106 false aborts.}
\label{tab:vla_decomp}
\begin{tabular}{@{}llrrrr@{}}
\toprule
Checkpoint & arm & valid completion (\%) & false aborts & correct refusals / 40 & mixed \\
\midrule
SmolVLA base & bare & 4.1 & 0 & 0 & 3.8 \\
SmolVLA base & canonicalization & 35.5 & 0 & 0 & 33.2 \\
SmolVLA base & human recipe v2 & 30.7 & 106 & 9 & 30.2 \\
SmolVLA base & HC $H_0$ & 39.6 & 0 & 0 & 37.0 \\
SmolVLA base & HC $H_{\mathrm{val}}$ (round 2) & 35.9 & 62 & 6 & 34.5 \\
SmolVLA plus-tuned & bare & 12.7 & 0 & 0 & 11.8 \\
SmolVLA plus-tuned & canonicalization & 63.2 & 0 & 0 & 59.0 \\
SmolVLA plus-tuned & transplanted base $H_0$ & 69.3 & 0 & 0 & 64.7 \\
SmolVLA plus-tuned & human recipe v2 & 51.4 & 106 & 9 & 49.5 \\
SmolVLA plus-tuned & own HC $H_0$ & 43.4 & 200 & 13 & 42.7 \\
SmolVLA plus-tuned & own HC $H_{\mathrm{val}}$ (round 2) & 67.7 & 0 & 0 & 63.2 \\
X-VLA-0.9B & bare & 63.9 & 0 & 0 & 59.7 \\
X-VLA-0.9B & canonicalization & 65.0 & 0 & 0 & 60.7 \\
X-VLA-0.9B & transplanted base $H_0$ & 67.9 & 0 & 0 & 63.3 \\
X-VLA-0.9B & student-routed & 64.5 & 0 & 0 & 60.2 \\
\bottomrule
\end{tabular}
\end{table}

\paragraph{Matched protocol comparison on LIBERO-plus.} The reflection-protocol comparison of Appendix~\ref{app:formal} was repeated on the base SmolVLA checkpoint under the VLA runtime (whose router already receives only observation-derived inputs) with the original mixed-score gate: three builds (replicates 1--3), each with a fresh $H_0$ and the two protocols run from it, every arm re-rolling $H_0$ on the 600 test episodes before its selected harness (Table~\ref{tab:vla_matched}). Rollouts are not deterministic: identical harnesses re-rolled on the same 208 gate episodes differ by up to three points (e.g.\ 33.2 / 33.2 / 30.3 in replicate 1), and two test rollouts of the same $H_0$ agree on 87--88\% of episodes with a net difference under one point, so gate acceptances of under three points lie within rollout noise. All three fresh initial harnesses were weak (23--30 mixed) because they aborted 200--279 of the 560 valid episodes by detector refusal, and both protocols gained mainly by removing those false aborts: refusals fall in every selected arm, so no gain comes from the refusal term. The reported protocol (A) selected the better harness in replicates 2 ($+6.7$, $p<10^{-3}$) and 3 ($+4.0$, $p=0.06$) and kept $H_0$ in replicate 1, where protocol B selected a round that is $+5.0$ better ($p=0.02$). Means over the three builds are $H_0$ 26.2, A 34.2, B 32.3. The comparison is therefore inconclusive at three builds and is not used for any main-text claim. It shows that the gate's three-point rollout noise, not the reflection protocol, limits selection on this bed. Teacher cost: \$1.94 (A) and \$1.81 (B) for the three builds.

\begin{table}[htbp]
\centering\scriptsize\setlength{\tabcolsep}{4pt}
\caption{LIBERO-plus matched protocol comparison on the base SmolVLA checkpoint: test outcomes of the shared $H_0$ (re-rolled in each arm) and of the harness each protocol selected, with completion on the 560 valid episodes, false aborts, correct refusals among the 40 mismatched instructions, and the paired difference against the arm's own $H_0$ rollout (exact McNemar).}
\label{tab:vla_matched}
\begin{tabular}{@{}llrrrrrl@{}}
\toprule
Replicate & arm & round & mixed & completion (\%) & false aborts & refusals /40 & vs own $H_0$ rollout \\
\midrule
1 & $H_0$ (rollout in A) & -- & 28.8 & 28.2 & 200 & 15 & -- \\
1 & A selected & 0 & 29.3 & 28.8 & 200 & 15 & $+0.5$ (39/36, $p=0.818$) \\
1 & $H_0$ (rollout in B) & -- & 29.7 & 29.1 & 200 & 15 & -- \\
1 & B selected & 2 & 34.3 & 35.9 & 91 & 5 & $+4.7$ (86/58, $p=0.0241$) \\
2 & $H_0$ (rollout in A) & -- & 23.7 & 22.3 & 279 & 17 & -- \\
2 & A selected & 1 & 36.3 & 37.3 & 71 & 9 & $+12.7$ (111/35, $p=2.09e-10$) \\
2 & $H_0$ (rollout in B) & -- & 23.2 & 21.8 & 272 & 17 & -- \\
2 & B selected & 2 & 29.7 & 28.9 & 200 & 16 & $+6.5$ (75/36, $p=0.000272$) \\
3 & $H_0$ (rollout in A) & -- & 25.8 & 24.8 & 232 & 16 & -- \\
3 & A selected & 1 & 36.8 & 39.5 & 0 & 0 & $+11.0$ (106/40, $p=4.45e-08$) \\
3 & $H_0$ (rollout in B) & -- & 25.8 & 24.6 & 229 & 17 & -- \\
3 & B selected & 3 & 32.8 & 32.9 & 129 & 13 & $+7.0$ (85/43, $p=0.000259$) \\
\bottomrule
\end{tabular}
\end{table}

These results illustrate interface repair and a limited external-control benefit. They do not isolate long-horizon planning, recovery across subtasks, or real industrial deployment. We therefore keep this as a boundary case outside the seven-setting VQA average. A stronger robotics claim would require task-level sequencing and recovery measurements beyond aggregate perturbed-policy success. The exploratory rollout cost was approximately 19 GPU-hours and is not included in the VQA build-cost range.

\subsection{Compilation and deployment cost}
\label{app:cost}

\begin{table}[htbp]
\centering\small\setlength{\tabcolsep}{4pt}
\caption{Representative VQA runtime costs, measured on historical reference records and not re-measured on the formal interface-v2 artifacts. HC output tokens include all student passes. Calls below one reflect direct code verdicts. CoT is a single reasoning rollout. The multi-sample CoT+vote comparison is in Table~\ref{tab:token_efficiency}. The InfoSeek and DocVQA cost records correspond to the earlier 39.7- and 85.7-score builds, respectively.}
\label{tab:cost}
\begin{tabular}{lrrrr}
\toprule
Setting & HC calls/item & HC output tokens & CoT output tokens & Teacher \$ / build \\
\midrule
InfoSeek & 1.6 & 4.9 & 262 & 6.9 (with store) \\
DocVQA & 1.0 & 11.2 & 97 & 0.9 \\
BLINK reflectance & 0.08 & 0.1 & 237 & 2.5 \\
BLINK correspondence & 0.45 & 0.7 & 278 & 3.0 \\
BLINK depth & 0.07 & 0.1 & 255 & 1.8 \\
\bottomrule
\end{tabular}
\end{table}

InfoSeek fact-store regeneration uses 138 teacher calls (approximately 230k tokens at low reasoning effort), costing about \$6 for a reusable store. Compilation and three reflections add approximately 65k tokens and \$1. Thus a \$1 build with an existing store and a \$7 build with a fresh store are different cost regimes. These are GPT-5.5 charges for the reported configurations, not model-independent prices. Student compute is measured on RTX 3090 GPUs: roughly 25 GPU-minutes for an InfoSeek build and 1--35 for a BLINK cross-build, compared with 2.25 GPU-hours for the 20k-pair reflectance LoRA. Search, trajectory generation and judging costs are reported with their respective comparisons. Deployment uses zero teacher calls. Output-token counts do not include image-prefill or tool-computation costs.

\paragraph{Accounting for Table~\ref{tab:token_efficiency}.}
The CoT-with-voting comparison uses five reasoning samples on the content tasks and nine on BLINK. Generated-token counts include all sampled rollouts and all student passes. A code-only verdict contributes zero student tokens. Table~\ref{tab:token_efficiency} pairs the scores and generated-token counts from the same formal-harness executions with their CoT+vote comparators. These counts do not measure latency. Table~\ref{tab:build_registry} identifies the reference harnesses, while the historical single-rollout CoT costs in Table~\ref{tab:cost} belong to a separate comparison.

\FloatBarrier

\end{document}